\documentclass[11pt,oneside,pagesize=auto]{scrartcl}

\newcommand{\titl}{Sobolev Norm Learning Rates for Regularized Least-Squares Algorithms}

\newcommand{\keyw}{statistical learning theory, regularized kernel methods, least-squares regression, interpolation norms, uniform convergence, learning rates}
\usepackage{tocbasic}
\usepackage[draft=false]{scrlayer-scrpage}
\usepackage[onehalfspacing]{setspace}

\KOMAoptions{mpinclude=false}
\KOMAoptions{headsepline=false}
\KOMAoptions{footsepline=false}

\KOMAoptions{titlepage=false}

\KOMAoptions{bibliography=totoc}

\KOMAoptions{abstract=true}

\KOMAoptions{BCOR=0cm, DIV=12}

\addtokomafont{title}{\boldmath}
\addtokomafont{section}{\boldmath}
\addtokomafont{subsection}{\boldmath}
\addtokomafont{paragraph}{\boldmath}
\addtokomafont{sectionentry}{\boldmath}

\usepackage[latin1]{inputenc} 
\usepackage[T1]{fontenc}      
\usepackage{lmodern}          
\usepackage[english]{babel}
\usepackage{microtype}        
\usepackage{enumitem}

\usepackage{amsmath}                 
\usepackage{amssymb}
\usepackage{amsthm}
\usepackage{mathtools} 
\usepackage{bbm} %

\usepackage[numbers,square]{natbib}

\usepackage{multicol}
\usepackage{multirow}
\usepackage{array}

\usepackage{tikz}

\usepackage[
pdfusetitle, 
pdfkeywords={\keyw}, 
backref = false, 
hidelinks,
pdfpagelabels = true, 
pdfstartview = FitH, 
bookmarksopen = true, 
bookmarksnumbered = true, 
plainpages = false, 
hypertexnames = false
] {hyperref}

\usepackage{comment}

\setlist{itemsep=0pt}
\setenumerate[1]{label={(\roman*)}}
\setenumerate[2]{label={(\alph*)}}

\newtheoremstyle{famous} 
{} 
{} 
{\itshape}
{} 
{\bfseries\sffamily\boldmath} 
{} 
{\newline} 
{\boldmath\thmnumber{\begin{footnotesize}#2\end{footnotesize} }\thmnote{#3}}

\newtheoremstyle{kursiv}
{} 
{} 
{\itshape} 
{} 
{\bfseries\sffamily\boldmath} 
{} 
{ } 
{\boldmath\thmnumber{\begin{footnotesize}#2\end{footnotesize} }\thmname{#1 }\thmnote{(#3)}}
  
\newtheoremstyle{normal}
{} 
{} 
{\rmfamily} 
{} 
{\bfseries\sffamily\boldmath} 
{} 
{ } 
{\boldmath\thmnumber{\begin{footnotesize}#2\end{footnotesize} }\thmname{#1 }\thmnote{(#3)}}
  
\newtheoremstyle{oNum}
{} 
{} 
{\rmfamily} 
{} 
{\bfseries\sffamily\boldmath} 
{} 
{\newline} 
{\thmname{#1 }\thmnote{(#3)}}
  
\makeatletter
\def\@endtheorem{\endtrivlist}
\makeatother
   
\theoremstyle{kursiv}
\newtheorem{thm}{Theorem}[section]
\newtheorem{cor}[thm]{Corollary}
\newtheorem{lem}[thm]{Lemma}

\newtheorem{rem}[thm]{Remark}

\theoremstyle{normal}

\theoremstyle{famous}

\ifdefined\verboseproofs

	\excludecomment{shortproof}

\else
	\excludecomment{longproof}

\fi

\DeclareFontFamily{U}{matha}{\hyphenchar\font45}
\DeclareFontShape{U}{matha}{m}{n}{
      <5> <6> <7> <8> <9> <10> gen * matha
      <10.95> matha10 <12> <14.4> <17.28> <20.74> <24.88> matha12
      }{}
\DeclareSymbolFont{matha}{U}{matha}{m}{n}
\DeclareFontFamily{U}{mathx}{\hyphenchar\font45}
\DeclareFontShape{U}{mathx}{m}{n}{
      <5> <6> <7> <8> <9> <10>
      <10.95> <12> <14.4> <17.28> <20.74> <24.88>
      mathx10
      }{}
\DeclareSymbolFont{mathx}{U}{mathx}{m}{n}
\DeclareMathDelimiter{\vvvert}{0}{matha}{"7E}{mathx}{"17}

\DeclareNewTOC[%
type=todo,%
types=todos,%
name=Todo,%
listname={Todo Liste}%
]{lotodo}
\setuptoc{lotodo}{chapteratlist} 

\newcounter{todono}
\renewcommand{\thetodono}{(\arabic{todono})}

\newcommand{\todo}[2]{\stepcounter{todono}
$^{\text{\thetodono}}$\marginline{
\begin{singlespace}
\footnotesize
\textbf{{\tiny\thetodono} #1}\par #2
\end{singlespace}}%
\addcontentsline{lotodo}{table}{#1 (#2)}}

\renewcommand{\d}{{\mathrm d}}

\newcommand{\R}{{\mathbb R}}

\newcommand{\N}{{\mathbb N}}

\newcommand{\E}{{\mathbb E}}

\DeclareMathOperator{\Id}{Id}

\DeclareMathOperator{\ran}{ran}

\DeclareMathOperator{\tr}{tr}
\DeclareMathOperator*{\argmin}{\arg\!\min}

\newcommand{\algBorel}[1]{\mathcal{B}(#1)} 

\newcommand{\cuball}[1]{\overline{B}_{#1}}

\newcommand{\indicator}[1]{\mathbbm 1_{#1}}

\newcommand{\sfrac}[2]{{#1}/{#2}}

\ifdefined\jmlr
	\newcommand{\jmlrtext}[2]{#1}
	\newcommand{\eqnr}[1]{#1}
	\let\cites\citet
	\let\citeidpar\citeyearpar

\else
	\newcommand{\jmlrtext}[2]{\ignorespaces#2\ignorespaces}
	\newcommand{\eqnr}[1]{(#1)}
	\let\cites\citep
	\let\citeidpar\citep

	\let\endproof\endproof
\fi

\newcommand{\optFZ}[1]{f^\ast_{#1}}
\newcommand{\optFP}{\optFZ{P}}
\newcommand{\optRegFP}[1][\lambda]{f_{P,#1}}
\newcommand{\optRegFD}[1][\lambda]{f_{D,#1}}

\newcommand{\g}{g}
\newcommand{\gP}{g_P}
\newcommand{\gD}{g_D}

\newcommand{\effdimMas}{\mathcal{N}_\nu}

\newcommand{\I}{I}           
\renewcommand{\S}{S}         
\newcommand{\T}{T}           
\newcommand{\rT}{C}          

\newcommand{\Tmas}{\T_\nu}
\newcommand{\rTmas}{\rT_\nu}
\newcommand{\Smas}{\S_\nu}
\newcommand{\Imas}{\I_\nu}

\newcommand{\rTX}{\rT_{x}}

\newcommand{\eigw}{\mu}      
\newcommand{\eigv}{e}        

\newcommand{\parSourceCond}{\beta}
\newcommand{\parPowerNorm}{\gamma}
\newcommand{\parEigDecay}{p}
\newcommand{\parEmbedding}{\alpha}

\newcommand{\constSourceCond}{B}
\newcommand{\constEmbedding}{A}
\newcommand{\constEigDecay}{C}
\newcommand{\constEigDecayLB}{\MakeLowercase{\constEigDecay}}
\newcommand{\constEffDim}{D}
\newcommand{\constInftyBound}{B_\infty}

\newcommand{\lebesgue}{\mu}
\newcommand{\parBesovSourceCond}{s}
\newcommand{\parBesovRKHS}{r}
\newcommand{\parBesovNorm}{t}

\newcommand{\zvi}{\xi}
\newcommand{\varb}{\sigma}
\newcommand{\supb}{L}

\newcommand{\constDensityUB}{G}
\newcommand{\constDensityLB}{g}

\newcommand{\indexBound}[1][\lambda]{A_{#1,\tau}}
\newcommand{\varLB}{\bar{\sigma}}

\newcommand{\constValidString}{U}
\newcommand{\constLB}{C}

\renewcommand{\todo}[2]{} 

\title{\titl}
\author{Simon Fischer and Ingo Steinwart}
\publishers{\normalsize Institute for Stochastics and Applications\\
Faculty 8: Mathematics and Physics\\
University of Stuttgart\\
D-70569 Stuttgart Germany \\
\texttt{\small $\{$simon.fischer, ingo.steinwart$\}$@mathematik.uni-stuttgart.de}}
\date{\normalsize\today}

\begin{document}


\maketitle
 
\begin{abstract}
Learning rates for least-squares regression are typically expressed in terms of $L_2$-norms. In this paper we extend these rates to norms stronger than the $L_2$-norm without requiring the regression function to be contained in the hypothesis space. In the special case of Sobolev reproducing kernel Hilbert spaces used as hypotheses spaces, these stronger norms coincide with fractional Sobolev norms between the used Sobolev space and $L_2$. As a consequence, not only the target function but also some of its derivatives can be estimated without changing the algorithm. From a technical point of view, we combine the well-known integral operator techniques with an embedding property, which so far has only been used in combination with empirical process arguments. This combination results in new finite sample bounds with respect to the stronger norms. From these finite sample bounds our rates easily follow. Finally, we prove the asymptotic optimality of our results in many cases.
\end{abstract} 





\section{Introduction}\label{sec:intro}

Given a data set $D=\{(x_i,y_i)\}_{i=1}^n$ independently sampled from an unknown distribution $P$ on $X\times Y$, the goal of non-parametric least-squares regression is to estimate the conditional mean function $\optFP:X\to Y$ given by $\optFP(x) \coloneqq \E(Y|X=x)$. The function $\optFP$ is also known as regression function, we refer to \citet{GyKoKrWa2002} for basic information as well as various algorithms for this  problem. In this work, we focus on kernel-based regularized least-squares algorithms, which are also known as least-squares support vector machines (LS-SVMs), see e.g.\ \citet{StCh2008}. Recall that LS-SVMs construct a predictor $\optRegFD$ by solving the convex optimization problem
\begin{equation}\label{eq:intro:optimization_problem}
\optRegFD = \argmin_{f\in H} \Bigl\{\lambda\|f\|_H^2 + \frac{1}{n}\sum_{i=1}^n(y_i - f(x_i))^2\Bigr\}\;\;,
\end{equation}
where a reproducing kernel Hilbert space (RKHS) $H$ over $X$ is used as hypothesis space and $\lambda>0$ is the so-called regularization parameter. For a definition and basic properties of RKHSs see e.g.\ \cites[Chapter~4]{StCh2008}. Probably the most interesting theoretical challenge for this problem is to establish learning rates, either in expectation or in probability, for the generalization error
\begin{equation}\label{eq:intro:generalization_error}
\|\optRegFD -\optFP\|\;\;.
\end{equation}
In this paper, we investigate \eqref{eq:intro:generalization_error} with respect to the norms of a continuous scale of suitable Hilbert spaces $[H]^\parPowerNorm$ with $H \subseteq [H]^\parPowerNorm \subseteq L_2$ in the \emph{hard learning} scenario $\optFP\not\in H$. For the sake of simplicity, we assume $[H]^0 = L_2$ and $[H]^1 = H$ for this introduction, see Section~\ref{sec:pre} for an exact definition.

Let us briefly compare the two main techniques previously used in the literature to establish learning rates for \eqref{eq:intro:generalization_error}: the \emph{integral operator} technique \citep[see e.g.,][and references therein]{DeCaRo2005,DeRoCaDeOd2005,DeRoCa2006,BaPeRo2007,SmZh2007,CaDe2007,BlM2017,DiFoHs2017, LiRuRoCe2018,LiCe2018} and the \emph{empirical process} technique \citep[see e.g.,][and references therein]{MeNe2010,StCh2008,StHuSc2009}. 
An advantage of the integral operator technique is that it can provide learning rates for \eqref{eq:intro:generalization_error} with respect to a continuous scale of $\parPowerNorm$, including the $L_2$-norm case $\parPowerNorm = 0$ \citep[see e.g.,][]{BlM2017,LiRuRoCe2018}. In addition, it can be used to establish learning rates for \emph{spectral regularization algorithms} \citep[see e.g.,][]{BaPeRo2007,BlM2017,LiRuRoCe2018} and further kernel-based learning algorithms \citep[see e.g.,][]{M2019,LiCe2018a,PiRuBa2018a,MBl2018,MNeRo2019}.
On the other hand, the empirical process techniques can so far only handle the $L_2$-norm in \eqref{eq:intro:generalization_error}, but in the hard learning scenario  $\optFP\not\in H$, which is rarely investigated by the integral operator technique, it provides the fastest, and in many cases minimax optimal, $L_2$-learning rates for \eqref{eq:intro:generalization_error}, see \cite{StHuSc2009}. This advantage of the empirical process technique in the hard learning scenario is based on the additional consideration of some \emph{embedding property} of the RKHS, which has hardly been considered in combination with the integral operator technique so far. In a nutshell, this embedding property allows for an improved bound on the $L_\infty$-norm of the regularized population predictor. In addition, the empirical process technique can be easily applied to learning algorithms \eqref{eq:intro:optimization_problem} in which the least-squares loss function is replaced by other convex loss functions, see  e.g.\ \cite{FaSt2018} for expectile regression and \cite{EbSt2013} for quantile regression.

In the present manuscript, which is an improvement of its first version \cite{FiSt2017}, we apply the integral operator technique in combination with some embedding property, see \eqref{eq:res:embedding_property} in Section~\ref{sec:res} below for details, to learning scenarios including the case $\optFP\not\in H$. Recall that such embedding properties---as far as we know---have only been used by \citet{StHuSc2009}, \citet{DiFoHs2017}, and \citet{PiRuBa2018a}. By doing so, we extend and improve the results of \citet{BlM2017} and \citet{LiRuRoCe2018}. To be more precise, we extend the results of \cite{BlM2017}, who only considered the case $\optFP\in H$, to the hard learning case and the largest possible scale of $\parPowerNorm$. Moreover, compared to \cite{LiRuRoCe2018} we obtain faster rates of convergence for \eqref{eq:intro:generalization_error}, if the RKHS enjoys a certain embedding property. In the hard learning scenario, we obtain, as a byproduct, the $L_2$-learning rates of \cite{StHuSc2009}, as well as the very first $L_\infty$-norm learning rates in the hard learning scenario. For a more detailed comparison with the literature see Section~\ref{sec:comparison} and in particular Table~\ref{tab:comparison:rates} and Figure~\ref{fig:comparison:L2rates}. 
Finally, we prove the minimax optimality of our $[H]^\parPowerNorm$-norm learning rates for all combinations of $H$ and $P$, for which the optimal $L_2$-norm learning rates are known.

The rest of this work is organized as follows: We start in Section~\ref{sec:pre} with an introduction of notations and general assumptions. In Section~\ref{sec:res} we present our learning rates. The consequences of our results for the special case of a Sobolev/Besov RKHS $H$ can be found in Section~\ref{sec:besov}. Note that in this case $[H]^\parPowerNorm$ coincide with the classical Besov spaces and the corresponding norms have a nice interpretation in terms of derivatives. Finally, we compare our result with other contributions in Section~\ref{sec:comparison}. All proofs can be found in Section~\ref{sec:proof}.

\subsection*{Acknowledgment} 
%
The authors are especially grateful to Nicole Mücke for pointing them to the article of \citet*{LiRuRoCe2018}.
Moreover, the authors thank the International Max Planck Research School for Intelligent Systems (IMPRS-IS) for supporting Simon Fischer. 
%
%


\section{Preliminaries}\label{sec:pre}

Let $(X,\mathcal{B})$ be a measurable space used as \emph{input space}, $Y=\R$ be the \emph{output space}, and $P$ be an \emph{unknown} probability distribution on $X\times \R$ with 
\begin{equation}\label{eq:pre:second_moment}
|P|_2^2 \coloneqq \int_{X\times \R} y^2\ \d P(x,y)<\infty\;\;.
\end{equation}
Moreover, we denote the marginal distribution of $P$ on $X$ by $\nu \coloneqq P_X$. In the following, we fix a (regular) conditional probability $P(\,\cdot\,|x)$ of $P$ given $x\in X$. Since the conditional mean function $\optFP$ is only $\nu$-almost everywhere uniquely determined we use the symbol $\optFP$ for both, the $\nu$-equivalence class and for the representative
\begin{equation}\label{eq:pre:regression_function}
\optFP(x)=\int_\R y\ P(\d y|x)\;\;.
\end{equation}
If we use another representative we will explicitly point this out.


In the following, we fix a separable RKHS $H$ on $X$ with respect to a measurable and bounded kernel $k$. Let us recall some facts about the interplay between $H$ and $L_2(\nu)$. Some of the following results have already be shown by \citet{SmZh2004,SmZh2005} and \citet{DeRoCa2006,DeRoCaDeOd2005}, but we follow the more recent contribution of \citet{StSc2012} because of its more general applicability. According to \cites[Lemma~2.2, Lemma~2.3]{StSc2012} and \cites[Theorem~4.27]{StCh2008} the---not necessarily injective---embedding $\Imas:H\to L_2(\nu)$, mapping a function $f\in H$ to its $\nu$-equivalence class $[f]_\nu$, is well-defined, Hilbert-Schmidt, and the Hilbert-Schmidt norm satisfies 
\begin{equation*}
\|\Imas\|_{\mathcal{L}_2(H,L_2(\nu))} = \|k\|_{L_2(\nu)} \coloneqq \biggl(\int_X k(x,x)\ \d\nu(x)\biggr)^{\sfrac{1}{2}} < \infty\;\;.
\end{equation*}
Moreover, the adjoint operator $\Smas \coloneqq \Imas^\ast: L_2(\nu)\to H$ is an integral operator with respect to the kernel $k$, i.e. for $f\in L_2(\nu)$ and $x\in X$ we have
\begin{equation}\label{eq:pre:integral_operator}
(\Smas f)(x) = \int_{X} k(x,x')f(x')\ \d\nu(x')\;\;.
\end{equation}
Next, we define the self-adjoint and positive semi-definite integral operators
\[
\Tmas \coloneqq \Imas\Smas:L_2(\nu)\to L_2(\nu)\qquad\text{ and }\qquad\rTmas \coloneqq \Smas\Imas:H\to H\;\;.
\]
These operators are trace class and their trace norms satisfy
\begin{equation*}
\|\Tmas\|_{\mathcal{L}_1(L_2(\nu))} = \|\rTmas\|_{\mathcal{L}_1(H)} = \|\Imas\|_{\mathcal{L}_2(H, L_2(\nu))}^2 = \|\Smas\|_{\mathcal{L}_2(L_2(\nu),H)}^2\;\;.  
\end{equation*}
If there is no danger of confusion we write $\|\cdot\|$ for the operator norm, $\|\cdot\|_2$ for the Hilbert-Schmidt norm, and $\|\cdot\|_1$ for the trace norm. The spectral theorem for self-adjoint compact operators yields an at most countable index set $I$, a non-increasing summable sequence $(\eigw_i)_{i\in I}\subseteq (0,\infty)$, and a family $(\eigv_i)_{i\in I}\subseteq H$, such that $([\eigv_i]_\nu)_{i\in I}$ is an orthonormal basis (ONB) of $\overline{\ran\Imas}\subseteq L_2(\nu)$ and $(\eigw_i^{\sfrac{1}{2}}\,\eigv_i)_{i\in I}$ is an ONB of $(\ker\Imas)^\perp \subseteq H$ with
\begin{equation}\label{eq:pre:spectral}
\Tmas = \sum_{i\in I}\eigw_i\,\langle\,\cdot\,,[\eigv_i]_\nu\rangle_{L_2(\nu)} [\eigv_i]_\nu
\qquad\text{ and }\qquad
\rTmas=\sum_{i\in I}\eigw_i\,\langle\,\cdot\,,\eigw_i^{\sfrac{1}{2}}\,\eigv_i\rangle_H\, \eigw_i^{\sfrac{1}{2}}\,\eigv_i\;\;,
\end{equation}
see \cites[Lemma~2.12]{StSc2012} for details. Since we are mainly interested in the hard learning scenario $\optFP\not\in H$ we exclude finite $I$ and assume $I=\N$ in the following.


Let us recall some intermediate spaces introduced by \citet[Equation~\eqnr{36}]{StSc2012}. We call them \emph{power spaces}. For $\parEmbedding\geq 0$, the \emph{$\parEmbedding$-power space} is defined by
\[
[H]_\nu^\parEmbedding \coloneqq \biggl\{\sum_{i\geq 1}a_i\eigw_i^{\sfrac{\parEmbedding}{2}}[\eigv_i]_\nu:\ (a_i)_{i\geq 1}\in\ell_2(\N)\biggr\} \subseteq L_2(\nu)
\]
and equipped with the \emph{$\parEmbedding$-power norm}
\[
\biggl\|\sum_{i\geq 1}a_i\eigw_i^{\sfrac{\parEmbedding}{2}} [\eigv_i]_\nu\biggr\|_{[H]_\nu^\parEmbedding} \coloneqq \bigl\|(a_i)_{i\geq 1}\bigr\|_{\ell_2(\N)} = \biggl(\sum_{i\geq 1} a_i^2\biggr)^{\sfrac{1}{2}}\;\;,
\]
for $(a_i)_{i\geq 1}\in\ell_2(\N)$, it becomes a Hilbert space. Moreover, $(\eigw_i^{\sfrac{\parEmbedding}{2}}[\eigv_i]_\nu)_{i\geq 1}$ forms an ONB of $[H]_\nu^\parEmbedding$ and consequently $[H]_\nu^\parEmbedding$ is a separable Hilbert space. If there is no danger of confusion we use the abbreviation $\|\cdot\|_\parEmbedding \coloneqq \|\cdot\|_{[H]_\nu^\parEmbedding}$. Furthermore, in the case of $\parEmbedding=1$ we introduce the notation $[H]_\nu \coloneqq [H]_\nu^1$. Recall that for $\parEmbedding=0$ we have $[H]_\nu^0 = \overline{\ran\Imas}\subseteq L_2(\nu)$ with $\|\cdot\|_0 = \|\cdot\|_{L_2(\nu)}$. Moreover, for $\parEmbedding = 1$ we have $[H]_\nu^1 = \ran\Imas$ and $[H]_\nu^1$ is isometrically isomorphic to the closed subspace $(\ker\Imas)^\perp$ of $H$ via $\Imas$, i.e.\ $\|[f]_\nu\|_{1} = \|f\|_H$ for $f\in(\ker\Imas)^\perp$. 
For $0<\parSourceCond<\parEmbedding$, the embeddings 
\begin{equation}\label{eq:pre:embeddings}
[H]_\nu^\parEmbedding \hookrightarrow[H]_\nu^\parSourceCond \hookrightarrow [H]_\nu^0 = \overline{\ran\Imas}\subseteq L_2(\nu)
\end{equation}
exist and they are compact. For $\parEmbedding>0$, the $\parEmbedding$-power space is given by the image of the fractional integral operator, namely 
\[ 
[H]_\nu^\parEmbedding=\ran\Tmas^{\sfrac{\parEmbedding}{2}} 
\qquad\text{and}\qquad
\|\Tmas^{\sfrac{\parEmbedding}{2}}f\|_\parEmbedding = \|f\|_{L_2(\nu)}
\]
for $f\in \overline{\ran\Imas}$. In addition, for $0<\parEmbedding<1$, the $\parEmbedding$-power space is characterized in terms of interpolation spaces of the real method, see e.g.\ \cites[Section~1.3.2]{Tr1978} for a definition. To be more precise, \citet[Theorem~4.6]{StSc2012} proved
\begin{equation}\label{eq:pre:interpolation_spaces}
[H]_\nu^\parEmbedding \cong \bigl[L_2(\nu),[H]_\nu\bigr]_{\parEmbedding,2}\;\;,
\end{equation}
where the symbol $\cong$ in \eqref{eq:pre:interpolation_spaces} means that these spaces are isomorphic, i.e.\ the sets coincide and the corresponding norms are equivalent. 
Note that for Sobolev/Besov RKHSs and marginal distributions that are essentially the uniform distribution, the interpolation space $\bigl[L_2(\nu),[H]_\nu\bigr]_{\parEmbedding,2}$ is well-known from the literature, see Section~\ref{sec:besov} for details. 


\section{Main Results}\label{sec:res}

Before we state the results we introduce the main assumptions. 
For $0<\parEigDecay\leq 1$ we assume that the \emph{eigenvalue decay} satisfies a polynomial upper bound of order $\sfrac{1}{\parEigDecay}$: There is a constant $\constEigDecay>0$ such that the eigenvalues $(\eigw_i)_{i\geq 1}$ of the integral operator satisfy 
\begin{gather}\label{eq:res:eigenvalue_decay}\tag{EVD}
\eigw_i \leq\constEigDecay\, i^{-\sfrac{1}{\parEigDecay}}
\end{gather}
for all $i\geq 1$. In order to establish the optimality of our results we need to assume an exact polynomial asymptotic behavior of order $\sfrac{1}{\parEigDecay}$: There are constants $\constEigDecayLB,\constEigDecay>0$ such that 
\begin{gather}\label{eq:res:eigenvalue_decay_exact}\tag{EVD+}
\constEigDecayLB\ i^{-\sfrac{1}{\parEigDecay}}\leq \eigw_i \leq\constEigDecay\, i^{-\sfrac{1}{\parEigDecay}}
\end{gather}
is satisfied for all $i\geq 1$. Our next assumption is the \emph{embedding property}, for $0<\parEmbedding\leq 1$: There is a constant $\constEmbedding>0$ with
\begin{gather}\label{eq:res:embedding_property}\tag{EMB}
\bigl\|[H]_\nu^\parEmbedding\hookrightarrow L_\infty(\nu)\bigr\| \leq\constEmbedding\;\;.
\end{gather}
This mean $[H]_\nu^\parEmbedding$ is continuously embedded into $L_\infty(\nu)$ and the operator norm of the embedding is bounded by $\constEmbedding$. Because of \eqref{eq:pre:embeddings} the larger $\parEmbedding$ is, the weaker the embedding property is. Since our kernel $k$ is bounded, \eqref{eq:res:embedding_property} is always satisfied for $\parEmbedding=1$. 
Moreover, Part~\ref{it:proof:embedding:eigenvalue_decay:iii} of Lemma~\ref{lem:proof:embedding:eigenvalue_decay} in Section~\ref{sec:proof} shows that \eqref{eq:res:embedding_property} implies a polynomial eigenvalue decay of order $\sfrac{1}{\parEmbedding}$ and hence we assume $\parEigDecay\leq\parEmbedding$ in the following. Observe that the converse does not hold in general and consequently it is possible that we even have the strict inequality $\parEigDecay < \parEmbedding$.

Note that the Conditions~\eqref{eq:res:embedding_property} and \eqref{eq:res:eigenvalue_decay}/\eqref{eq:res:eigenvalue_decay_exact} just describe the interplay between the marginal distribution $\nu=P_X$ and the RKHS $H$. Consequently, they are independent of the conditional distribution $P(\,\cdot\,|x)$ and especially independent of the regression function $\optFP$. In the following, we use a \emph{source condition}, for $0<\parSourceCond\leq 2$, to measure the smoothness of the regression function: There is a constant $\constSourceCond>0$ such that $\optFP\in [H]_\nu^\parSourceCond$ and 
\begin{gather}\label{eq:res:source_condition}\tag{SRC}
\|\optFP\|_\parSourceCond\leq\constSourceCond\;\;.
\end{gather}
Note that $|P|_2<\infty$, defined in \eqref{eq:pre:second_moment}, already implies $\optFP\in L_2(\nu)$. Moreover, \eqref{eq:res:source_condition} with $\parSourceCond\geq 1$ implies that $\optFP$ has a representative from $H$---in short $\optFP\in H$---and hence $\parSourceCond\geq 1$ excludes the hard learning scenario we are mainly interested in. Nonetheless, we included the case $1\leq\parSourceCond\leq 2$ because it is no extra effort in the proof. Since we want to estimate $\| [\optRegFD]_\nu - \optFP\|_\gamma$ and this expression is well-defined if and only if $\optFP\in[H]_\nu^\parPowerNorm$, we naturally have to assume $\parSourceCond\geq\parPowerNorm$ in the following. Finally, we introduce a \emph{moment condition} to control the noise of the observations: There are constants $\varb,\supb>0$ such that
\begin{gather}\label{eq:res:moment_condition}\tag{MOM}
\int_\R |y - \optFP(x)|^m\ P(\d y|x) \leq \frac{1}{2}m!\,\varb^2\,\supb^{m-2} 
\end{gather}
is satisfied for $\nu$-almost all $x\in X$ and all $m\geq 2$. Note that \eqref{eq:res:moment_condition} is satisfied for Gaussian noise with bounded variance, i.e.\ $P(\,\cdot\,|x) = \mathcal{N}(\optFP(x),\varb_x^2)$, where $x\mapsto\varb_x \in (0,\infty)$ is a measurable and $\nu$-almost surely bounded function. Another sufficient condition is that $P$ is concentrated on $X\times[-M,M]$ for some constant $M>0$, i.e.\ $P(X\times[-M,M]) = 1$.

The Conditions~\eqref{eq:res:eigenvalue_decay} and \eqref{eq:res:source_condition} are well-recognized in the statistical analysis of regularized least-squares algorithms \citep[see e.g.,][]{CaDe2007,BlM2017,LiCe2018,LiRuRoCe2018}. However, there is a whole zoo of moment conditions. We use \eqref{eq:res:moment_condition} because \eqref{eq:res:moment_condition} only constraints the discrepancy of the observation $y$ to the \emph{true} value $\optFP(x)$ and hence does \emph{not} imply additional constraints, such as boundedness, on $\optFP$. An embedding property slightly weaker than \eqref{eq:res:embedding_property} was used by \citet{StHuSc2009} in combination with empirical process arguments. \citet{DiFoHs2017} used \eqref{eq:res:embedding_property} to investigate benign scenarios with exponentially decreasing eigenvalues and $\optFP\in H$, and \citet{PiRuBa2018a} used \eqref{eq:res:embedding_property} to investigate stochastic gradient methods. But embedding properties are new in  combination with the integral operator technique in the hard learning scenario for the learning scheme \eqref{eq:intro:optimization_problem} and enable us to prove the following result.

\begin{thm}[\boldmath$\parPowerNorm$-Learning Rates]\label{thm:res:upper_rates}
Let $(X,\mathcal{B})$ be a measurable space, $H$ be a separable RKHS on $X$ with respect to a bounded and measurable kernel $k$, $P$ be a probability distribution on $X\times \R$ with $|P|_2<\infty$, and $\nu \coloneqq P_X$ be the marginal distribution on $X$. 
Furthermore, let $\constInftyBound>0$ be a constant with $\|\optFP\|_{L_\infty(\nu)}\leq\constInftyBound$ and the Conditions~\eqref{eq:res:embedding_property}, \eqref{eq:res:eigenvalue_decay} \eqref{eq:res:source_condition}, and \eqref{eq:res:moment_condition} be satisfied for some 
$0<\parEigDecay\leq\parEmbedding\leq 1$ and $0<\parSourceCond\leq 2$. 
Then, for $0\leq\parPowerNorm\leq 1$ with $\parPowerNorm<\parSourceCond$ and a regularization parameter sequence $(\lambda_n)_{n\geq 1}$, the LS-SVM $D\mapsto\optRegFD[\lambda_n]$ with respect to $H$ defined by \eqref{eq:intro:optimization_problem} satisfies the following statements:
\begin{enumerate}
\item\label{it:res:upper_rates:i} In the case of $\parSourceCond + \parEigDecay \leq \parEmbedding$ and $\lambda_n\asymp(\sfrac{n}{\log^r(n)})^{-\sfrac{1}{\parEmbedding }}$ for some $r>1$ there is a constant $K>0$ independent of $n\geq 1$ and $\tau\geq 1$ such that
\begin{equation}\label{eq:res:upper_rate:i}
\bigl\|[\optRegFD[\lambda_n]]_\nu - \optFP\bigr\|_{\parPowerNorm}^2 
\leq \tau^2 K \biggl(\frac{\log^r(n)}{n}\biggr)^{\frac{\parSourceCond-\parPowerNorm}{\parEmbedding}}
\end{equation}
is satisfied for sufficiently large $n\geq 1$ with $P^n$-probability not less than $1 - 4 e^{-\tau}$.
\item\label{it:res:upper_rates:ii} In the case of $\parSourceCond + \parEigDecay>\parEmbedding$ and $\lambda_n\asymp n^{-\sfrac{1}{(\parSourceCond + \parEigDecay)}}$ there is a constant $K>0$ independent of $n\geq 1$ and $\tau\geq 1$ such that
\begin{equation}\label{eq:res:upper_rate:ii}
\bigl\|[\optRegFD[\lambda_n]]_\nu - \optFP\bigr\|_{\parPowerNorm}^2 
\leq \tau^2 K \biggl(\frac{1}{n}\biggr)^{\frac{\parSourceCond-\parPowerNorm}{\parSourceCond + \parEigDecay}}
\end{equation}
is satisfied for sufficiently large $n\geq 1$ with $P^n$-probability not less than $1 - 4 e^{-\tau}$.
\end{enumerate}
\end{thm}

Theorem~\ref{thm:res:upper_rates} is mainly based on a finite sample bound given in Section~\ref{sec:proof}, see Theorem~\ref{thm:proof:upper:oi}. We think that the statement of Theorem~\ref{thm:res:upper_rates} can be proved for general regularization methods if one combines our technique, especially Lemma~\ref{lem:proof:upper:oi_part_i} and Lemma~\ref{lem:proof:upper:oi_part_ii} from Section~\ref{sec:upper}, with the results of \citet{LiRuRoCe2018} and \citet{LiCe2018}. However, we stick to the learning scheme \eqref{eq:intro:optimization_problem} for simplicity. The proof of Theorem~\ref{thm:res:upper_rates} reveals that the constants $K>0$ just depend on the parameters and constants from \eqref{eq:res:embedding_property}, \eqref{eq:res:eigenvalue_decay}, \eqref{eq:res:source_condition}, and \eqref{eq:res:moment_condition}, on the considered norm, i.e.\ on $\parPowerNorm$, on $\constInftyBound$, and on the regularization parameter sequence $(\lambda_n)_{n\geq 1}$. Moreover, the index bound hidden in the phrase \emph{for sufficient large $n\geq 1$} just depends on the parameters and constants from \eqref{eq:res:embedding_property} and \eqref{eq:res:eigenvalue_decay}, on $\tau$, on a lower bound $0<c\leq 1$ for the operator norm $c\leq\|\rTmas\|$, and on the regularization parameter sequence $(\lambda_n)_{n\geq 1}$. The asymptotic behavior in $n$ of the right hand side in \eqref{eq:res:upper_rate:i} and \eqref{eq:res:upper_rate:ii}, respectively, is called \emph{learning rate} with respect to the $\parPowerNorm$-power norm or abbreviated $\parPowerNorm$-learning rate. Recall, for $\parPowerNorm=0$, the norms on left hand sides of \eqref{eq:res:upper_rate:i} and \eqref{eq:res:upper_rate:ii} coincide with the $L_2(\nu)$-norm. 

Note that, for $\parSourceCond\geq\parEmbedding$, the conditional mean function $\optFP$ is automatically $\nu$-almost surely bounded, since we have $\optFP\in[H]_\nu^\parSourceCond\hookrightarrow[H]_\nu^\parEmbedding\hookrightarrow L_\infty(\nu)$, and in this case always Situation~\eqref{eq:res:upper_rate:ii} applies. Moreover, in the case of $\parEmbedding=\parEigDecay$, which was also considered by \citet[Corollary~6]{StHuSc2009}, we are always in Situation~\eqref{eq:res:upper_rate:ii}, too. 

If we ignore the $\log$-term in the obtained $\parPowerNorm$-learning rates then in both cases, $\parSourceCond+\parEigDecay\leq\parEmbedding$ and $\parSourceCond+\parEigDecay>\parEmbedding$, the $\parPowerNorm$-learning rate coincides with
\[ 
n^{-\frac{\parSourceCond-\parPowerNorm}{\max\{\parSourceCond+ \parEigDecay,\parEmbedding\}}}\;\;.
\]
Finally, note that the asymptotic behavior of the regularization parameter sequence \emph{does not depend} on the considered $\parPowerNorm$-power norm. Consequently, we get convergence with respect to \emph{all} $\parPowerNorm$-power norms $0\leq\parPowerNorm<\parSourceCond$ \emph{simultaneously}. 
In order to investigate the optimality of our $\parPowerNorm$-learning rates the next theorem yields $\parPowerNorm$-lower rates. In doing so, we have to assume \eqref{eq:res:eigenvalue_decay_exact} to make sure that the eigenvalues do not decay faster than \eqref{eq:res:eigenvalue_decay} guarantees.

\begin{thm}[\boldmath$\parPowerNorm$-Lower Rates]\label{thm:res:lower_rate}
Let $(X,\mathcal{B})$ be a measurable space, $H$ be a separable RKHS on $X$ with respect to a bounded and measurable kernel $k$, and $\nu$ be a probability distribution on $X$ such that \eqref{eq:res:embedding_property} and \eqref{eq:res:eigenvalue_decay_exact} are satisfied for some $0<\parEigDecay\leq\parEmbedding\leq 1$. 
Then, for all parameters $0<\parSourceCond\leq 2$, $0\leq\parPowerNorm\leq 1$ with $\parPowerNorm<\parSourceCond$ and all constants $\varb,\supb, \constSourceCond,\constInftyBound>0$, there exist $K_0,K,r>0$ such that for all learning methods $D\mapsto f_D$, all $\tau>0$, and all sufficiently large $n\geq 1$ there is a distribution $P$ on $X\times \R$ with $P_X=\nu$ satisfying $\|\optFP\|_{L_\infty(\nu)}\leq\constInftyBound$, \eqref{eq:res:source_condition} with respect to $\parSourceCond,\constSourceCond$, \eqref{eq:res:moment_condition} with respect to $\varb,\supb$, and with $P^n$-probability not less than $1 - K_0\tau^{\sfrac{1}{r}}$
\begin{equation}\label{eq:res:lower_rate}
\bigl\|[f_{D}]_\nu - \optFP\bigr\|_{\parPowerNorm}^2 \geq \tau^2 K \biggl(\frac{1}{n}\biggr)^{\frac{\max\{\parEmbedding,\parSourceCond\} - \parPowerNorm}{\max\{\parEmbedding,\parSourceCond\} + \parEigDecay}}\;\;.
\end{equation}
\end{thm}

In short, Theorem~\ref{thm:res:lower_rate} states that there is no learning method satisfying a faster decaying $\parPowerNorm$-learning rate than
\[
n^{-\frac{\max\{\parEmbedding,\parSourceCond\} - \parPowerNorm}{\max\{\parEmbedding,\parSourceCond\} + \parEigDecay}}
\]
under the assumptions of Theorem~\ref{thm:res:upper_rates} and \eqref{eq:res:eigenvalue_decay_exact}. The asymptotic behavior in $n$ of the right hand side in \eqref{eq:res:lower_rate} is called \emph{(minimax) lower rate} with respect to the $\parPowerNorm$-power norm or abbreviated $\parPowerNorm$-lower rate. Theorem~\ref{thm:res:lower_rate} extends the lower bounds previously obtained by \citet{CaDe2007}, \citet{StHuSc2009}, and \citet{BlM2017}. To be more precise, \citet[Theorem~2]{CaDe2007} considered only the case $\optFP\in H$ and $\parPowerNorm=0$, \citet[Theorem~9]{StHuSc2009} considered only the case $\parSourceCond\geq\parEmbedding$ and $\parPowerNorm=0$, and \citet[Theorem~3.5]{BlM2017} restricted their considerations to $\optFP\in H$. In the case of $\parEmbedding\leq\parSourceCond$, which implies the boundedness of $\optFP$, the $\parPowerNorm$-learning rate of LS-SVMs stated in Theorem~\ref{thm:res:upper_rates} coincides with the $\parPowerNorm$-lower rate from Theorem~\ref{thm:res:lower_rate} and hence is optimal. The optimal rate in the case of $\parEmbedding>\parSourceCond$, which does \emph{not} imply the boundedness of $\optFP$, is, \emph{even for the $L_2$-norm}, an outstanding problem for several decades, which we cannot address, either.

\begin{rem}[Optimality and Boundedness]
Under the assumptions of Theorem~\ref{thm:res:lower_rate}, \emph{but without} requiring the uniform boundedness of $\optFP$ by some constant $\constInftyBound$, we can \emph{improve} the $\parPowerNorm$-lower rate of Theorem~\ref{thm:res:lower_rate}. More precisely, a straightforward modification of Lemma~\ref{lem:proof:lower:valid_strings} in Section~\ref{sec:proof} gives in the case of not uniformly bounded $\optFP$ the $\parPowerNorm$-lower rate
\begin{equation*}\label{eq:res:lower_rate_improved} 
n^{-\frac{\parSourceCond - \parPowerNorm}{\parSourceCond + \parEigDecay}}\;\;.
\end{equation*}
Moreover, if we would be able to prove the $\parPowerNorm$-learning rates of Theorem~\ref{thm:res:upper_rates} with a constant $K>0$ independent of $\|\optFP\|_{L_\infty(\nu)}$ then we would have optimality for our $\parPowerNorm$-learning rates in the case of $\parSourceCond > \parEmbedding-\parEigDecay$ instead of $\parSourceCond\geq\parEmbedding$.
\end{rem}


Because of \eqref{eq:res:embedding_property}, the next remark is a direct consequence of Theorem~\ref{thm:res:upper_rates} for $\parPowerNorm = \parEmbedding$.

\begin{rem}[\boldmath$L_\infty$-Learning Rates]\label{rem:res:Linfinity_rate}
Under the assumptions of Theorem~\ref{thm:res:upper_rates} in the case of $\parSourceCond>\parEmbedding$ the following statement is true. 
For all regularization parameter sequences $(\lambda_n)_{n\geq 1}$ with $\lambda_n\asymp n^{\sfrac{1}{(\parSourceCond + \parEigDecay)}}$ there is a constant $K>0$ independent of $n\geq 1$ and $\tau\geq 1$ such that the LS-SVM $D\mapsto\optRegFD[\lambda_n]$ with respect to $H$ defined by \eqref{eq:intro:optimization_problem} satisfies
\[ 
\bigl\|[\optRegFD[\lambda_n]]_\nu - \optFP\bigr\|_{L_\infty(\nu)}^2 
\leq \tau^2 K \biggl(\frac{1}{n}\biggr)^{\frac{\parSourceCond-\parEmbedding}{\parSourceCond + \parEigDecay}}
\]
for sufficiently large $n\geq 1$ with $P^n$-probability not less than $1 - 4 e^{-\tau}$.
\end{rem}

Note that all previous efforts to get $L_\infty$-learning rates for the learning scheme \eqref{eq:intro:optimization_problem} need to assume $\optFP\in H$. Consequently, Remark~\ref{rem:res:Linfinity_rate} establishes the very first $L_\infty$-learning rates in the hard learning scenario.


\section{Example: Besov RKHSs}\label{sec:besov} 

In this section we illustrate our main results in the case of Besov RKHSs. To this end, we assume that $X$ is a benign domain:
Let $X\subseteq\R^d$ be a non-empty, open, connected, and bounded set with a 
\begin{gather}\label{eq:besov:domain}\tag{DOM}
C_\infty\text{-boundary}
\end{gather}
and be equipped with the Lebesgue-Borel $\sigma$-algebra $\mathcal{B}$. Furthermore, $L_2(X) \coloneqq L_2(\lebesgue)$ denotes the corresponding $L_2$-space.

Let us briefly introduce Sobolev and Besov Hilbert spaces. For a more detailed introduction see e.g.\ \cites{AdFo2003}. For $m\in\N$ we denote the \emph{Sobolev space} of smoothness $m$ by $W_m(X) \coloneqq W_{m,2}(X)$, see e.g.\ \cites[Definition~3.2]{AdFo2003} for a definition.
For $\parBesovRKHS>0$ the \emph{Besov space} $B^{\parBesovRKHS}_{2,2}(X)$ is defined by means of the real interpolation method, namely $B^{\parBesovRKHS}_{2,2}(X) \coloneqq \bigl[L_2(X), W_{m}(X)\bigr]_{\sfrac{\parBesovRKHS}{m},2}$, where $m \coloneqq \min\{k\in\N:\ k>\parBesovRKHS\}$ see e.g.\ \cites[Section~7.30]{AdFo2003} for details. For $\parBesovRKHS=0$ we define $B^{0}_{2,2}(X) \coloneqq L_2(X)$.
It is well-known that the Besov spaces $B^{\parBesovRKHS}_{2,2}(X)$ are separable Hilbert spaces and that they satisfy
\begin{equation}\label{eq:besov:reiteration}
B^{\parBesovRKHS}_{2,2}(X) \cong \bigl[L_2(X),B^\parBesovNorm_{2,2}(X)\bigr]_{\sfrac{\parBesovRKHS}{\parBesovNorm},2}
\end{equation}
for all $\parBesovNorm>\parBesovRKHS>0$, see e.g.\ \cites[Section~7.32]{AdFo2003} for details. Moreover, an extension of the Sobolev embedding theorem to Besov spaces guarantees that, for $\parBesovRKHS>\sfrac{d}{2}$, each $\lebesgue$-equivalence class in $B^{\parBesovRKHS}_{2,2}(X)$ has a unique continuous and bounded representative, see e.g.\ \cites[Part~\eqnr{c} of Theorem~7.24]{AdFo2003}. In fact, for $\parBesovRKHS > j + \sfrac{d}{2}$, this representative is from the space $C_j(X)$ of $j$-times continuous differentiable and bounded functions with bounded derivatives. More precisely, the mapping of a $\lebesgue$-equivalence class to its (unique) continuous representative is linear and continuous, in short, for $\parBesovRKHS > j + \sfrac{d}{2}$, 
\begin{equation}\label{eq:besov:embedding}
B^{\parBesovRKHS}_{2,2}(X) \hookrightarrow C_j(X)\;\;.
\end{equation}
Consequently, we define, for $\parBesovRKHS>\sfrac{d}{2}$, the \emph{Besov RKHS} as the set of continuous representatives $H_\parBesovRKHS(X) \coloneqq \{f\in C_0(X):\ [f]_\lebesgue\in B^{\parBesovRKHS}_{2,2}(X)\}$ and equip this space with the norm $\|f\|_{H_\parBesovRKHS(X)} \coloneqq \|[f]_\lebesgue\|_{B^{\parBesovRKHS}_{2,2}(X)}$. The Besov RKHS $H_\parBesovRKHS(X)$ is a separable RKHS with respect to a kernel $k_\parBesovRKHS$. Moreover, $k_\parBesovRKHS$ is bounded and measurable, see e.g.\ \cites[Lemma~4.28 and Lemma~4.25]{StCh2008}.

In the following, we fix a Besov RKHS $H_\parBesovRKHS(X)$ for some $\parBesovRKHS>\sfrac{d}{2}$ and a probability measure $P$ on $X\times \R$ such that the marginal distribution $\nu = P_X$ on $X$ satisfies the following condition: The probability measure $\nu$ is equivalent to the Lebesgue measure $\lebesgue$ on $X$, i.e.\ $\lebesgue\ll\nu$, $\nu\ll\lebesgue$, and there are constants $\constDensityLB,\constDensityUB>0$ such that
\begin{gather}\label{eq:besov:marginal_distribution}\tag{LEB}
\constDensityLB \leq \frac{\d\nu}{\d\lebesgue}\leq \constDensityUB
\end{gather}
is $\lebesgue$-almost surely satisfied. For marginal distributions $\nu$ satisfying \eqref{eq:besov:marginal_distribution} we have $L_2(\nu)\cong L_2(X)$ and we can describe the power spaces of $H_\parBesovRKHS(X)$ according to \eqref{eq:pre:interpolation_spaces}, the interpolation property, and \eqref{eq:besov:reiteration} by 
\begin{equation}\label{eq:besov:power_spaces}
[H_\parBesovRKHS(X)]_\nu^{\sfrac{u}{\parBesovRKHS}} 
\cong \bigl[L_2(\nu),[H_\parBesovRKHS(X)]_\nu\bigr]_{\sfrac{u}{\parBesovRKHS},2} 
\cong \bigl[L_2(X),[H_\parBesovRKHS(X)]_\lebesgue\bigr]_{\sfrac{u}{\parBesovRKHS},2} 
\cong B^{u}_{2,2}(X)
\end{equation}
for $0<u<\parBesovRKHS$.
As a consequence of \eqref{eq:besov:power_spaces}, we have $\optFP\in B^{\parBesovSourceCond}_{2,2}(X)$ for some $0<\parBesovSourceCond<\parBesovRKHS$ if and only if \eqref{eq:res:source_condition} is satisfied for $\parSourceCond=\sfrac{\parBesovSourceCond}{\parBesovRKHS}$. Next, if we combine \eqref{eq:besov:power_spaces} and \eqref{eq:besov:embedding} then we get \eqref{eq:res:embedding_property} for all $\parEmbedding$ with $\frac{d}{2\parBesovRKHS}<\parEmbedding<1$: 
\[
[H_\parBesovRKHS(X)]_\nu^\parEmbedding 
\cong B^{\parEmbedding \parBesovRKHS}_{2,2}(X) 
\hookrightarrow C_0(X)
\hookrightarrow L_\infty(\nu)\;\;.
\] 
Finally, we consider the asymptotic behavior of the eigenvalues $(\eigw_i)_{i\geq 1}$ of the integral operator $\Tmas$. \citet[Equation~\eqnr{4.4.12}]{CaSt1990} show that the eigenvalue $\eigw_i$ of $\Tmas$ equals the squares of the approximation number $a_i^2(\Imas)$ of the embedding $\Imas:H_\parBesovRKHS(X)\to L_2(\nu)$. Since $L_2(\nu)\cong L_2(X)$ these approximation numbers are described by \citet[Equation~\eqnr{4} on p.~119]{EdTr1996}, namely
\[
\eigw_i=a_i^2(\Imas) \asymp i^{-\sfrac{2\parBesovRKHS}{d}}\;\;.
\]
To sum up, the eigenvalues satisfy \eqref{eq:res:eigenvalue_decay_exact} for $\parEigDecay = \frac{d}{2\parBesovRKHS}$. 
The following corollaries are direct consequences of Part~\ref{it:res:upper_rates:ii} of Theorem~\ref{thm:res:upper_rates} and Theorem~\ref{thm:res:lower_rate} with $\parEigDecay=\frac{d}{2\parBesovRKHS}$, $\parSourceCond=\sfrac{\parBesovSourceCond}{\parBesovRKHS}$, $\parPowerNorm = \sfrac{\parBesovNorm}{\parBesovRKHS}$, and an $\parEmbedding>\parEigDecay$ that is chosen sufficiently close to $\parEigDecay$.

\begin{cor}[Besov-Learning Rates]\label{cor:besov:upper_rates}
Let $X\subseteq\R^d$ be a set satisfying \eqref{eq:besov:domain}, $H_r(X)$ be a Besov RKHS on $X$ with $\parBesovRKHS>\sfrac{d}{2}$, $P$ be a probability distribution on $X\times \R$ with $|P|_2<\infty$, and $\nu \coloneqq P_X$ be the marginal distribution on $X$ such that \eqref{eq:besov:marginal_distribution} is satisfied. 
Furthermore, let $\constSourceCond,\constInftyBound>0$ be constants with $\|\optFP\|_{L_\infty(\lebesgue)}\leq\constInftyBound$ and $\|\optFP\|_{B^{\parBesovSourceCond}_{2,2}(X)}\leq\constSourceCond$ for some $0<\parBesovSourceCond < \parBesovRKHS$, and the Condition~\eqref{eq:res:moment_condition} be satisfied. 
Then, for $0\leq\parBesovNorm<\parBesovSourceCond$ and a regularization parameter sequence $(\lambda_n)_{n\geq 1}$ with $\lambda_n\asymp n^{-\sfrac{\parBesovRKHS}{(\parBesovSourceCond+\sfrac{d}{2})}}$, there is a constant $K>0$ independent of $n\geq 1$ and $\tau\geq 1$ such that the LS-SVM $D\mapsto\optRegFD[\lambda_n]$ with respect to the Besov RKHS $H_\parBesovRKHS(X)$ defined by \eqref{eq:intro:optimization_problem} satisfies
\[ 
\bigl\|[\optRegFD[\lambda_n]]_\lebesgue - \optFP\bigr\|_{B^{\parBesovNorm}_{2,2}(X)}^2 \leq \tau^2 K \biggl(\frac{1}{n}\biggr)^{\frac{\parBesovSourceCond - \parBesovNorm}{\parBesovSourceCond + \sfrac{d}{2}}}
\]
for sufficiently large $n\geq 1$ with $P^n$-probability not less than $1- 4 e^{-\tau}$.
\end{cor}

Note that the $B^{\parBesovNorm}_{2,2}$-learning rate is independent of the chosen Besov RKHS $H_\parBesovRKHS(X)$. Besides $\parBesovRKHS>\sfrac{d}{2}$ the only requirement on the choice of $H_\parBesovRKHS(X)$, a user has to take care of, is $\parBesovRKHS>\parBesovSourceCond$, i.e.\ to pick a sufficiently small $H_\parBesovRKHS(X)$. Recall that the case $\parBesovNorm=0$ corresponds to $L_2$-norm learning rates.

\begin{cor}[Besov-Lower Rates]\label{cor:besov:lower_rates}
Let $X\subseteq\R^d$ be a set satisfying \eqref{eq:besov:domain}, $H_r(X)$ be a Besov RKHS on $X$ with $\parBesovRKHS>\sfrac{d}{2}$, and $\nu$ be a probability distribution on $X$ satisfying \eqref{eq:besov:marginal_distribution}. 
Then, for all parameters $0\leq\parBesovNorm<\parBesovSourceCond<\parBesovRKHS$ with $s>\sfrac{d}{2}$ and all constants $\varb,\supb,\constSourceCond,\constInftyBound>0$, there exist $K_0,K,r>0$ such that for all learning methods $D\mapsto f_D$, all $\tau>0$, and all sufficiently large $n\geq 1$ there is a distribution $P$ on $X\times \R$ with $P_X=\nu$ satisfying $\|\optFP\|_{L_\infty(\nu)}\leq\constInftyBound$, $\|\optFP\|_{B^{\parBesovNorm}_{2,2}(X)}\leq\constSourceCond$, \eqref{eq:res:moment_condition} with respect to $\varb,\supb$, and with $P^n$-probability not less than $1 - K_0\tau^{\sfrac{1}{r}}$
\[
\bigl\|[f_D]_\lebesgue - \optFP\bigr\|_{B^{\parBesovNorm}_{2,2}(X)}^2 \geq \tau^2 K \biggl(\frac{1}{n}\biggr)^{\frac{\parBesovSourceCond - \parBesovNorm}{\parBesovSourceCond + \sfrac{d}{2}}}\;\;.
\]
\end{cor}

In short, Corollary~\ref{cor:besov:lower_rates} states that the rates from Corollary~\ref{cor:besov:upper_rates} are optimal for $\parBesovSourceCond>\sfrac{d}{2}$. 

\begin{rem}\label{rem:besov:lower_rate}
Under the assumptions of Corollary~\ref{cor:besov:lower_rates} in the case of $\parBesovSourceCond\leq \sfrac{d}{2}$ for all sufficiently small $\varepsilon>0$ the following lower bound is satisfied
\[
\bigl\|[f_D]_\lebesgue - \optFP\bigr\|_{B^{\parBesovNorm}_{2,2}(X)}^2 \geq \tau^2 K \biggl(\frac{1}{n}\biggr)^{\sfrac{1}{2} - \sfrac{\parBesovNorm}{d} + \varepsilon}\;\;.
\]
\end{rem}

Finally, if we have $\parBesovSourceCond > j + \sfrac{d}{2}$, for some integer $j\geq 0$, then the combination of Corollary~\ref{cor:besov:upper_rates} and \eqref{eq:besov:embedding} yields $C_j(X)$-norm learning rates. To this end, we denote by $\optFP$ the unique continuous representative of the $\nu$-equivalence class $\optFP$ and apply Corollary~\ref{cor:besov:upper_rates} with a sufficiently small $\parBesovNorm>j+\sfrac{d}{2}$.

\begin{rem}[\boldmath$C_j(X)$-Learning Rates]\label{rem:besov:Cj_rate}
Under the assumption of Corollary~\ref{cor:besov:upper_rates} in the case of $\parBesovSourceCond > j + \sfrac{d}{2}$ for some integer $j\geq 0$ the following statement is true. 
For all $0<\varepsilon<\frac{\parBesovSourceCond - (j + \sfrac{d}{2})}{\parBesovSourceCond + \sfrac{d}{2}}$ and each regularization parameter sequence $(\lambda_n)_{n\geq 1}$ with $\lambda_n\asymp n^{-\sfrac{\parBesovRKHS}{(\parBesovSourceCond + \sfrac{d}{2}})}$ there is a constant $K>0$ independent of $n\geq 1$ and $\tau\geq 1$ such that the LS-SVM $D\mapsto\optRegFD[\lambda_n]$ with respect to the Besov RKHS $H_\parBesovRKHS(X)$ defined by \eqref{eq:intro:optimization_problem} satisfies
\[ 
\bigl\|\optRegFD[\lambda_n] - \optFP\bigr\|_{C_j(X)}^2 \leq \tau^2 K \biggl(\frac{1}{n}\biggr)^{\frac{\parBesovSourceCond - (j + \sfrac{d}{2})}{\parBesovSourceCond + \sfrac{d}{2}} - \varepsilon}
\]
for sufficiently large $n\geq 1$ with $P^n$-probability not less than $1- 4 e^{-\tau}$.
\end{rem}

Remark~\ref{rem:besov:Cj_rate} suggests that $D\mapsto \partial^\alpha\optRegFD$, for some multi-index $\alpha=(\alpha_1,\ldots,\alpha_d)\in\N_0^d$, is a reasonable estimator for the $\alpha$-th derivative of the regression function $\partial^\alpha\optFP$ if $\optFP\in B^{\parBesovSourceCond}_{2,2}(X)$ with some $s>|\alpha|+\sfrac{d}{2}=\alpha_1+\ldots+\alpha_d +\sfrac{d}{2}$. 
Note that the $\varepsilon>0$ appears in the rates of Remark~\ref{rem:besov:lower_rate} and Remark~\ref{rem:besov:Cj_rate} because we have to choose $\parEmbedding>\parEigDecay$ and $\parBesovNorm > j + \sfrac{d}{2}$, respectively. 


\section{Comparison}\label{sec:comparison} 

In this section we compare our results with learning rates previously obtained in the literature. Since in the case of $\optFP\in[H]_\nu^\parSourceCond$ with $1\leq\parSourceCond\leq 2$ we just recover the well-known optimal rates 
obtained by many authors, see e.g.\ \cite{CaDe2007, LiCe2018} for $L_2$-rates and \cite{BlM2017, LiRuRoCe2018} for general $\parPowerNorm$-rates, we focus on the hard learning scenario $0<\parSourceCond<1$. Furthermore, due to the large amount of results in the literature we limit our considerations to the best known results for the learning scheme \eqref{eq:intro:optimization_problem}, namely \cite{StCh2008,StHuSc2009}, which use empirical process techniques and \cite{LiCe2018,LiRuRoCe2018}, which use integral operator techniques. Moreover, we assume that $P$ is concentrated on $X\times[-M,M]$ for some $M>0$ and that $k$ is a bounded measurable kernel with separable RKHS $H$. Note that these assumptions form the largest common ground under which all the considered contributions achieve $L_2$-learning rates. In addition, the article of \citet{LiRuRoCe2018} is the only one of the four articles listed above that considers general $\parPowerNorm$-learning rates. Finally, in order to keep the comparison clear we ignore $\log$-terms in the learning rates. In Table~\ref{tab:comparison:rates} we give a short overview of the learning rates and in Figure~\ref{fig:comparison:L2rates} we plot the exponent $r$ of the polynomial $L_2$-learning rates $n^{-r}$ over the smoothness $0<\parSourceCond<1$ of $\optFP\in[H]_\nu^\parSourceCond$ for some fixed $0<\parEigDecay\leq\parEmbedding\leq 1$.


\begin{table}[t]
\centering
\renewcommand{\arraystretch}{1.5}
%
\definecolor{opt}{RGB}{0,0,0}
\definecolor{bad}{RGB}{255,0,0}
\definecolor{emp}{RGB}{17,166,0}
\definecolor{int}{RGB}{3,94,167}
\newcommand{\missing}{x}
\newcommand{\centercell}[1]{\begin{tabular}{c}#1\end{tabular}}
\newcolumntype{P}[1]{>{\centering\arraybackslash}p{#1}}
\newcolumntype{M}[1]{>{\centering\arraybackslash}m{#1}}
\resizebox{0.95\textwidth}{!}{
\begin{tabular}{P{5.3cm}|c|c|c|c}
\emph{Articles} & 
\multicolumn{2}{c|}{\emph{Assumptions}} & 
\multicolumn{2}{c}{\emph{Exponent $r$ of the}}\\[-0.5ex]
%
%
&
\eqref{eq:res:embedding_property} &
\eqref{eq:res:eigenvalue_decay} &
\multicolumn{2}{c}{\emph{Learning Rate $n^{-r}$}}\\[-0.5ex]
& 
{\small $[H]_\nu^\parEmbedding\hookrightarrow L_\infty(\nu)$} & 
{\small $\eigw_i\preccurlyeq i^{-\frac{1}{p}}$ }& 
$L_2(\nu)$&
$[H]_\nu^\parPowerNorm$ for $\parPowerNorm<\parSourceCond$\\
%
\hline\hline
%
\textcolor{int}{our results} & 
\multirow{2}{*}[-1.5ex]{\textcolor{opt}{$0<\parEmbedding\leq 1$}} &
\multirow{2}{*}[-1.5ex]{\textcolor{opt}{$0<\parEigDecay\leq \parEmbedding$}} &
\multirow{2}{*}[-1.5ex]{\textcolor{opt}{$\frac{\parSourceCond}{\max\{\parSourceCond + \parEigDecay,\parEmbedding\}}$}} &
\textcolor{opt}{$\frac{\parSourceCond - \parPowerNorm}{\max\{\parSourceCond + \parEigDecay,\parEmbedding\}}$}\\
%
\cline{1-1}\cline{5-5}
%
\textcolor{emp}{\citet[Thm.~7.23]{StCh2008} + \eqref{eq:res:embedding_property}} &
& 
& 
& 
\multirow{5}{*}[-3ex]{\missing}\\
%
\cline{1-4}
%
\textcolor{emp}{\citet[Thm.~1]{StHuSc2009}}&
\textcolor{opt}{$0<\parEmbedding\leq 1$} &
\textcolor{opt}{$0<\parEigDecay \leq \parEmbedding$} & 
\textcolor{bad}{$\frac{\parSourceCond}{\max\{\parSourceCond +\parEigDecay,\parSourceCond +\parEmbedding(1-\parSourceCond)\}}$} &
\\ 
%
\cline{1-4}
%
\textcolor{emp}{\citet[Cor.~6]{StHuSc2009}}&
\textcolor{opt}{$0<\parEmbedding\leq 1$} &
\textcolor{bad}{$\parEigDecay = \parEmbedding$} & 
\textcolor{bad}{$\frac{\parSourceCond}{\parSourceCond + \parEmbedding}$} &
\\ 
%
\cline{1-4}
%
\textcolor{emp}{\citet[Eq.~(7.54)]{StCh2008}}&
\multirow{3}{*}[-2ex]{\textcolor{bad}{$\parEmbedding = 1$}} &
\multirow{3}{*}[-2ex]{\textcolor{opt}{$0<\parEigDecay \leq 1$}} & 
\multirow{3}{*}[-2ex]{\textcolor{bad}{$\frac{\parSourceCond}{\max\{\parSourceCond + \parEigDecay, 1\}}$}}&
\\ 
%
\cline{1-1}
%
\textcolor{int}{\citet[Cor.~6]{LiCe2018}}&
& 
& 
& 
\\ 
%
\cline{1-1}\cline{5-5}
%
\textcolor{int}{\citet[Cor.~4.4]{LiRuRoCe2018}}&
& 
& 
& 
\textcolor{bad}{$\frac{\parSourceCond-\parPowerNorm}{\max\{\parSourceCond + \parEigDecay, 1\}}$}\\
%
\end{tabular}
} 
\caption{Learning rates established by different authors for $\optFP\in[H]_\nu^\parSourceCond$ with $0<\parSourceCond<1$. In order to keep the comparison clear we ignore $\log$-terms in the learning rates. The \textcolor{int}{\emph{blue}} results are based on integral operator techniques and the \textcolor{emp}{\emph{green}} ones are based on empirical process techniques. The \textcolor{bad}{\emph{marked}} parameter ranges are more restrictive than ours and the \textcolor{bad}{\emph{marked}} rates are never better than our rates and at least for some parameter ranges worse than our rates.}\label{tab:comparison:rates}
\end{table}

\begin{figure}[t]
\centering
\newlength{\origtabcolsep}
\setlength\origtabcolsep{\tabcolsep}
\setlength\tabcolsep{0pt}
\input{Color.tex}
\newcommand{\myline}[1]{\raisebox{0.4ex}{\tikz \draw[#1, line width= 1.2pt, line cap=round] (0pt,0pt) -- (20pt,0pt);}}
\newcommand{\mydashedline}[1]{\raisebox{0.4ex}{\tikz \draw[#1, line width= 1.2pt, line cap=round, dash pattern=on 4pt off 4pt] (0pt,0pt) -- (20pt,0pt);}}

\begin{tabular}{cc} 
\begin{tabular}{c} 
\resizebox{0.4\textwidth}{!}{
\begin{tikzpicture}[x=1pt,y=1pt]
\path[use as bounding box,fill=transparent,fill opacity=0.00] (0,0) rectangle (144.54,144.54);
\begin{scope}
\path[clip] (  0.00,  0.00) rectangle (144.54,144.54);

\path[draw=black,line width= 0.4pt,line join=round,line cap=round] ( 22.62, 26.23) -- (136.38, 26.23);

\path[draw=black,line width= 0.4pt,line join=round,line cap=round] ( 22.62, 26.23) -- ( 22.62, 20.23);

\path[draw=black,line width= 0.4pt,line join=round,line cap=round] ( 51.06, 26.23) -- ( 51.06, 20.23);

\path[draw=black,line width= 0.4pt,line join=round,line cap=round] ( 79.50, 26.23) -- ( 79.50, 20.23);

\path[draw=black,line width= 0.4pt,line join=round,line cap=round] (136.38, 26.23) -- (136.38, 20.23);

\node[text=black,anchor=base,inner sep=0pt, outer sep=0pt, scale=  1.00] at ( 22.62,  4.63) {0};

\node[text=black,anchor=base,inner sep=0pt, outer sep=0pt, scale=  1.00] at ( 51.06,  4.63) {$p$};

\node[text=black,anchor=base,inner sep=0pt, outer sep=0pt, scale=  1.00] at ( 79.50,  4.63) {$\alpha$};

\node[text=black,anchor=base,inner sep=0pt, outer sep=0pt, scale=  1.00] at (136.38,  4.63) {1};

\path[draw=black,line width= 0.4pt,line join=round,line cap=round] ( 22.62, 26.23) -- ( 22.62,139.99);

\path[draw=black,line width= 0.4pt,line join=round,line cap=round] ( 22.62, 26.23) -- ( 16.62, 26.23);

\path[draw=black,line width= 0.4pt,line join=round,line cap=round] ( 22.62, 83.11) -- ( 16.62, 83.11);

\path[draw=black,line width= 0.4pt,line join=round,line cap=round] ( 22.62,139.99) -- ( 16.62,139.99);

\node[text=black,rotate= 90.00,anchor=base,inner sep=0pt, outer sep=0pt, scale=  1.00] at (  8.22, 26.23) {$0$};

\node[text=black,rotate= 90.00,anchor=base,inner sep=0pt, outer sep=0pt, scale=  1.00] at (  8.22, 83.11) {$1/2$};

\node[text=black,rotate= 90.00,anchor=base,inner sep=0pt, outer sep=0pt, scale=  1.00] at (  8.22,139.99) {$1$};
\end{scope}
\begin{scope}
\path[clip] ( 14.45, 21.68) rectangle (144.54,144.54);

\path[fill=99999977,fill opacity=0.47] ( 79.50, 26.23) --
	(136.38, 26.23) --
	(136.38,139.99) --
	( 79.50,139.99) --
	cycle;

\path[draw=black,line width= 0.4pt,dash pattern=on 4pt off 4pt ,line join=round,line cap=round] ( 51.06, 26.23) --
	( 51.06,139.99);

\path[draw=black,line width= 0.4pt,dash pattern=on 4pt off 4pt ,line join=round,line cap=round] ( 79.50, 26.23) --
	( 79.50,139.99);

\path[draw=black,line width= 0.8pt,line join=round,line cap=round] ( 22.62, 26.23) --
	( 24.99, 30.97) --
	( 27.36, 35.71) --
	( 29.73, 40.45) --
	( 32.10, 45.19) --
	( 34.47, 49.93) --
	( 36.84, 54.67) --
	( 39.21, 59.41) --
	( 41.58, 64.15) --
	( 43.95, 68.89) --
	( 46.32, 73.63) --
	( 48.69, 78.37) --
	( 51.06, 83.11) --
	( 53.43, 85.39) --
	( 55.80, 87.49) --
	( 58.17, 89.43) --
	( 60.54, 91.24) --
	( 62.91, 92.92) --
	( 65.28, 94.49) --
	( 67.65, 95.95) --
	( 70.02, 97.33) --
	( 72.39, 98.62) --
	( 74.76, 99.84) --
	( 77.13,100.99) --
	( 79.50,102.07) --
	( 81.87,103.10) --
	( 84.24,104.07) --
	( 86.61,104.99) --
	( 88.98,105.86) --
	( 91.35,106.69) --
	( 93.72,107.49) --
	( 96.09,108.24) --
	( 98.46,108.96) --
	(100.83,109.65) --
	(103.20,110.31) --
	(105.57,110.94) --
	(107.94,111.55) --
	(110.31,112.13) --
	(112.68,112.69) --
	(115.05,113.22) --
	(117.42,113.74) --
	(119.79,114.23) --
	(122.16,114.71) --
	(124.53,115.17) --
	(126.90,115.61) --
	(129.27,116.04) --
	(131.64,116.45) --
	(134.01,116.85) --
	(136.38,117.24);

\path[draw=red,line width= 0.8pt,dash pattern=on 4pt off 4pt ,line join=round,line cap=round] ( 22.62, 26.23) --
	( 24.99, 30.87) --
	( 27.36, 35.33) --
	( 29.73, 39.61) --
	( 32.10, 43.73) --
	( 34.47, 47.70) --
	( 36.84, 51.51) --
	( 39.21, 55.19) --
	( 41.58, 58.73) --
	( 43.95, 62.16) --
	( 46.32, 65.46) --
	( 48.69, 68.65) --
	( 51.06, 71.73) --
	( 53.43, 74.72) --
	( 55.80, 77.61) --
	( 58.17, 80.40) --
	( 60.54, 83.11) --
	( 62.91, 85.74) --
	( 65.28, 88.28) --
	( 67.65, 90.75) --
	( 70.02, 93.15) --
	( 72.39, 95.48) --
	( 74.76, 97.74) --
	( 77.13, 99.93) --
	( 79.50,102.07) --
	( 81.87,103.10) --
	( 84.24,104.07) --
	( 86.61,104.99) --
	( 88.98,105.86) --
	( 91.35,106.69) --
	( 93.72,107.49) --
	( 96.09,108.24) --
	( 98.46,108.96) --
	(100.83,109.65) --
	(103.20,110.31) --
	(105.57,110.94) --
	(107.94,111.55) --
	(110.31,112.13) --
	(112.68,112.69) --
	(115.05,113.22) --
	(117.42,113.74) --
	(119.79,114.23) --
	(122.16,114.71) --
	(124.53,115.17) --
	(126.90,115.61) --
	(129.27,116.04) --
	(131.64,116.45) --
	(134.01,116.85) --
	(136.38,117.24);

\path[draw=orange,line width= 0.8pt,dash pattern=on 4pt off 4pt ,line join=round,line cap=round] ( 22.62, 26.23) --
	( 24.99, 30.78) --
	( 27.36, 34.98) --
	( 29.73, 38.87) --
	( 32.10, 42.48) --
	( 34.47, 45.84) --
	( 36.84, 48.98) --
	( 39.21, 51.92) --
	( 41.58, 54.67) --
	( 43.95, 57.26) --
	( 46.32, 59.69) --
	( 48.69, 61.98) --
	( 51.06, 64.15) --
	( 53.43, 66.20) --
	( 55.80, 68.14) --
	( 58.17, 69.98) --
	( 60.54, 71.73) --
	( 62.91, 73.40) --
	( 65.28, 74.98) --
	( 67.65, 76.50) --
	( 70.02, 77.94) --
	( 72.39, 79.32) --
	( 74.76, 80.64) --
	( 77.13, 81.90) --
	( 79.50, 83.11) --
	( 81.87, 84.27) --
	( 84.24, 85.39) --
	( 86.61, 86.46) --
	( 88.98, 87.49) --
	( 91.35, 88.48) --
	( 93.72, 89.43) --
	( 96.09, 90.35) --
	( 98.46, 91.24) --
	(100.83, 92.09) --
	(103.20, 92.92) --
	(105.57, 93.72) --
	(107.94, 94.49) --
	(110.31, 95.23) --
	(112.68, 95.95) --
	(115.05, 96.65) --
	(117.42, 97.33) --
	(119.79, 97.99) --
	(122.16, 98.62) --
	(124.53, 99.24) --
	(126.90, 99.84) --
	(129.27,100.42) --
	(131.64,100.99) --
	(134.01,101.54) --
	(136.38,102.07);

\path[draw=violet,line width= 0.8pt,dash pattern=on 4pt off 4pt ,line join=round,line cap=round] ( 22.62, 26.23) --
	( 24.99, 28.60) --
	( 27.36, 30.97) --
	( 29.73, 33.34) --
	( 32.10, 35.71) --
	( 34.47, 38.08) --
	( 36.84, 40.45) --
	( 39.21, 42.82) --
	( 41.58, 45.19) --
	( 43.95, 47.56) --
	( 46.32, 49.93) --
	( 48.69, 52.30) --
	( 51.06, 54.67) --
	( 53.43, 57.04) --
	( 55.80, 59.41) --
	( 58.17, 61.78) --
	( 60.54, 64.15) --
	( 62.91, 66.52) --
	( 65.28, 68.89) --
	( 67.65, 71.26) --
	( 70.02, 73.63) --
	( 72.39, 76.00) --
	( 74.76, 78.37) --
	( 77.13, 80.74) --
	( 79.50, 83.11) --
	( 81.87, 85.48) --
	( 84.24, 87.85) --
	( 86.61, 90.22) --
	( 88.98, 92.59) --
	( 91.35, 94.96) --
	( 93.72, 97.33) --
	( 96.09, 99.70) --
	( 98.46,102.07) --
	(100.83,104.44) --
	(103.20,106.81) --
	(105.57,109.18) --
	(107.94,111.55) --
	(110.31,112.13) --
	(112.68,112.69) --
	(115.05,113.22) --
	(117.42,113.74) --
	(119.79,114.23) --
	(122.16,114.71) --
	(124.53,115.17) --
	(126.90,115.61) --
	(129.27,116.04) --
	(131.64,116.45) --
	(134.01,116.85) --
	(136.38,117.24);
\end{scope}
\end{tikzpicture}\nolinebreak
} 
\end{tabular} & 
\setlength\tabcolsep{\origtabcolsep}
\begin{tabular}[c]{l>{\raggedright\arraybackslash}p{6.3cm}} 
\multirow{3}{*}{\myline{black}} 
& our result and \\
& \citeauthor{StCh2008}\\
& \citeidpar[Thm.~7.23]{StCh2008} + \eqref{eq:res:embedding_property}.\\[2ex]
\mydashedline{red} & \citet[Thm.~1]{StHuSc2009}\\[2ex]
\mydashedline{orange} & \citet[Cor.~6]{StHuSc2009}\\[2ex]
\multirow{4}{*}{\mydashedline{violet}} 
& \citeauthor{StCh2008}\\
& \citeidpar[Eq.~(7.54)]{StCh2008},\\
& \citet[Cor.~6]{LiCe2018}, and\\
& \citet[Cor.~4.4]{LiRuRoCe2018}.\\
\end{tabular} 
\end{tabular} 
\setlength\tabcolsep{\origtabcolsep}
\caption{Plot of the exponent $r$ of the $L_2$-learning rate $n^{-r}$ over the smoothness $\parSourceCond$ of $\optFP$ for a fixed RKHS $H$ and a fixed marginal distribution $\nu=P_X$ which satisfy \eqref{eq:res:embedding_property} and \eqref{eq:res:eigenvalue_decay} with respect to $\parEmbedding=\sfrac{1}{2}$ and $\parEigDecay=\sfrac{1}{4}$, respectively. Consequently, higher values correspond to faster learning rates. In the gray shaded range the best rates are know to be optimal.}
\label{fig:comparison:L2rates}
\end{figure}

\textbf{Integral operator techniques.} The article of \citet{LiCe2018} is an extended version of the conference paper \cite{LiCe2018a}. \citet{LiCe2018} investigate distributed gradient decent methods and spectral regularization algorithms. In Corollary~6 they provide the $L_2$-learning rate $n^{-\sfrac{\parSourceCond}{\max\{\parSourceCond+\parEigDecay,1\}}}$ in expectation for spectral regularization algorithms, containing the learning scheme \eqref{eq:intro:optimization_problem} as special case.
\citet{LiRuRoCe2018} establish the $\parPowerNorm$-learning rate $n^{-\sfrac{(\parSourceCond-\parPowerNorm)}{\max\{\parSourceCond+\parEigDecay,1\}}}$ in probability for spectral regularization algorithms under more general source conditions, see \cites[Equation~\eqnr{18}]{LiRuRoCe2018} for a definition. 
Both articles do not take any embedding property into account and hence we get at least the same rates and in case of \eqref{eq:res:embedding_property} with $\parEmbedding<1$ we actually improve their rates iff $\parSourceCond + \parEigDecay < 1$.
Let us illustrate this improvement in the case of a Besov RKHS $H_\parBesovRKHS(X)$ with smoothness $r$. To this end, we assume $\optFP\in B^{\parBesovSourceCond}_{2,2}(X)$ for some $\parBesovSourceCond>0$. Besides the condition $\parBesovRKHS>\sfrac{d}{2}$, which ensures that $H_\parBesovRKHS(X)$ is a RKHS, the only requirement for our Corollary~\ref{cor:besov:upper_rates} is $\parBesovRKHS>\parBesovSourceCond$ in order to achieve the fastest known $L_2$-learning rate $n^{-\sfrac{\parBesovSourceCond}{(\parBesovSourceCond+\sfrac{d}{2})}}$. Recall that this rate is independent of the smoothness $\parBesovRKHS$ of the hypothesis space and is known to be optimal for $\parBesovSourceCond>\sfrac{d}{2}$, see e.g.\ Corollary~\ref{cor:besov:lower_rates}. In order to get the same $L_2$-learning rate by the results of \citet{LiCe2018} or \citet{LiRuRoCe2018} the \emph{additional} constraint $\parBesovRKHS\leq\parBesovSourceCond+\sfrac{d}{2}$ has to be satisfied. Otherwise, they only yield the $L_2$-rate $n^{-\sfrac{\parBesovSourceCond}{\parBesovRKHS}}$, which gets worse with increasing smoothness $\parBesovRKHS$. Consequently, taking \eqref{eq:res:embedding_property} into account facilitates the choice of $\parBesovRKHS$. Moreover, for learning rates with respect to Besov norms our results improve those of \citet{LiRuRoCe2018} in a similar way, i.e.\ to get our Besov-learning rates with the help of the results of \citet{LiRuRoCe2018} the \emph{additional} constraint $\parBesovRKHS\leq\parBesovSourceCond+\sfrac{d}{2}$ has to be satisfied.

\textbf{Empirical process techniques.} \citet{StCh2008} provide an oracle inequality in Theorem~7.23 under a slightly weaker assumption than \eqref{eq:res:eigenvalue_decay}. As already mentioned there \citep[Equation~\eqnr{7.54}]{StCh2008}, this oracle inequality leads, under a slightly weaker assumption than \eqref{eq:res:source_condition}, to the $L_2$-rate $n^{-\sfrac{\parSourceCond}{\max\{\parSourceCond+\parEigDecay,1\}}}$. This rate coincides with the results of \citet{LiCe2018} and \citet{LiRuRoCe2018}, and is even better by a logarithmic factor. Inspired by \citet[Lemma~5.1]{MeNe2010}, \citet{StHuSc2009} were the first using an embedding property, slightly weaker than \eqref{eq:res:embedding_property}, to derive finite sample bounds, see \cites[Theorem~1]{StHuSc2009}. Moreover, Theorem~1 of \citet{StHuSc2009} was used in Corollary~6 of that article to establish, in the case of $\parEigDecay=\parEmbedding$, the $L_2$-rate $n^{-\sfrac{\parSourceCond}{(\parSourceCond+\parEmbedding)}}$. But the proof remains valid in the general case $\parEigDecay\leq \parEmbedding$ and hence \citet[Theorem~1]{StHuSc2009} get the $L_2$-rate $n^{-\sfrac{\parSourceCond}{\max\{\parSourceCond +\parEigDecay,\parSourceCond +\parEmbedding(1-\parSourceCond)\}}}$. This rate is never better than ours and is worse than ours iff $\parEmbedding<1$ and $\parSourceCond<1-\sfrac{\parEigDecay}{\parEmbedding}$. If we combine the oracle inequality of \citet[Theorem~7.23]{StCh2008} with \eqref{eq:res:embedding_property} then we recover our $L_2$-rate from Theorem~\ref{thm:res:upper_rates} even without logarithmic factor.
However, recall that the empirical process technique is not able to provide general $\parPowerNorm$-learning rates yet. 
Finally, it is to mention that both contributions, \cite{StCh2008} and \cite{StHuSc2009}, consider the \emph{clipped} predictor. The influence of this clipping is not clear, but it could be the reason for avoiding the logarithmic factors appearing in some learning rates obtained by integral operator techniques.

To sum up, we use the integral operator technique to recover the best known, and in many cases optimal, $L_2$-learning rates previously only obtained by the empirical process technique. In addition, we improve the best known $\parPowerNorm$-learning rates from \cite{LiRuRoCe2018} for the learning scheme \eqref{eq:intro:optimization_problem} whenever \eqref{eq:res:embedding_property} is satisfied for some $0<\parEmbedding<1$ as well as  \eqref{eq:res:source_condition} and \eqref{eq:res:eigenvalue_decay} are satisfied for $\parSourceCond+\parEigDecay < 1$. Finally, we show that our $\parPowerNorm$-learning rates are optimal in all cases in which the optimal $L_2$-norm learning rate is known.


\section{Proofs}\label{sec:proof}

First, we summarize some well-known facts that we need for the proofs of our main results. To this end, we use the notation and general assumptions from Section~\ref{sec:pre}.



Since we assume that $H$ is separable, \citet[Corollary~3.2]{StSc2012} show that there exists a $\nu$-zero set $N\subseteq X$, such that $k$ is given by
\begin{equation}\label{eq:proof:embedding:mercer}
k(x,x') = \sum_{i\geq 1}\eigw_i\,\eigv_i(x)\eigv_i(x')
\end{equation}
for all $x,x'\in X\backslash N$. Furthermore, the boundedness of $k$ implies $\sum_{i\geq 1}\eigw_i\eigv_i^2(x)\leq\constEmbedding^2$ for $\nu$-almost all $x\in X$ and a constant $\constEmbedding\geq 0$. Motivated by this statement we say, for $\parEmbedding> 0$, that the \emph{$\parEmbedding$-power of $k$ is $\nu$-a.s.\ bounded} if there exists a constant $\constEmbedding\geq 0$ with
\begin{equation}\label{eq:proof:embedding:bounded_power}
\sum_{i\geq 1}\eigw_i^\parEmbedding \eigv_i^2(x) \leq \constEmbedding^2
\end{equation}
for $\nu$-almost all $x\in X$. Furthermore, we write $\|k_\nu^\parEmbedding\|_{\infty}$ for the smallest constant with this property and set $\|k_\nu^\parEmbedding\|_{\infty} \coloneqq \infty$ if there is no such constant. Consequently,  $\|k_\nu^\parEmbedding\|_{\infty}<\infty$ is an abbreviation of the phrase \emph{the $\parEmbedding$-power of $k$ is $\nu$-a.s.\ bounded}. We refer to \cites[Proposition~4.2]{StSc2012} for the logic behind this notation. Because of the representation in \eqref{eq:proof:embedding:mercer} and the boundedness of $k$ we always have $\|k_\nu^1\|_{\infty}<\infty$. The following theorem allows an alternative characterization of \eqref{eq:res:embedding_property}.

\begin{thm}[\boldmath$L_\infty$-Embedding]\label{thm:proof:embedding:equivalence}
Let $(X,\mathcal{B})$ be a measurable space, $H$ be a separable RKHS on $X$ w.r.t.\ a bounded and measurable kernel $k$, and $\nu$ be a probability distribution on $X$. Then the following equality is satisfied, for $\parEmbedding>0$,
\begin{equation}\label{eq:proof:embedding:norm-identity}
\bigl\|[H]_\nu^\parEmbedding\hookrightarrow L_\infty(\nu)\bigr\| 
= \|k_\nu^\parEmbedding\|_{\infty}\;\;.
\end{equation}
\end{thm}

Note that with the help of \eqref{eq:proof:embedding:norm-identity} the Condition~\eqref{eq:res:embedding_property} can be written as $\|k_\nu^\parEmbedding\|_\infty \leq \constEmbedding$. The statement of Theorem~\ref{thm:proof:embedding:equivalence} is part of \cites[Theorem~5.3]{StSc2012}, but we give an alternative proof below, which is more basic and does \emph{not} require the $\nu$-completeness of the measurable space $(X,\mathcal{B})$. Moreover, our proof of Theorem~\ref{thm:proof:embedding:equivalence} remains true in the situation considered by \citet[Theorem~5.3]{StSc2012}, i.e.\ for $\sigma$-finite measures $\nu$ and (possibly unbounded) kernels $k$ whose RKHS $H$ is compactly embedded into $L_2(\nu)$. In this respect, we generalize Theorem~5.3 of \cites{StSc2012}. We restricted our consideration to bounded kernels and probability measures only for convenience since we do not need this generality in the rest of this work.

\begin{proof}
First we prove `$\geq$'. To this end, we assume that $\Id:[H]_\nu^\parEmbedding \to L_\infty(\nu)$ exists and is bounded and hence $\|\Id\|<\infty$. Since $(\mu_i^{\parEmbedding/2}[e_i]_\nu)_{i\in I}$ is an ONB of $[H]_\nu^\parEmbedding$ for every sequence $a=(a_i)_{i\geq 1}\in\ell_2(\N)$ the series $\sum_{i\geq 1} a_i \mu_i^{\sfrac{\parEmbedding}{2}}[e_i]_\nu$ converges in $[H]_\nu^\parEmbedding$ and hence it also converges in $L_\infty(\nu)$. As a result, there is a representative $f_a:X\to\R$ with $[f_a]_\nu = \sum_{i\geq 1}a_i \mu_i^{\sfrac{\parEmbedding}{2}}[e_i]_\nu\in[H]_\nu^\parEmbedding$ and a set $N_a\subseteq X$ with $\nu(N_a)=0$ such that
\begin{equation*}
f_a(x) = \sum_{i\geq 1} a_i \mu_i^{\sfrac{\parEmbedding}{2}}e_i(x)
\qquad\text{and}\qquad
|f_a(x)| 
\leq \Bigl\|\sum_{i\geq 1}a_i \mu_i^{\sfrac{\parEmbedding}{2}}[e_i]_\nu\Bigr\|_{L_\infty(\nu)}
\end{equation*}
for all $x\in X\backslash N_a$. Consequently, for all $x\in X\backslash N_a$, we find
\begin{equation*}
|f_a(x)| 
\leq \Bigl\|\sum_{i\geq 1}a_i \mu_i^{\sfrac{\parEmbedding}{2}}[e_i]_\nu\Bigr\|_{L_\infty(\nu)}
\leq \|\Id\|\cdot \Bigl\|\sum_{i\geq 1}a_i \mu_i^{\sfrac{\parEmbedding}{2}}[e_i]_\nu\Bigr\|_{[H]_\nu^\parEmbedding}
= \|\Id\|\cdot\|a\|_{\ell_2(\N)}
\;\;.
\end{equation*}
Since the closed unit ball $\cuball{\ell_2(\N)}$ of $\ell_2(\N)$ is separable there is a countable dense subset $B\subseteq\cuball{\ell_2(\N)}$. If we define the set $N:=\bigcup_{a\in B}N_a\subseteq X$ then we have $\nu(N)=0$ since $B$ is countable. Moreover, the denseness of $B$ in $\cuball{\ell_2(\N)}$ implies
\begin{equation*}
\sum_{i\geq 1}\mu_i^\parEmbedding e_i^2(x) 
= \|(\mu_i^{\sfrac{\parEmbedding}{2}}e_i(x))\|_{\ell_2(\N)}^2
= \sup_{a\in B} |\langle a, (\mu_i^{\sfrac{\parEmbedding}{2}}e_i(x))_{i\geq 1}\rangle_{\ell_2(\N)}|^2
= \sup_{a\in B} |f_a(x)|^2
\leq \|\Id\|^2
\end{equation*}
for all $x\in X\backslash N$ and hence $\|k_\nu^\parEmbedding\|_\infty\leq \|\Id\|$.

Now we prove `$\leq$'. To this end, we assume $\|k_\nu^\parEmbedding\|_\infty < \infty$ and choose some $[f]_\nu \in [H]_\nu^\parEmbedding$ with $\|[f]_\nu\|_\parEmbedding \leq 1$. Then there is a (unique) sequence $a=(a_i)_{i\geq 1}\in\ell_2(\N)$ with $\|a\|_{\ell_2(\N)}\leq 1$ and $[f]_\nu = \sum_{i\geq 1} a_i\eigw_i^{\sfrac{\parEmbedding}{2}}[\eigv_i]_\nu$. Using H\"older's inequality we get
\begin{equation*}
|f(x)| 
\leq \|a\|_{\ell_2(\N)} \biggl(\sum_{i\geq 1}\eigw_i^\parEmbedding\eigv_i^2(x)\biggr)^{\sfrac{1}{2}} 
\leq \|k_\nu^\parEmbedding\|_\infty
\end{equation*}
for $\nu$-almost all $x\in X$. Consequently, we have $\|[f]_\nu\|_{L_\infty(\nu)}\leq \|k_\nu^\parEmbedding\|_\infty$ for all $[f]_\nu\in[H]_\nu^\parEmbedding$ with $\|[f]_\nu\|_\parEmbedding\leq 1$ and this proves $\|\Id\|\leq \|k_\nu^\parEmbedding\|_\infty$.
\end{proof}

The following lemma summarizes further implications of \eqref{eq:res:embedding_property}.

\begin{lem}\label{lem:proof:embedding:eigenvalue_decay}
Let $(X,\mathcal{B})$ be a measurable space, $H$ be a separable RKHS on $X$ w.r.t.\ a bounded and measurable kernel $k$, and $\nu$ be a probability distribution on $X$. Then the following statements are true, for $0<\parEigDecay,\parEmbedding\leq 1$:
\begin{enumerate}
\item\label{it:proof:embedding:eigenvalue_decay:i} \eqref{eq:res:embedding_property} implies $\|[\eigv_i]_\nu\|_{L_\infty(\nu)} \leq \|k_\nu^\parEmbedding\|_\infty \,\eigw_i^{-\sfrac{\parEmbedding}{2}}$ for all $i\geq 1$.
\item\label{it:proof:embedding:eigenvalue_decay:ii} \eqref{eq:res:embedding_property} implies $(\eigw_i)_{i\geq 1}\in\ell_\parEmbedding(\N)$. If, in addition, the eigenfunctions are uniformly bounded, i.e.\ $\sup_{i\geq 1}\|[\eigv_i]_\nu\|_{L_\infty(\nu)}<\infty$, then the converse implication is true.
\item\label{it:proof:embedding:eigenvalue_decay:iii} \eqref{eq:res:embedding_property} implies \eqref{eq:res:eigenvalue_decay} for $\parEigDecay=\parEmbedding$. If, in addition, the eigenfunctions are uniformly bounded, then \eqref{eq:res:eigenvalue_decay} w.r.t.\ $0<\parEigDecay<1$ implies \eqref{eq:res:embedding_property} for all $\parEmbedding>\parEigDecay$.
\end{enumerate}
\end{lem}

Note that uniformly bounded eigenfunction have been considered e.g.\ by \cites[Assumption~4.1]{MeNe2010} and \cites[Theorem~2]{StHuSc2009}, see also the discussion after Theorem~5.3 of \cites{StSc2012}.

\begin{proof}
For the proof we silently use the Identity~\eqref{eq:proof:embedding:norm-identity} in Theorem~\ref{thm:proof:embedding:equivalence}.

\ref{it:proof:embedding:eigenvalue_decay:i} Using \eqref{eq:res:embedding_property} and the fact that $(\eigw_i^{\sfrac{\parEmbedding}{2}}[\eigv_i]_\nu)_{i\geq 1}$ is an ONB of $[H]_\nu^\parEmbedding$ yields the assertion 
\begin{equation*}
\|\eigw_i^{\sfrac{\parEmbedding}{2}}[\eigv_i]_\nu\|_{L_\infty(\nu)}
\leq \|k_\nu^\parEmbedding\|_\infty \bigl\|\eigw_i^{\sfrac{\parEmbedding}{2}}[\eigv_i]_\nu\bigr\|_\parEmbedding 
= \|k_\nu^\parEmbedding\|_\infty\;\;.
\end{equation*}

\ref{it:proof:embedding:eigenvalue_decay:ii} The first statement in \ref{it:proof:embedding:eigenvalue_decay:ii} is from \cites[Theorem~5.3]{StSc2012}. The converse under the additional assumption of uniformly bounded eigenfunctions is a direct consequence of \eqref{eq:proof:embedding:bounded_power}. 

\ref{it:proof:embedding:eigenvalue_decay:iii} If \eqref{eq:res:embedding_property} is satisfied for $\alpha$, then the monotonicity of the eigenvalues $(\eigw_i)_{i\geq 1}$ and Statement~\ref{it:proof:embedding:eigenvalue_decay:ii} imply, for $i\geq 1$,
\[ 
i\eigw_i^\parEmbedding 
\leq \sum_{j=1}^i \eigw_j^\parEmbedding 
\leq \sum_{j\geq 1}\eigw_j^\parEmbedding 
=: D<\infty\;\;.
\]
Consequently, \eqref{eq:res:eigenvalue_decay} is satisfied for $\parEigDecay=\parEmbedding$ and $\constEigDecay \coloneqq D^{\sfrac{1}{\parEmbedding}}$. For the converse we assume \eqref{eq:res:eigenvalue_decay} w.r.t.\ $0<\parEigDecay< 1$. As a consequence, we have $\sum_{i\geq 1}\eigw_i^\parEmbedding \leq \constEigDecay^\parEmbedding\sum_{i\geq 1} i^{-\sfrac{\parEmbedding}{\parEigDecay}}<\infty$ for all $\parEmbedding>\parEigDecay$ and together with Part~\ref{it:proof:embedding:eigenvalue_decay:ii} this gives the assertion.
\end{proof}


Recall that the \emph{effective dimension} $\effdimMas:(0,\infty)\to [0,\infty)$ is defined by 
\[ 
\effdimMas(\lambda) \coloneqq \tr\bigl((\rTmas + \lambda)^{-1}\rTmas\bigr) = \sum_{i\geq 1}\frac{\eigw_i}{\eigw_i + \lambda}\;\;,
\]
where $\tr$ denotes the trace operator. The effective dimension is widely used in the statistical analysis of LS-SVMs \citep[see e.g.,][]{CaDe2007,BlM2017, LiCe2018, LiRuRoCe2018}. The following lemma establishes a connection between \eqref{eq:res:eigenvalue_decay} and the asymptotic behavior of $\effdimMas(\lambda)$ for $\lambda\to 0^+$. 

\begin{lem}\label{lem:proof:effective_dimension}
Let $(X,\mathcal{B})$ be a measurable space, $H$ be a separable RKHS on $X$ w.r.t.\ a bounded and measurable kernel $k$, and $\nu$ be a probability distribution on $X$. 
Then the following statements are equivalent, for $0< \parEigDecay\leq 1$:
\begin{enumerate}
\item\label{it:proof:effective_dimension:i} There is a constant $\constEffDim>0$ such that the following inequality is satisfied, for $\lambda>0$,
\begin{equation*}
\effdimMas(\lambda) \leq\constEffDim \lambda^{-\parEigDecay}\;\;.
\end{equation*}
\item\label{it:proof:effective_dimension:ii} \eqref{eq:res:eigenvalue_decay} is satisfied for $\parEigDecay$, i.e.\ there is a constant $\constEigDecay>0$ with $\eigw_i\leq \constEigDecay i^{-\sfrac{1}{\parEigDecay}}$ for all $i\geq 1$.
\end{enumerate}
\end{lem}

Note that \ref{it:proof:effective_dimension:i}$\Rightarrow$\ref{it:proof:effective_dimension:ii} for $\parEigDecay<1$ is from \citet[Proposition~3]{CaDe2007}.

\begin{proof}
\ref{it:proof:effective_dimension:i}$\Leftarrow$\ref{it:proof:effective_dimension:ii} For $\parEigDecay<1$ this implication is a consequence of \cites[Proposition~3]{CaDe2007} for $\constEffDim \coloneqq \sfrac{\constEigDecay^\parEigDecay}{(1 - \parEigDecay)}$. For $\parEigDecay=1$ the properties of the trace operator yields $\effdimMas(\lambda) \leq \|\rTmas\|_{1}\,\|(\rTmas + \lambda)^{-1}\|$. Since $\rTmas$ is a positive semi-definite operator we have $\|(\rTmas + \lambda)^{-1}\|\leq \lambda^{-1}$.
Moreover, using the ONS $([\eigv_i]_\nu)_{i\geq 1}$ in $L_2(\nu)$ and the monotone convergence theorem, the trace norm can be bounded by 
\begin{equation*}
\|\rTmas\|_1
= \sum_{i\geq 1}\eigw_i 
= \sum_{i\geq 1}\eigw_i \int_X \eigv_i^2(x)\ \d\nu(x)
= \int_X\sum_{i\geq 1}\eigw_i  \eigv_i^2(x)\ \d\nu(x)
\leq \|k_\nu^1\|_{\infty}^{2}
=:\constEffDim
\end{equation*}

\ref{it:proof:effective_dimension:i}$\Rightarrow$\ref{it:proof:effective_dimension:ii} Since $(\eigw_i)_{i\geq 1}$ is non-increasing also $(\sfrac{\eigw_i}{(\eigw_i +\lambda)})_{i\geq 1}$ is non-increasing for all $\lambda>0$. Consequently, we have, for $i\geq 1$ and $\lambda>0$,
\[ 
i \frac{\eigw_i}{\eigw_i + \lambda} 
\leq \sum_{j=1}^i \frac{\eigw_j}{\eigw_j + \lambda} 
\leq \effdimMas(\lambda)
\leq \constEffDim \lambda^{-\parEigDecay}\;\;.
\]
Using this inequality for $\lambda=\eigw_i$ we get $i \leq 2 D \eigw_i^{-\parEigDecay}$ for all $i\geq 1$ and this yields \eqref{eq:res:eigenvalue_decay} w.r.t.\ $\parEigDecay$ and $\constEigDecay=(2\constEffDim)^{\sfrac{1}{\parEigDecay}}$.
\end{proof}


The \emph{LS-risk} of a measurable function $f:X\to\R$ is defined by
\[ 
\mathcal{R}_{P}(f) \coloneqq \int_{X\times \R} \bigl(y - f(x)\bigr)^2\ \d P(x,y)
\]
and the \emph{Bayes-LS-risk} $\mathcal{R}_{P}^\ast \coloneqq \inf_{f:X\to\R}\mathcal{R}_{P}(f)$ is achieved by the conditional mean function $\optFP$, see e.g.\ \cites[Example~2.6]{StCh2008}. Moreover, the \emph{LS-excess-risk} is given by $\mathcal{R}_{P}(f) - \mathcal{R}_{P}^\ast = \|[f]_\nu - \optFP\|_{L_2(\nu)}^2$, see e.g.\ \cites[Example~2.6]{StCh2008}, and minimizing the LS-risk is therefore equivalent to approximating the conditional mean function in the $L_2(\nu)$-norm. For $\lambda>0$ the unique minimizer of
\begin{equation}\label{eq:proof:svm:problem}
\inf_{f\in H}\lambda\|f\|_H^2 + \mathcal{R}_{P}(f)
\end{equation}
can be easily calculated by means of derivatives and is given by
\begin{equation}\label{eq:proof:svm:solution} 
\optRegFP \coloneqq (\rTmas + \lambda)^{-1}\gP \in H \qquad \text{with }\ \gP \coloneqq \Smas\optFP\;\;,
\end{equation}
see e.g.\ \cites[Equations~\eqnr{7.4} and Equation~\eqnr{7.5}]{SmZh2005}. Note that \eqref{eq:pre:integral_operator} and \eqref{eq:pre:regression_function} together with the properties of the conditional distribution $P(\,\cdot\,|x)$ yield
\begin{equation}\label{eq:proof:svm:gP}
\gP = \int_X k(x,\,\cdot\,) \int_\R y\ P(\d y|x)\ \d P_X(x) = \int_{X\times \R} y k(x,\,\cdot\,)\ \d P(x,y)\;\;.
\end{equation}
The predictor $\optRegFD$, for a data set $D=\{(x_i,y_i)\}_{i=1}^n$, given in \eqref{eq:intro:optimization_problem} is the unique minimizer of \eqref{eq:proof:svm:problem} w.r.t.\ the \emph{empirical} measure $D \coloneqq \frac{1}{n}\sum_{i=1}^n \delta_{(x_i,y_i)}$ where $\delta_{(x,y)}$ denotes the Dirac measure at $(x,y)$. Consequently, $\optRegFD$ is given by \eqref{eq:proof:svm:solution} w.r.t.\ the corresponding empirical quantities, namely
\begin{equation}\label{eq:proof:svm:solution_empirical}
\optRegFD = (\rT_\delta + \lambda)^{-1} \gD \in H\;\;,
\end{equation}
where $\delta$ denotes the marginal distribution of $D$ on $X$, i.e.\ $\delta=D_X=\frac{1}{n}\sum_{i=1}^n\delta_{x_i}$.


\subsection{Some Bounds} \label{sec:bounds}

In this subsection we further exploit the spectral representations in \eqref{eq:pre:spectral} in order to establish some bounds which we use several times in the proofs of our main results.

Recall from \cites[Theorem~2.11 and Lemma~2.12]{StSc2012} that $(\eigw_i^{\sfrac{1}{2}}\eigv_i)_{i\geq 1}$ is an ONB of $(\ker\Imas)^\perp$, $([\eigv_i]_\nu)_{i\geq 1}$ is an ONB of $\overline{\ran\Imas}=[H]_\nu^0$, and 
\begin{equation}\label{eq:proof:bounds:operator:i}
\Smas = \sum_{i\geq 1}\eigw_i^{\sfrac{1}{2}} \langle [\eigv_i]_\nu,\,\cdot\,\rangle_{L_2(\nu)} \eigw_i^{\sfrac{1}{2}}\eigv_i\;\;.
\end{equation}
As the representation in \eqref{eq:proof:svm:solution} indicates, the operator $(\rTmas + \lambda)^{-a}$, for $a>0$, plays a crucial role in the following. To this end, we fix an arbitrary ONB $(\tilde{\eigv}_j)_{j\in J}$ of $\ker\Imas$, with $J\cap\N=\emptyset$, and bring up the following spectral representation
\begin{equation}\label{eq:proof:bounds:operator:ii}
(\rTmas + \lambda)^{-a}
= \sum_{i\geq 1} (\eigw_i + \lambda)^{-a} \langle \eigw_i^{\sfrac{1}{2}}\eigv_i,\,\cdot\,\rangle_H \:\eigw_i^{\sfrac{1}{2}}\eigv_i + \lambda^{-a}\sum_{j\in J}\langle \tilde{\eigv}_j, \,\cdot\,\rangle_H\:\tilde{\eigv}_j\;\;.
\end{equation}
Note that $(\tilde{\eigv}_j)_{j\in J}\subseteq H$ are normalized in contrast to $(\eigv_i)_{i\geq 1}\subseteq H$, which are not normalized to be aligned with the literature. Moreover, by normalizing $(\eigv_i)_{i\geq 1}$ we get the ONB $(\eigw_i^{\sfrac{1}{2}}\eigv_i)_{i\geq 1}\cup(\tilde{\eigv}_j)_{j\in J}$ of $H$, where $J$ is at most countably infinite since $H$ is separable.

Next, we present a spectral representation for $\optRegFP$ which is well-known from \cites[proof of Theorem~4]{SmZh2005}. To this end, we use the abbreviation $a_i \coloneqq \langle\optFP,[\eigv_i]_\nu\rangle_{L_2(\nu)}$, for $i\geq 1$. A combination of \eqref{eq:proof:svm:solution} with the representations in \eqref{eq:proof:bounds:operator:i} and \eqref{eq:proof:bounds:operator:ii}, for $a=1$, yields
\begin{equation}\label{eq:proof:bounds:svm}
\optRegFP = \sum_{i\geq 1}\frac{\eigw_i^{\sfrac{1}{2}}}{\eigw_i + \lambda}\,a_i\, \eigw_i^{\sfrac{1}{2}}\eigv_i \in (\ker\Imas)^\perp\;.
\end{equation}
If we additionally assume $\optFP\in\overline{\ran\Imas} = [H]_\nu^0$, then $\optFP = \sum_{i\geq 1} a_i [\eigv_i]_\nu$ holds and together with \eqref{eq:proof:bounds:svm} we have
\begin{equation}\label{eq:proof:bounds:svm_diff}
\optFP - [\optRegFP]_\nu = \sum_{i\geq 1}\frac{\lambda}{\eigw_i + \lambda}\,a_i\,[\eigv_i]_\nu\;\;.
\end{equation}
The first lemma describes the connection of the $\parPowerNorm$-power norm and the $H$-norm.

\begin{lem}\label{lem:proof:bounds:rkhs_norm}
Let $(X,\mathcal{B})$ be a measurable space, $H$ be a separable RKHS on $X$ w.r.t.\ a bounded and measurable kernel $k$, and $\nu$ be a probability distribution on $X$. 
Then, for $0\leq\parPowerNorm\leq 1$ and $f\in H$, the inequality 
\begin{equation*}
\|[f]_\nu\|_{\parPowerNorm}\leq \bigl\|\rTmas^{\frac{1-\parPowerNorm}{2}}f\bigr\|_H
\end{equation*}
is satisfied. If, in addition, $\parPowerNorm<1$ or $f\perp\ker\Imas$ is satisfied, then equality holds.
\end{lem}

\begin{proof}
Let us fix a $f\in H$. Since $(\eigw_i^{\sfrac{1}{2}}\eigv_i)_{i\geq 1}$ is an ONB of $(\ker\Imas)^\perp$, there exits a $g\in\ker\Imas$ with $f = \sum_{i\geq 1}b_i\,\eigw_i^{\sfrac{1}{2}}\eigv_i + g$, where $b_i = \langle f,\eigw_i^{\sfrac{1}{2}}\eigv_i\rangle_H$ for all $i\geq 1$. Since $[g]_\nu = 0$ we have $[f]_\nu= \sum_{i\geq 1}b_i\eigw_i^{\sfrac{1}{2}}[\eigv_i]_\nu$ and together with Parseval's identity w.r.t.\ the ONB $(\eigw_i^{\sfrac{\parPowerNorm}{2}}[\eigv_i]_\nu)_{i\geq 1}$ of $[H]_\nu^\parPowerNorm$ this yields
\[
\|[f]_\nu\|_{\parPowerNorm}^2 
= \Bigl\|\sum_{i\geq 1} b_i \eigw_i^{\frac{1-\parPowerNorm}{2}}\eigw_i^{\sfrac{\parPowerNorm}{2}}[\eigv_i]_\nu\Bigr\|_{\parPowerNorm}^2 
= \sum_{i\geq 1}\eigw_i^{1-\parPowerNorm}b_i^2\;\;.
\]
For $\parPowerNorm<1$ the spectral decomposition in \eqref{eq:pre:spectral} together with Parseval's identity w.r.t.\ the ONS $(\eigw_i^{\sfrac{1}{2}}\eigv_i)_{i\geq 1}$ in $H$ yields
\[
\|\rTmas^\frac{1-\parPowerNorm}{2}f\|_H^2 = \Bigl\|\sum_{i\geq 1} \eigw_i^\frac{1-\parPowerNorm}{2} b_i\, \eigw_i^{\sfrac{1}{2}}\eigv_i\Bigr\|_H^2 = \sum_{i\geq 1}\eigw_i^{1-\parPowerNorm}b_i^2\;\;.
\]
This proves the claimed equality in the case of $\parPowerNorm<1$. For $\parPowerNorm=1$ we have $\rTmas^\frac{1-\parPowerNorm}{2} = \Id_H$ and the Pythagorean theorem together with Parseval's identity 
yields
\[
\|\rTmas^\frac{1-\parPowerNorm}{2}f\|_H^2 
= \Bigl\|\sum_{i\geq 1} b_i\, \eigw_i^{\sfrac{1}{2}}\eigv_i + g\Bigr\|_H^2 
= \Bigl\|\sum_{i\geq 1} b_i\, \eigw_i^{\sfrac{1}{2}}\eigv_i\Bigr\|_H^2 + \|g\|_H^2
= \sum_{i\geq 1}b_i^2 + \|g\|_H^2\;\;.
\]
This gives the claimed equality if $f\perp\ker\Imas$, i.e.\ $g=0$, as well as the claimed inequality for general $f\in H$.
\end{proof}

The next lemma describes how the effective dimension comes into play. Note that parts of the next lemma are already mentioned by \citet{RuCaRo2015} in the discussion after Assumption~3.

\begin{lem}\label{lem:proof:bounds:help}
Let $(X,\mathcal{B})$ be a measurable space, $H$ be a separable RKHS on $X$ w.r.t.\ a bounded and measurable kernel $k$, and $\nu$ be a probability distribution on $X$.
Then the following equality is satisfied, for $\lambda>0$,
\begin{equation}\label{eq:proof:bounds:help:i}
\int_X \bigl\|(\rTmas + \lambda)^{-\sfrac{1}{2}} k(x,\,\cdot\,)\bigr\|_H^2\ \d\nu(x) = \effdimMas(\lambda)\;\;.
\end{equation}
If, in addition, $\|k_\nu^\parEmbedding\|_\infty < \infty$ is satisfied, then the following inequality is satisfied, for $\lambda>0$ and $\nu$-almost all $x\in X$,
\begin{equation}\label{eq:proof:bounds:help:ii}
\bigl\|(\rTmas + \lambda)^{-\sfrac{1}{2}} k(x,\,\cdot\,)\bigr\|_H^2 \leq \|k_\nu^\parEmbedding\|_{\infty}^2 \lambda^{-\parEmbedding}\;\;.
\end{equation}
\end{lem}

Note that the inequality in \eqref{eq:proof:bounds:help:ii} is the place where we benefit from \eqref{eq:res:embedding_property}.

\begin{proof}
Let us fix a $\lambda>0$. Since $H$ is separable and $k$ is measurable the map $X\to H$ given by $x\mapsto k(x,\,\cdot\,)$ is measurable, see e.g.\ \cites[Lemma~4.25]{StCh2008} and hence $x\mapsto\| (\rTmas + \lambda)^{-\sfrac{1}{2}} k(x,\,\cdot\,)\|_H^2$ is measurable, too. Using the ONB $(\tilde{\eigv}_j)_{j\in J}$ of $\ker\Imas$ introduced before Equation~\eqref{eq:proof:bounds:operator:ii} and the reproducibility property of the kernel $k$ we get the following series representation which converges in $H$
\begin{align*}
k(x,\,\cdot\,) 
&=\sum_{i\geq 1} \langle\eigw_i^{\sfrac{1}{2}}\eigv_i, k(x,\,\cdot\,)\rangle_H\, \eigw_i^{\sfrac{1}{2}}\eigv_i + \sum_{j\in J}\langle\tilde{\eigv}_j,k(x,\,\cdot\,)\rangle_H\,\tilde{\eigv}_j\\
&= \sum_{i\geq 1} \eigw_i^{\sfrac{1}{2}}\eigv_i(x)\, \eigw_i^{\sfrac{1}{2}}\eigv_i + \sum_{j\in J}\tilde{\eigv}_j(x)\,\tilde{\eigv}_j
\end{align*}
for all $x\in X$. Together with \eqref{eq:proof:bounds:operator:ii}, for $a=\sfrac{1}{2}$, and Parseval's identity we get
\begin{align*}
\|(\rTmas + \lambda)^{-\sfrac{1}{2}} k(x,\,\cdot\,)\|_H^2 
&= \biggl\|\sum_{i\geq 1} \frac{\eigw_i^{\sfrac{1}{2}}\eigv_i(x)}{(\eigw_i + \lambda)^{\sfrac{1}{2}}}\eigw_i^{\sfrac{1}{2}}\eigv_i + \lambda^{-\sfrac{1}{2}}\sum_{j\in J}\tilde{\eigv}_j(x)\, \tilde{\eigv}_j\biggr\|_H^2\\
&= \sum_{i\geq 1}\frac{\eigw_i}{\eigw_i + \lambda}\eigv_i^2(x) + \frac{1}{\lambda}\sum_{j\in J}\tilde{\eigv}_j^2(x)
\end{align*}
for all $x\in X$. Recall that the index set $J$ is at most countable since $H$ is separable. Moreover, $\tilde{\eigv}_j\in\ker\Imas$ for all $j\in J$ implies that the second summand on the right hand side vanishes for $\nu$-almost all $x\in X$. Consequently, we have
\begin{equation}\label{eq:proof:bounds:help:temp}
\|(\rTmas + \lambda)^{-\sfrac{1}{2}} k(x,\,\cdot\,)\|_H^2 
= \sum_{i\geq 1}\frac{\eigw_i}{\eigw_i + \lambda}\eigv_i^2(x)
\end{equation}
for $\nu$-almost all $x\in X$. Now, \eqref{eq:proof:bounds:help:i} is a consequence of \eqref{eq:proof:bounds:help:temp}, the monotone convergence theorem, and the fact that $([\eigv_i])_{i\geq 1}$ is an ONS in $L_2(\nu)$, namely
\[
\int_X \|(\rTmas + \lambda)^{-\sfrac{1}{2}} k(x,\,\cdot\,)\|_H^2\ \d\nu(x) = \sum_{i\geq 1} \frac{\eigw_i}{\eigw_i + \lambda}\int_X\eigv_i^2(x)\ \d\nu(x) = \tr\bigr((\rTmas + \lambda)^{-1}\rTmas\bigl)\;\;.
\]
Finally, \eqref{eq:proof:bounds:help:ii} is a consequence of \eqref{eq:proof:bounds:help:temp} and Lemma~\ref{lem:apx:estimate}, namely
\[
\|(\rTmas + \lambda)^{-\sfrac{1}{2}} k(x,\,\cdot\,)\|_H^2
= \sum_{i\geq 1}\frac{\eigw_i^{1-\parEmbedding}}{\eigw_i + \lambda}\,\eigw_i^\parEmbedding\eigv_i^2(x)
\leq \Bigl(\sum_{i\geq 1}\eigw_i^\parEmbedding\eigv_i^2(x)\Bigr)\ \sup_{i\geq 1}\frac{\eigw_i^{1-\parEmbedding}}{\eigw_i + \lambda} 
\leq \|k_\nu^\parEmbedding\|_{\infty}^2 \lambda^{-\parEmbedding}
\]
is satisfied for $\nu$-almost all $x\in X$.
\end{proof}

The next lemma uses the representations in \eqref{eq:proof:bounds:svm} and \eqref{eq:proof:bounds:svm_diff} to provide bounds on the $\parPowerNorm$-power norm of $[\optRegFP]_\nu - \optFP$ and $[\optRegFP]_\nu$.

\begin{lem}\label{lem:proof:bounds:power_norm}
Let $(X,\mathcal{B})$ be a measurable space, $H$ be a separable RKHS on $X$ w.r.t.\ a bounded and measurable kernel $k$, $P$ be a probability distribution on $X\times \R$ with $|P|_2<\infty$, and $\nu \coloneqq P_X$ be the marginal distribution on $X$. 
If $\optFP \in [H]_\nu^\parSourceCond$ is satisfied for some $0\leq \parSourceCond\leq 2$, then the following bounds are satisfied, for all $\lambda>0$:
\begin{align}
\label{it:proof:bounds:power_norm:svm_diff} \|[\optRegFP]_\nu - \optFP\|_{\parPowerNorm}^2 
&\leq \|\optFP\|_{\parSourceCond}^2\,\lambda^{\parSourceCond-\parPowerNorm} 
&\text{for all } 0\leq\parPowerNorm\leq\parSourceCond,\\
\label{it:proof:bounds:power_norm:svm} \|[\optRegFP]_\nu\|_\parPowerNorm^2 
&\leq \|\optFP\|_{\min\{\parPowerNorm,\parSourceCond\}}^2 \lambda^{-(\parPowerNorm-\parSourceCond)_+} 
&\text{for all }\parPowerNorm\geq 0.
\end{align}
\end{lem}

Here we used the abbreviation $t_+ \coloneqq \max\{0,t\}$ for $t\in\R$. Note that \eqref{it:proof:bounds:power_norm:svm_diff} in the case of $\parPowerNorm\in\{0,1\}$ is covered by \cites[Theorem~4]{SmZh2005}.
Since, in the case $\parSourceCond\geq\parPowerNorm=1$, the $\nu$-equivalence class $\optFP$ has a (unique) representative $\optFP\in H$ with $\optFP\perp\ker\Imas$ and $\optRegFP\perp\ker\Imas$ holds according to \eqref{eq:proof:bounds:svm}, we can use the equality from Lemma~\ref{lem:proof:bounds:rkhs_norm} and exchange the left hand sides of \eqref{it:proof:bounds:power_norm:svm_diff} by $\|\optRegFP - \optFP\|_H^2$ in the case of $\parSourceCond\geq\parPowerNorm=1$. Analogously, we can exchange the left hand side in \eqref{it:proof:bounds:power_norm:svm} by $\|\optRegFP\|_H^2$ for $\parPowerNorm=1$.

\begin{proof} 
Let us first show \eqref{it:proof:bounds:power_norm:svm_diff}. Since $\optFP\in[H]_\nu^\parSourceCond\subseteq[H]_\nu^0$ we can use the spectral representation in \eqref{eq:proof:bounds:svm_diff}. Then, Parseval's identity w.r.t.\ the ONB $(\eigw_i^{\sfrac{\parPowerNorm}{2}}[\eigv_i]_\nu)_{i\geq 1}$ of $[H]_\nu^\parPowerNorm$ yields
\[ 
\bigl\|\optFP - [\optRegFP]_\nu\bigr\|_{\parPowerNorm}^2 
= \lambda^2 \sum_{i\geq 1}\Bigl(\frac{\eigw_i^{-\sfrac{\parPowerNorm}{2}}}{\eigw_i + \lambda}\Bigr)^2 a_i^2
= \lambda^2 \sum_{i\geq 1}\Bigl(\frac{\eigw_i^{\frac{\parSourceCond-\parPowerNorm}{2}}}{\eigw_i + \lambda}\Bigr)^2 \eigw_i^{-\parSourceCond}a_i^2\;\;.
\]
If we estimate the fraction on the right hand side with Lemma~\ref{lem:apx:estimate} and apply Parseval's identity w.r.t.\ the ONB $(\eigw_i^{\sfrac{\parSourceCond}{2}}[\eigv_i]_\nu)_{i\geq 1}$ of $[H]_\nu^\parSourceCond$, then we get
\[ 
\bigl\|\optFP - [\optRegFP]_\nu\bigr\|_{\parPowerNorm}^2 
\leq \biggl(\lambda \sup_{i\geq 1} \frac{\eigw_i^{\frac{\parSourceCond - \parPowerNorm}{2}}}{\eigw_i + \lambda}\biggr)^2 \sum_{i\geq 1}\eigw_i^{-\parSourceCond}a_i^2
\leq \lambda^{\parSourceCond-\parPowerNorm} \sum_{i\geq 1} \eigw_i^{-\parSourceCond}a_i^2 = \lambda^{\parSourceCond-\parPowerNorm} \|\optFP\|_{\parSourceCond}^2\;\;.
\]

In order to show \eqref{it:proof:bounds:power_norm:svm} we use the spectral representation in \eqref{eq:proof:bounds:svm} and Parseval's identity
\[ 
\|[\optRegFP]_\nu\|_{\parPowerNorm}^2 
=\sum_{i\geq 1} \Bigl(\frac{\eigw_i}{\eigw_i + \lambda}\Bigr)^2 \eigw_i^{-\parPowerNorm} a_i^2\;\;.
\]
In the case of $\parPowerNorm\leq\parSourceCond$ we estimate the fraction by $1$ and then Parseval's identity gives us
\[ 
\|[\optRegFP]_\nu\|_{\parPowerNorm}^2 
\leq \sum_{i\geq 1} \eigw_i^{-\parPowerNorm} a_i^2 =\|\optFP\|_\parPowerNorm^2\;\;.
\]
In the case of $\parPowerNorm>\parSourceCond$ we additionally use Lemma~\ref{lem:apx:estimate} and get
\[ 
\|[\optRegFP]_\nu\|_{\parPowerNorm}^2 
=\sum_{i\geq 1} \Bigl(\frac{\eigw_i^{1-\frac{\parPowerNorm-\parSourceCond}{2}}}{\eigw_i + \lambda}\Bigr)^2 \eigw_i^{-\parSourceCond} a_i^2
\leq \lambda^{-(\parPowerNorm-\parSourceCond)} \sum_{i\geq 1} \eigw_i^{-\parSourceCond}a_i^2
= \lambda^{-(\parPowerNorm-\parSourceCond)} \|\optFP\|_{\parSourceCond}^2\;\;.
\]
Whereby, in the last equality we used Parseval's identity again.
\end{proof}

If we combine the bounds from Lemma~\ref{lem:proof:bounds:power_norm} with \eqref{eq:res:embedding_property} we directly obtain the following $L_\infty(\nu)$ bounds. Note that some parts of the following lemma are already stated by \citet[Corollary~5.5]{StSc2012}.

\begin{cor}\label{cor:proof:bounds:infinity_norm}
Let $(X,\mathcal{B})$ be a measurable space, $H$ be a separable RKHS on $X$ w.r.t.\ a bounded and measurable kernel $k$, $P$ be a probability distribution on $X\times \R$ with $|P|_2<\infty$, and $\nu \coloneqq P_X$ be the marginal distribution on $X$.
If $\optFP \in [H]_\nu^\parSourceCond$ and \eqref{eq:res:embedding_property} are satisfied for some $0\leq \parSourceCond\leq 2$ and $0<\parEmbedding\leq 1$, respectively, then the following bounds are satisfied, for all $0<\lambda\leq 1$:
\begin{align}
\label{it:proof:bounds:infinity_norm:svm_diff} \|[\optRegFP]_\nu - \optFP\|_{L_\infty(\nu)}^2 
&\leq \bigl(\|\optFP\|_{L_\infty(\nu)} + \|k_\nu^\parEmbedding\|_{\infty}\|\optFP\|_{\parSourceCond}\bigr)^2\,\lambda^{\parSourceCond-\parEmbedding} \\
\label{it:proof:bounds:infinity_norm:svm} \|[\optRegFP]_\nu\|_{L_\infty(\nu)}^2
&\leq \|k_\nu^\parEmbedding\|_{\infty}^2\|\optFP\|_{\min\{\parEmbedding,\parSourceCond\}}^2 \lambda^{-(\parEmbedding-\parSourceCond)_+}\;\;.
\end{align}
\end{cor}

\begin{proof}
The bound in \eqref{it:proof:bounds:infinity_norm:svm} is a direct consequence of the Identity~\eqref{eq:proof:embedding:norm-identity} in Theorem~\ref{thm:proof:embedding:equivalence} and \eqref{it:proof:bounds:power_norm:svm} with $\parPowerNorm=\parEmbedding$.

To prove \eqref{it:proof:bounds:infinity_norm:svm_diff} we can assume without loss of generality $\optFP\in L_\infty(\nu)$. In the case of $\parSourceCond\leq\parEmbedding$ we use the triangle inequality, Inequality~\eqref{it:proof:bounds:infinity_norm:svm}, and $\lambda\leq 1$ to find
\begin{align*}
\|\optFP - [\optRegFP]_\nu\|_{L_\infty(\nu)}
&\leq \|\optFP\|_{L_\infty(\nu)} + \|[\optRegFP]_\nu\|_{L_\infty(\nu)}\\
&\leq \bigl(\|\optFP\|_{L_\infty(\nu)} + \|k_\nu^\parEmbedding\|_{\infty}\|\optFP\|_{\parSourceCond}\bigr)\,\lambda^{-\frac{\parEmbedding-\parSourceCond}{2}}\;\;.
\end{align*}
In the case $\parSourceCond>\parEmbedding$, Bound~\eqref{it:proof:bounds:infinity_norm:svm_diff} is a consequence of the Identity~\eqref{eq:proof:embedding:norm-identity} in Theorem~\ref{thm:proof:embedding:equivalence} and \eqref{it:proof:bounds:power_norm:svm_diff} with $\parPowerNorm=\parEmbedding$.
\end{proof}


\subsection{Upper Rates}\label{sec:upper}

In order to establish upper bounds, we split $\|[\optRegFD]_\nu - \optFP\|_{\parPowerNorm}$ into two parts:
\begin{equation}\label{eq:proof:upper:prep}
\bigl\|[\optRegFD]_\nu - \optFP\bigr\|_{\parPowerNorm}
\leq \bigl\|[\optRegFD-\optRegFP]_\nu\bigr\|_{\parPowerNorm} + \bigl\|[\optRegFP]_\nu - \optFP\bigr\|_{\parPowerNorm}\;\;,
\end{equation}
the \emph{estimation error} $\|[\optRegFD-\optRegFP]_\nu\|_{\parPowerNorm}$ and the \emph{approximation error} $\|[\optRegFP]_\nu - \optFP\|_{\parPowerNorm}$. A bound on the approximation error has already been given in Lemma~\ref{lem:proof:bounds:power_norm} and the following inequality controls the estimation error.

\begin{thm}[Error Control Inequality]\label{thm:proof:upper:oi}
Let $(X,\mathcal{B})$ be a measurable space, $H$ be a separable RKHS on $X$ w.r.t.\ a bounded and measurable kernel $k$, $P$ be a probability distribution on $X\times \R$ with $|P|_2<\infty$, and $\nu \coloneqq P_X$ be the marginal distribution on $X$.
Furthermore, let $\|\optFP\|_{L_\infty(\nu)}<\infty$, $\|k_\nu^\parEmbedding\|_\infty<\infty$, and \eqref{eq:res:moment_condition} be satisfied.
Then for the abbreviations
\begin{align}
\label{eq:proof:upper:oi:glambda}
g_\lambda 
&\coloneqq \log\biggl(2 e\effdimMas(\lambda) \frac{\|\rTmas\| + \lambda}{\|\rTmas\|}\biggr),\\
\label{eq:proof:upper:oi:indexBound}
\indexBound
&\coloneqq 8\|k_\nu^\parEmbedding\|_{\infty}^2 \tau g_\lambda\lambda^{-\parEmbedding}, \text{ and}\\[1ex]
\label{eq:proof:upper:oi:Llambda}
\supb_\lambda
&\coloneqq \max\{\supb, \|\optFP - [\optRegFP]_\nu\|_{L_\infty(\nu)}\}
\end{align}
and $0\leq\parPowerNorm\leq 1$, $\tau\geq 1$, $\lambda>0$, and $n\geq\indexBound$, the following bound is satisfied with $P^n$-probability not less than $1 - 4 e^{-\tau}$
\[
\Bigl\|\rTmas^{\frac{1-\parPowerNorm}{2}}\bigl(\optRegFD - \optRegFP\bigr)\Bigr\|_H^2
\leq \frac{576\tau^2}{n \lambda^\parPowerNorm}\biggl(\varb^2\effdimMas(\lambda) + \|k_\nu^\parEmbedding\|_{\infty}^2\frac{\|\optFP - [\optRegFP]_\nu\|_{L_2(\nu)}^2}{\lambda^\parEmbedding} + 2\|k_\nu^\parEmbedding\|_{\infty}^2\frac{\supb_\lambda^2}{n\lambda^{\parEmbedding}} \biggr)\;\;.
\]
\end{thm}

According to Lemma~\ref{lem:proof:bounds:rkhs_norm} the same result is true for $\|[\optRegFD - \optRegFP]_\nu\|_\parPowerNorm^2$. Moreover, in the case of $\parPowerNorm=1$ the left hand side coincides with $\|\optRegFD - \optRegFP\|_H$. Our proof is based on an argument tracing back to \cite{SmZh2007}. We refine the analysis with some ideas of \cite{CaDe2007} and \cite{LiCe2018} under the embedding property. We split the proof into several lemmas: the first one improves Lemma~18 of \cite{LiCe2018} under the additional Assumption~\eqref{eq:res:embedding_property}.

\begin{lem}\label{lem:proof:upper:oi_part_i}
Let the assumptions of Theorem~\ref{thm:proof:upper:oi} be satisfied and $g_\lambda$ as defined in \eqref{eq:proof:upper:oi:glambda}. Then, for $\tau\geq 1$, $\lambda>0$, and $n\geq 1$, the following operator norm bound is satisfied with $\nu^n$-probability not less than $1-2 e^{-\tau}$
\begin{equation}\label{eq:proof:upper:oi_part_i}
\bigl\|(\rTmas + \lambda)^{-\sfrac{1}{2}}(\rTmas - \rT_\delta)(\rTmas + \lambda)^{-\sfrac{1}{2}}\bigr\|
\leq \frac{4\|k_\nu^\parEmbedding\|_{\infty}^2 \tau g_\lambda}{3n\lambda^\parEmbedding} + \sqrt{\frac{2\|k_\nu^\parEmbedding\|_{\infty}^2\tau g_\lambda}{n\lambda^\parEmbedding}}\;\;.
\end{equation}
\end{lem}

\begin{proof}
This is a consequence of the concentration inequality in Theorem~\ref{thm:apx:bernstein_operator}, but before we start with the main part of the proof we recall some well-known facts about the mapping $\otimes: H\times H\to \mathcal{L}_2(H)$ into the space of Hilbert-Schmidt operators defined by $f\otimes g \coloneqq \langle f,\,\cdot\,\rangle_H\, g$. Since $f\otimes g$ has rank one, $f\otimes g$ is a Hilbert-Schmidt operator. Furthermore, $\otimes$ is bilinear, satisfies the following Hilbert-Schmidt norm and operator norm identity
\begin{equation}\label{eq:proof:upper:oi_part_i:hs_operator}
\|f\otimes g\|_2 = \|f\otimes g\| = \|f\|_H \|g \|_H\;\;,
\end{equation}
and hence $\otimes$ is continuous. Moreover, the adjoint operator is given by $(f\otimes g)^\ast = g\otimes f$ and $\langle (f\otimes g) h, p\rangle_H = \langle f, h\rangle_H\langle g,p\rangle_H$ for $h,p\in H$.
As a result, $f\otimes f$ is, for all $f\in H$, a self-adjoint positive semi-definite Hilbert-Schmidt operator.

Now, we consider $\rTX:H\to H$ the \emph{integral} operator w.r.t.\ the point measure at $x\in X$,
\[
\rTX f 
\coloneqq f(x) k(x,\,\cdot\,) 
= \langle f, k(x,\,\cdot\,)\rangle_H\, k(x,\,\cdot\,)\;\;,
\]
and define the random variables $\zvi_0,\zvi_1:X\to\mathcal{L}_2(H)$ by
\begin{equation*}
\zvi_0(x) \coloneqq \rTX
\qquad\text{and}\qquad
\zvi_1(x) \coloneqq (\rTmas + \lambda)^{-\sfrac{1}{2}}\rTX(\rTmas + \lambda)^{-\sfrac{1}{2}}\;\;.
\end{equation*}
Using the definition of the bilinear operator $\otimes$, the self-adjointness of $(\rTmas + \lambda)^{-\sfrac{1}{2}}$, and the abbreviation $h_x \coloneqq (\rTmas + \lambda)^{-\sfrac{1}{2}}k(x,\,\cdot\,)$ we can represent $\zvi_0$ and $\zvi_1$ as follows
\begin{equation}\label{eq:proof:upper:oi_part_i:representation}
\begin{aligned}
\zvi_0(x)f 
&= \bigl(k(x,\,\cdot\,)\otimes k(x,\,\cdot\,)\bigr) f\\
\zvi_1(x)f 
&= \bigl\langle k(x,\,\cdot\,), (\rTmas + \lambda)^{-\sfrac{1}{2}}f\bigr\rangle_H\ (\rTmas + \lambda)^{-\sfrac{1}{2}}k(x,\,\cdot\,)\\
&= \bigl\langle(\rTmas + \lambda)^{-\sfrac{1}{2}}  k(x,\,\cdot\,), f\bigr\rangle_H\ (\rTmas + \lambda)^{-\sfrac{1}{2}}k(x,\,\cdot\,)\\
&= (h_x \otimes h_x)f\;\;.
\end{aligned}
\end{equation} 
Since $H$ is a separable RKHS w.r.t.\ a measurable kernel, the map $X\to H$, $x\mapsto k(x,\,\cdot\,)$ is measurable, see e.g.\ \cites[Lemma~4.25]{StCh2008}. Consequently, $\zvi_0$ and $\zvi_1$ are measurable, as compositions of measurable functions. Combining \eqref{eq:proof:upper:oi_part_i:hs_operator} with the representations in \eqref{eq:proof:upper:oi_part_i:representation} and Lemma~\ref{lem:proof:bounds:help} we get the supremum bounds, w.r.t.\ the Hilbert-Schmidt norm and the operator norm, 
\begin{align}
\notag
\|\zvi_0(x)\|_2
&=\|\zvi_0(x)\| = \|k(x,\,\cdot\,)\|_H^2 = k(x,x) \leq \|k_\nu^1\|_{\infty}^2 \qquad \text{and}\\
\label{eq:proof:upper:oi_part_i:sup_bound}
\|\zvi_1(x)\|_2
&= \|\zvi_1(x)\|
= \|(\rTmas + \lambda)^{-\sfrac{1}{2}}k(x,\,\cdot\,)\|_H^2 
\leq \|k_\nu^\parEmbedding\|_{\infty}^2 \lambda^{-\parEmbedding} =: B
\end{align}
for $\nu$-almost all $x\in X$. As a consequence of the boundedness w.r.t.\ the Hilbert-Schmidt norm, the mappings $\zvi_0$ and $\zvi_1$ are Bochner-integrable w.r.t. every probability measure $\mu$ on $X$. Combining \cites[Theorem~6 in Chapter~II.2]{DiUh1977} and $\E_{x\sim\mu}\rTX = \rT_\mu$ yields
\begin{equation}\label{eq:proof:upper:oi_part_i:integral}
\E_{\mu}\zvi_1 = (\rTmas + \lambda)^{-\sfrac{1}{2}}\bigl(\E_{x\sim\mu}\rTX\bigr)(\rTmas + \lambda)^{-\sfrac{1}{2}} = (\rTmas + \lambda)^{-\sfrac{1}{2}}\rT_\mu(\rTmas + \lambda)^{-\sfrac{1}{2}}\;\;.
\end{equation}
If we exploit \eqref{eq:proof:upper:oi_part_i:integral} in the case of $\mu=\nu=P_X$ and $\mu = \delta= D_X$, then we get
\begin{equation*}
\frac{1}{n}\sum_{i=1}^n\bigl(\zvi_1(x_i) - \E_{\nu}\zvi_1\bigr) 
= \E_{\delta}\zvi_1 - \E_{\nu}\zvi_1
= (\rTmas + \lambda)^{-\sfrac{1}{2}}(\rT_\delta - \rTmas)(\rTmas + \lambda)^{-\sfrac{1}{2}}
\end{equation*}
for all $D=\bigl((x_i,y_i)\bigr)_{i=1}^n\in(X\times \R)^n$. Consequently, the left hand side of our claimed inequality \eqref{eq:proof:upper:oi_part_i} coincides with the left hand side in Theorem~\ref{thm:apx:bernstein_operator} w.r.t.\ the random variable $\zvi_1$. A supremum bound for $\zvi_1$ is already established in \eqref{eq:proof:upper:oi_part_i:sup_bound} and $\zvi_1(x)$ is a positive semi-definite self-adjoint Hilbert-Schmidt operator because of the representation in \eqref{eq:proof:upper:oi_part_i:representation} and the properties of $\otimes$. Finally, we have to provide a \emph{variance} bound for $\zvi_1$. To this end, recall that for two self-adjoint operators $R$ and $S$ on a Hilbert space we write $R\preccurlyeq S$ iff $S-R$ is a positive semi-definite operator. The representation $\zvi_1(x) = h_x\otimes h_x$ from \eqref{eq:proof:upper:oi_part_i:representation} together with the supremum bound in \eqref{eq:proof:upper:oi_part_i:sup_bound} yields
\[ 
\zvi_1(x)^2 
= \zvi_1(x)\zvi_1(x)
= \|h_x\|_H^2\langle h_x,\,\cdot\,\rangle_H h_x
= \|h_x\|_H^2 \zvi_1(x) 
\preccurlyeq B \zvi_1(x)
\]
for all $\nu$-almost all $x\in X$. Since the relation $\preccurlyeq$ remains true if we integrate both sides we get from the identity in \eqref{eq:proof:upper:oi_part_i:integral} with $\mu=\nu$ the variance bound
\[
\E_\nu(\zvi_1^2)
\preccurlyeq B\E_\nu\zvi_1 
=B(\rTmas + \lambda)^{-\sfrac{1}{2}}\rTmas(\rTmas + \lambda)^{-\sfrac{1}{2}} =:V\;\;.
\]
Note that $V$ is a self-adjoint positive semi-definite operator as an integral over self-adjoint positive semi-definite operators. Moreover, using the spectral representation of $\rTmas$ in \eqref{eq:pre:spectral} and the spectral representation of $(\rTmas + \lambda)^{-\sfrac{1}{2}}$ in \eqref{eq:proof:bounds:operator:ii} with $a=\sfrac{1}{2}$ we get
\[ 
V = B \sum_{i\geq 1}\frac{\eigw_i}{\eigw_i + \lambda}\langle \eigw_i^{\sfrac{1}{2}}\eigv_i,\,\cdot\,\rangle_H\, \eigw_i^{\sfrac{1}{2}}\eigv_i\;\;.
\]
Since the operator norm coincides with the largest eigenvalue we get
\[ 
\|V\| = B\frac{\eigw_1}{\eigw_1 + \lambda} = B\frac{\|\rTmas\|}{\|\rTmas\| + \lambda}\;\;.
\]
Moreover, the trace coincides with the sum of the eigenvalues and hence
\[ 
\tr(V) = B\sum_{i\geq 1}\frac{\eigw_i}{\eigw_i + \lambda} = B\effdimMas(\lambda)\;\;.
\]
Consequently, Theorem~\ref{thm:apx:bernstein_operator} is applicable and together with $\|V\|\leq B$ and $g(V) = g_\lambda$ Theorem~\ref{thm:apx:bernstein_operator} yields the assertion.
\end{proof}

\begin{lem}\label{lem:proof:upper:oi_part_ii}
Let the assumptions of Theorem~\ref{thm:proof:upper:oi} be satisfied and $L_\lambda$ as in \eqref{eq:proof:upper:oi:Llambda}. Then, for $\tau\geq 1$, $\lambda>0$, and $n\geq 1$, the following bound is satisfied with $P^n$-probability not less than $1-2 e^{-\tau}$
\begin{equation}\label{eq:proof:upper:oi_part_ii}
\begin{aligned}
&\bigl\|(\rTmas + \lambda)^{-\sfrac{1}{2}}\bigl((\gD - \rT_\delta\optRegFP) - (\gP - \rTmas\optRegFP)\bigr)\bigr\|_H^2 \\
\leq\ &\frac{64\tau^2}{n}\biggl(\varb^2\effdimMas(\lambda) + \|k_\nu^\parEmbedding\|_{\infty}^2\frac{\|\optFP - [\optRegFP]_\nu\|_0^2}{\lambda^\parEmbedding} + 2\|k_\nu^\parEmbedding\|_{\infty}^2\frac{L_\lambda^2}{n\lambda^{\parEmbedding}} \biggr)\;\;.
\end{aligned}
\end{equation}
\end{lem}

\begin{proof}
We consider the random variables $\zvi_0,\zvi_2:X\times \R\to H$ defined by
\begin{align*}
\zvi_0(x,y) &\coloneqq (y - \optRegFP(x)) k(x,\,\cdot\,)\\
\zvi_2(x,y) &\coloneqq (\rTmas + \lambda)^{-\sfrac{1}{2}} \zvi_0(x,y)\;\;.
\end{align*}
Since $H$ is a separable RKHS w.r.t.\ the measurable kernel $k$ the mappings $x\mapsto k(x,\,\cdot\,)$ and $\optRegFP$ are measurable, see \cites[Lemma~4.24 and Lemma~4.25]{StCh2008}. Consequently, $\zvi_0$ and $\zvi_2$ are measurable, as compositions of measurable functions. Moreover, since our kernel $k$ is bounded also $\optRegFP$ is bounded and
\[ 
\|\zvi_0(x,y)\|_H 
= |y - \optRegFP(x)|\,\|k(x,\,\cdot\,)\|_H 
\leq (|y| + \|\optRegFP\|_{L_\infty(\nu)}) \|k_\nu^1\|_\infty
\]
is satisfied for $\nu$-almost all $x\in X$. As a result $\zvi_0$ is Bochner-integrable w.r.t.\ all probability measures $Q$ on $X\times \R$ with 
\begin{equation*}
|Q|_1 \coloneqq \int_{X\times \R} |y|\ \d Q(x,y) < \infty\;\;. 
\end{equation*}
An analogous bound shows that $\zvi_2$ is Bochner-integrable w.r.t.\ such measures $Q$. Combining \cites[Theorem~6 in Chapter~II.2]{DiUh1977} and \eqref{eq:proof:svm:gP} yields
\begin{align*}
\E_Q\zvi_2 
&= (\rTmas + \lambda)^{-\sfrac{1}{2}} \Bigl(\E_{(x,y)\sim Q}y k(x,\,\cdot\,) - \E_{x\sim Q_X}\optRegFP(x) k(x,\,\cdot\,)\Bigr)\\
&= (\rTmas + \lambda)^{-\sfrac{1}{2}} (\g_{Q} - \rT_{Q_X}\optRegFP)\;\;.
\end{align*}
If we use this identity for $Q=D$ and $Q=P$, then we get
\begin{align*}
\frac{1}{n}\sum_{i=1}^n \Bigl(\zvi_2(x_i,y_i) - \E_P\zvi_2\Bigr) 
&= \E_D\zvi_2 - \E_P\zvi_2\\
&= (\rTmas + \lambda)^{-\sfrac{1}{2}}\Bigl((\gD - \rT_\delta\optRegFP) - (\gP - \rTmas\optRegFP)\Bigr)
\end{align*}
and therefore the left hand side of our claimed Inequality~\eqref{eq:proof:upper:oi_part_ii} coincides with the left hand side of Bernstein's inequality for $H$-valued random variables from Theorem~\ref{thm:apx:bernstein}. Consequently, it remains to bound the $m$-th moment of $\zvi_2$, for $m\geq 2$,
\begin{equation*}
\E_P\|\zvi_2\|_H^m = \int_X \|(\rTmas + \lambda)^{-\sfrac{1}{2}} k(x,\,\cdot\,)\|_H^m \int_\R |y-\optRegFP(x)|^m\ P(\d y|x)\ \d\nu(x)\;\;.
\end{equation*}
First, we consider the inner integral: Using the triangle inequality and \eqref{eq:res:moment_condition} yields
\begin{align*}
\int_\R |y-\optRegFP(x)|^m\ P(\d y|x) 
&\leq 2^{m-1} \Bigl(\|\Id_\R - \optFP(x)\|_{L_m(P(\,\cdot\,|x))}^m + |\optFP(x) - \optRegFP(x)|^m\Bigr)\\
& \leq \frac{1}{2} m! (2\supb)^{m-2} 2\varb^2 + 2^{m-1}|\optFP(x) - \optRegFP(x)|^m\;\;.
\end{align*}
for $\nu$-almost all $x\in X$. If we plug this bound into the outer integral and use the abbreviation $h_x \coloneqq (\rTmas + \lambda)^{-\sfrac{1}{2}} k(x,\,\cdot\,)$ we get 
\begin{equation}\label{eq:proof:upper:oi_part_ii:temp}
\begin{aligned}
\E_P\|\zvi_2\|_H^m
\leq &\ \frac{1}{2} m! (2\supb)^{m-2} 2\varb^2\int_X \|h_x\|_H^m\ \d\nu(x)\\
 +&\  2^{m-1}\int_X \|h_x\|_H^m\, |\optFP(x) - \optRegFP(x)|^m\ \d\nu(x)\;\;.
\end{aligned}
\end{equation}
Using Lemma~\ref{lem:proof:bounds:help}, the first term in \eqref{eq:proof:upper:oi_part_ii:temp} can be bounded by
\begin{align*}
\frac{1}{2} m! (2\supb)^{m-2} 2\varb^2\int_X \|h_x\|_H^m\ \d\nu(x) 
&\leq \frac{1}{2} m! (2\supb)^{m-2} 2\varb^2 \biggl(\frac{\|k_\nu^\parEmbedding\|_\infty}{\lambda^{\sfrac{\parEmbedding}{2}}}\biggr)^{m-2}\int_X \|h_x\|_H^2\ \d\nu(x)\\
&= \frac{1}{2} m! \biggl(\frac{2\supb\|k_\nu^\parEmbedding\|_{\infty}}{\lambda^{\sfrac{\parEmbedding}{2}}}\biggr)^{m-2} 2\varb^2\effdimMas(\lambda)\\
&\leq \frac{1}{2} m! \biggl(\frac{2\supb_\lambda\|k_\nu^\parEmbedding\|_{\infty}}{\lambda^{\sfrac{\parEmbedding}{2}}}\biggr)^{m-2} 2\varb^2\effdimMas(\lambda)\;\;,
\end{align*}
where we only used $\supb\leq \supb_\lambda$ in the last step. Again, using Lemma~\ref{lem:proof:bounds:help}, the second term in \eqref{eq:proof:upper:oi_part_ii:temp} can be bounded by
\begin{align*}
&\ 2^{m-1}\int_X \|h_x\|_H^m\, |\optFP(x) - \optRegFP(x)|^m\ \d\nu(x)\\
\leq&\ \frac{1}{2}\biggl(\frac{2\|k_\nu^\parEmbedding\|_{\infty}}{\lambda^{\sfrac{\parEmbedding}{2}}}\biggr)^m \|\optFP - [\optRegFP]_\nu\|_{L_\infty(\nu)}^{m-2}\int_X |\optFP(x) - \optRegFP(x)|^2\ \d\nu(x)\\
=&\ \frac{1}{2}\biggl(\frac{2\|k_\nu^\parEmbedding\|_{\infty}\|\optFP - [\optRegFP]_\nu\|_{L_\infty(\nu)}}{\lambda^{\sfrac{\parEmbedding}{2}}}\biggr)^{m-2} \|\optFP - [\optRegFP]_\nu\|_{L_2(\nu)}^2\frac{4\|k_\nu^\parEmbedding\|_{\infty}^2}{\lambda^{\parEmbedding}}\\
\leq&\ \frac{1}{2}m!\biggl(\frac{2\supb_\lambda\|k_\nu^\parEmbedding\|_{\infty}}{\lambda^{\sfrac{\parEmbedding}{2}}}\biggr)^{m-2} \|\optFP - [\optRegFP]_\nu\|_{L_2(\nu)}^2\frac{2\|k_\nu^\parEmbedding\|_{\infty}^2}{\lambda^{\parEmbedding}}\;\;,
\end{align*}
where we only used $\|\optFP - [\optRegFP]_\nu\|_{L_\infty(\nu)}\leq \supb_\lambda$ and $2\leq m!$ in the last step. Continuing Estimate~\eqref{eq:proof:upper:oi_part_ii:temp} we get
\begin{align*}
\E_P\|\zvi_2\|_H^m \leq \frac{1}{2}m!\biggl(\frac{2\supb_\lambda\|k_\nu^\parEmbedding\|_{\infty}}{\lambda^{\sfrac{\parEmbedding}{2}}}\biggr)^{m-2} 2\Bigl(\varb^2\effdimMas(\lambda) + \|\optFP - [\optRegFP]_\nu\|_0^2\frac{\|k_\nu^\parEmbedding\|_{\infty}^2}{\lambda^{\parEmbedding}}\Bigr)
\end{align*}
and an application of Bernstein's inequality from Theorem~\ref{thm:apx:bernstein} with $\supb = 2\supb_\lambda\|k_\nu^\parEmbedding\|_{\infty}\lambda^{-\sfrac{\parEmbedding}{2}}$ and $\varb^2 = 2\bigl(\varb^2\effdimMas(\lambda) + \|\optFP - [\optRegFP]_\nu\|_0^2\,\|k_\nu^\parEmbedding\|_{\infty}^2\lambda^{-\parEmbedding}\bigr)$ yield the assertion.
\end{proof}

\begin{myproof}[\jmlrtext{}{Proof }of Theorem~\ref{thm:proof:upper:oi}]
Let us fix some $\tau\geq 1$, $\lambda >0$, and $n\geq \indexBound$. For $D\in(X\times \R)^n$ the representation $\optRegFD=(\rT_\delta + \lambda)^{-1}\gD$ from \eqref{eq:proof:svm:solution_empirical} yields
\[
\rTmas^\frac{1-\parPowerNorm}{2}(\optRegFD - \optRegFP) = \rTmas^\frac{1-\parPowerNorm}{2}(\rT_\delta + \lambda)^{-1}(\gD - (\rT_\delta + \lambda)\optRegFP)\;\;.
\]
When we combine this with the identity $\Id_H = (\rTmas + \lambda)^{-\sfrac{1}{2}}(\rTmas + \lambda)^{\sfrac{1}{2}}$ then we obtain
\begin{subequations}\label{eq:proof:upper:oi:splitting}
\begin{align}
\Bigl\|\rTmas^{\frac{1-\parPowerNorm}{2}}\bigl(\optRegFD - \optRegFP\bigr)\Bigr\|_H^2
\leq &\ \bigl\|\rTmas^\frac{1-\parPowerNorm}{2}(\rTmas + \lambda)^{-\sfrac{1}{2}}\bigr\|^2 \label{eq:proof:upper:oi:splitting:a}\\
\cdot &\ \bigl\| (\rTmas + \lambda)^{\sfrac{1}{2}}(\rT_\delta + \lambda)^{-1}(\rTmas + \lambda)^{\sfrac{1}{2}} \bigr\|^2\label{eq:proof:upper:oi:splitting:b}\\
\cdot &\ \|(\rTmas + \lambda)^{-\sfrac{1}{2}}(\gD - (\rT_\delta + \lambda)\optRegFP)\|_H^2\label{eq:proof:upper:oi:splitting:c}
\end{align}
\end{subequations}
for all $D\in(X\times \R)^n$. Now, we consider the three factors on the right hand side separately. Let us start with Term~\eqref{eq:proof:upper:oi:splitting:a}. An application of Lemma~\ref{lem:apx:estimate} yields
\begin{equation}\label{eq:proof:upper:oi:splitting:a_estimate}
\bigl\|\rTmas^\frac{1-\parPowerNorm}{2}(\rTmas + \lambda)^{-\sfrac{1}{2}}\bigr\|^2 
= \sup_{i\geq 1} \frac{\eigw_i^{1-\parPowerNorm}}{\eigw_i + \lambda} 
\leq \lambda^{-\parPowerNorm}\;\;.
\end{equation}
Next, Factor~\eqref{eq:proof:upper:oi:splitting:c} can be rearranged using $\optRegFP=(\rTmas + \lambda)^{-1}\gP$ from \eqref{eq:proof:svm:solution}:
\begin{align*}
(\rTmas + \lambda)^{-\sfrac{1}{2}}\bigl(\gD - (\rT_\delta + \lambda)\optRegFP\bigr)
&= (\rTmas + \lambda)^{-\sfrac{1}{2}}\bigl(\gD - (\rT_\delta - \rTmas + \rTmas + \lambda)\optRegFP\bigr)\\
&=(\rTmas + \lambda)^{-\sfrac{1}{2}}\bigl((\gD - \rT_\delta\optRegFP) - (\gP - \rTmas\optRegFP)\bigr)\;\;.
\end{align*}
Consequently, the Factor~\eqref{eq:proof:upper:oi:splitting:c} coincides with the right hand side in Lemma~\ref{lem:proof:upper:oi_part_ii} and this lemma yields
\begin{equation}\label{eq:proof:upper:oi:splitting:c_estimate}
\begin{aligned}
&\ \|(\rTmas + \lambda)^{-\sfrac{1}{2}}(\gD - (\rT_\delta + \lambda)\optRegFP)\|_H^2\\
\leq&\ \frac{64\tau^2}{n}\biggl(\varb^2\effdimMas(\lambda) + \|k_\nu^\parEmbedding\|_{\infty}^2\frac{\|\optFP - [\optRegFP]_\nu\|_0^2}{\lambda^\parEmbedding} + 2\|k_\nu^\parEmbedding\|_{\infty}^2\frac{L_\lambda^2}{n\lambda^{\parEmbedding}} \biggr)
\end{aligned}
\end{equation}
with $P^n$-probability not less than $1 - 2 e^{-\tau}$. Finally, in order to estimate \eqref{eq:proof:upper:oi:splitting:b} we start with the following identity
\begin{align*}
\rT_\delta + \lambda
&= \rT_\delta - \rTmas + \rTmas + \lambda\\ 
&= -(\rTmas - \rT_\delta) + (\rTmas + \lambda)^{\sfrac{1}{2}}(\rTmas + \lambda)^{\sfrac{1}{2}}\\
&= (\rTmas + \lambda)^{\sfrac{1}{2}}\ \Bigl(\Id - (\rTmas + \lambda)^{-\sfrac{1}{2}}(\rTmas - \rT_\delta)(\rTmas + \lambda)^{-\sfrac{1}{2}}\Bigr)\ (\rTmas + \lambda)^{\sfrac{1}{2}}\;\;.
\end{align*}
Plugging this into \eqref{eq:proof:upper:oi:splitting:b}, we get
\begin{align*}
&\ \bigl\| (\rTmas + \lambda)^{\sfrac{1}{2}}(\rT_\delta + \lambda)^{-1}(\rTmas + \lambda)^{\sfrac{1}{2}} \bigr\|^2\\
=&\ \Bigl\| \Bigl(\Id - (\rTmas + \lambda)^{-\sfrac{1}{2}}(\rTmas - \rT_\delta)(\rTmas + \lambda)^{-\sfrac{1}{2}}\Bigr)^{-1} \Bigr\|^2\;\;.
\end{align*}
Lemma~\ref{lem:proof:upper:oi_part_i} gives us an estimate for the operator norm of $(\rTmas + \lambda)^{-\sfrac{1}{2}}(\rTmas - \rT_\delta)(\rTmas + \lambda)^{-\sfrac{1}{2}}$. Continuing the estimate from Lemma~\ref{lem:proof:upper:oi_part_i} with $n\geq \indexBound$ and $\indexBound = 8\|k_\nu^\parEmbedding\|_{\infty}^2 \tau g_\lambda\lambda^{-\parEmbedding}$ from \eqref{eq:proof:upper:oi:indexBound} yields
\begin{align*}
\bigl\|(\rTmas + \lambda)^{-\sfrac{1}{2}}(\rTmas - \rT_\delta)(\rTmas + \lambda)^{-\sfrac{1}{2}}\bigr\|
&\leq \frac{4}{3}\cdot\frac{\|k_\nu^\parEmbedding\|_\infty^2\tau g_\lambda}{n\lambda^\parEmbedding}  + \sqrt{2\cdot\frac{\|k_\nu^\parEmbedding\|_\infty^2\tau g_\lambda}{n\lambda^\parEmbedding}}\\
&\leq \frac{4}{3}\cdot\frac{1}{8} + \sqrt{2\cdot\frac{1}{8}}
=\frac{2}{3}
\end{align*}
with $\nu^n$-probability not less than $1-2 e^{-\tau}$. Consequently, the inverse of 
\[
\Id - (\rTmas + \lambda)^{-\sfrac{1}{2}}(\rTmas - \rT_\delta)(\rTmas + \lambda)^{-\sfrac{1}{2}} 
\]
can be represented by the Neumann series. In particular, the Neumann series gives us the following bound on \eqref{eq:proof:upper:oi:splitting:b}

\begin{equation}\label{eq:proof:upper:oi:splitting:b_estimate}
\begin{aligned}
&\ \bigl\| (\rTmas + \lambda)^{\sfrac{1}{2}}(\rT_\delta + \lambda)^{-1}(\rTmas + \lambda)^{\sfrac{1}{2}} \bigr\|^2\\
=&\ \Bigl\| \Bigl(\Id - (\rTmas + \lambda)^{-\sfrac{1}{2}}(\rTmas - \rT_\delta)(\rTmas + \lambda)^{-\sfrac{1}{2}}\Bigr)^{-1} \Bigr\|^2\\
\leq&\ \biggl(\sum_{k=0}^\infty \bigl\|(\rTmas + \lambda)^{-\sfrac{1}{2}}(\rTmas - \rT_\delta)(\rTmas + \lambda)^{-\sfrac{1}{2}}\bigr\|^k\biggr)^2\\
\leq&\ \biggl(\sum_{k=0}^\infty \Bigl(\frac{2}{3}\Bigr)^k\biggr)^2 = 9
\end{aligned}
\end{equation}
with $\nu^n$-probability not less than $1 - 2 e^{-\tau}$. Now, if we combine the estimate in \eqref{eq:proof:upper:oi:splitting} with \eqref{eq:proof:upper:oi:splitting:a_estimate}, \eqref{eq:proof:upper:oi:splitting:c_estimate}, and \eqref{eq:proof:upper:oi:splitting:b_estimate}, then we get the claimed bound, with $P^n$-probability not less than $1-4 e^{-\tau}$.
\end{myproof}

\begin{myproof}[\jmlrtext{}{Proof }of Theorem~\ref{thm:res:upper_rates}]
Let us fix some $\tau\geq 1$ and some lower bound $0<c\leq 1$ with $c\leq\|\rTmas\|$.
First, we show that Theorem~\ref{thm:proof:upper:oi} is applicable. To this end, we prove in both cases, $\parSourceCond+\parEigDecay\leq\parEmbedding$ and $\parSourceCond+\parEigDecay>\parEmbedding$, that there is an index bound $n_0\geq 1$ such that $n\geq \indexBound[\lambda_n]$ is satisfied for all $n\geq n_0$. Since $\lambda_n\to 0$ we choose $n'_0\geq 1$ such that $\lambda_n\leq c \leq \min\{1,\|\rTmas\|\}$ for all $n\geq n'_0$. 
Using the definitions of $\indexBound[\lambda_n]$ and $g_\lambda$ in \eqref{eq:proof:upper:oi:indexBound} and \eqref{eq:proof:upper:oi:glambda}, respectively, $\lambda_n\leq c \leq \|\rTmas\|$, $\effdimMas(\lambda_n)\leq\constEffDim \lambda_n^{-\parEigDecay}$ from Lemma~\ref{lem:proof:effective_dimension}, and $\|k_\nu^\parEmbedding\|_\infty\leq\constEmbedding$ from \eqref{eq:res:embedding_property} and \eqref{eq:proof:embedding:norm-identity} we get, for $n\geq n'_0$,
\begin{align*}
\frac{\indexBound[\lambda_n]}{n} 
&= 8\|k_\nu^\parEmbedding\|_{\infty}^2 \tau \frac{ g_{\lambda_n}}{n\lambda_n^{\parEmbedding}}\\
&= 8\|k_\nu^\parEmbedding\|_{\infty}^2 \tau \frac{ \log\bigl(2 e \effdimMas(\lambda_n) (1+\sfrac{\lambda_n}{\|\rTmas\|})\bigr)}{n\lambda_n^{\parEmbedding}}\\
&\leq 8\constEmbedding^2 \tau \frac{ \log\bigl(4 e\constEffDim \lambda_n^{-\parEigDecay} \bigr)}{n\lambda_n^{\parEmbedding}}\\
&= 8\constEmbedding^2 \tau \biggl(\frac{ \log\bigl(4 e\constEffDim\bigr)}{n\lambda_n^{\parEmbedding}} + \parEigDecay\frac{\log\bigl(\lambda_n^{-1} \bigr)}{n\lambda_n^{\parEmbedding}}\biggr)\;\;.
\end{align*}
Consequently, it is enough to show $\frac{\log(\lambda_n^{-1} )}{n\lambda_n^{\parEmbedding}} \to 0$. To this end, we consider the cases $\parSourceCond+\parEigDecay\leq\parEmbedding$ and $\parSourceCond+\parEigDecay>\parEmbedding$ separately.

\ref{it:res:upper_rates:i} In the case of $\parSourceCond+\parEigDecay\leq\parEmbedding$ we have $\lambda_n\asymp\bigl(\sfrac{n}{\log^r(n)}\bigr)^{-\sfrac{1}{\parEmbedding }}$ for some $r>1$ and hence
\begin{equation*}
\frac{\log\bigl(\lambda_n^{-1} \bigr)}{n\lambda_n^{\parEmbedding}}
\asymp \frac{\log(n)}{n (n/\log^r(n))^{-1}} 
= \frac{1}{\log^{r-1}(n)}\to 0\;\;.
\end{equation*}

\ref{it:res:upper_rates:ii} In the case of $\parSourceCond + \parEigDecay>\parEmbedding$ we have $1-\frac{\parEmbedding}{\parSourceCond + \parEigDecay}>0$, $\lambda_n\asymp n^{-\sfrac{1}{(\parSourceCond + \parEigDecay)}}$, and hence
\begin{equation*}
\frac{\log\bigl(\lambda_n^{-1} \bigr)}{n\lambda_n^{\parEmbedding}}
\asymp \frac{\log(n)}{n^{1-\frac{\parEmbedding}{\parSourceCond + \parEigDecay}}}\to 0\;\;.
\end{equation*}
Consequently, there is a $n_0\geq n'_0$ with $n\geq\indexBound[\lambda_n]$ for all $n\geq n_0$. Moreover, $n_0$ just depends on $(\lambda_n)_{n\geq 1}$, $c$, $\tau$, $\constEmbedding$, $\constEffDim$, and on the parameters $\parEmbedding,\parEigDecay$.

Let $n\geq n_0$ be fixed. From Lemma~\ref{lem:proof:bounds:rkhs_norm} and Theorem~\ref{thm:proof:upper:oi} we get the bound
\begin{equation*}
\|[\optRegFD[\lambda_n] - \optRegFP[\lambda_n]]_\nu\|_\parPowerNorm^2
\leq \frac{576\tau^2}{n \lambda_n^\parPowerNorm}\biggl(\varb^2\effdimMas(\lambda_n) + \|k_\nu^\parEmbedding\|_{\infty}^2\frac{\|\optFP - [\optRegFP[\lambda_n]]_\nu\|_{L_2(\nu)}^2}{\lambda_n^\parEmbedding} + 2\|k_\nu^\parEmbedding\|_{\infty}^2\frac{\supb_{\lambda_n}^2}{n\lambda_n^{\parEmbedding}} \biggr)\;\;.
\end{equation*}
Continuing this estimate by using $\effdimMas(\lambda_n)\leq\constEffDim \lambda_n^{-\parEigDecay}$ from Lemma~\ref{lem:proof:effective_dimension}, $\|k_\nu^\parEmbedding\|_\infty\leq\constEmbedding$ from \eqref{eq:res:embedding_property} and \eqref{eq:proof:embedding:norm-identity}, and $\|\optFP - [\optRegFP[\lambda_n]]_\nu\|_{L_2(\nu)}^2 \leq \constSourceCond^2 \lambda_n^{\parSourceCond}$ from Lemma~\ref{lem:proof:bounds:power_norm} and \eqref{eq:res:source_condition} we get
\begin{equation}\label{eq:proof:upper:thm:temp}
\|[\optRegFD[\lambda_n] - \optRegFP[\lambda_n]]_\nu\|_\parPowerNorm^2
\leq 576 \frac{\tau^2}{n \lambda_n^\parPowerNorm}\biggl(\varb^2\constEffDim\lambda_n^{-\parEigDecay} + \constEmbedding^2\constSourceCond^2\lambda_n^{\parSourceCond-\parEmbedding} + 2\constEmbedding^2\frac{\supb_{\lambda_n}^2}{n\lambda_n^{\parEmbedding}} \biggr)\;\;.
\end{equation}
Combining the definition of $\supb_\lambda$ in \eqref{eq:proof:upper:oi:Llambda} with Corollary~\ref{cor:proof:bounds:infinity_norm} and $\lambda_n\leq 1$ we get 
\begin{align*}
\supb_{\lambda_n}^2 
&= \max\bigl\{\supb^2, \|\optFP - [\optRegFP[\lambda_n]]_\nu\|_{L_\infty(\nu)}^2\bigr\}\\
&\leq \max\bigl\{\supb^2, \bigl(\|\optFP\|_{L_\infty(\nu)} + \|k_\nu^\parEmbedding\|_{\infty}\|\optFP\|_{\parSourceCond}\bigr)^2\lambda_n^{-(\parEmbedding-\parSourceCond)}\bigr\}\\
&\leq K_0\, \lambda_n^{-(\parEmbedding-\parSourceCond)_+}
\end{align*}
with $K_0 \coloneqq \max\bigl\{\supb^2, \bigl(\constInftyBound + \constEmbedding\constSourceCond\bigr)^2\bigr\}$. For the first and second addend in \eqref{eq:proof:upper:thm:temp} we use again $\lambda_n\leq 1$ and get
\begin{equation*}
\varb^2\constEffDim\lambda_n^{-\parEigDecay} + \constEmbedding^2\constSourceCond^2\lambda_n^{\parSourceCond-\parEmbedding}
\leq \bigl(\varb^2\constEffDim + \constEmbedding^2\constSourceCond^2\bigr)\max\{\lambda_n^{-\parEigDecay}, \lambda_n^{-(\parEmbedding-\parSourceCond)}\}
= K_1 \lambda_n^{-\max\{\parEigDecay, \parEmbedding-\parSourceCond\}}
\end{equation*}
with $K_1 \coloneqq \varb^2\constEffDim + \constEmbedding^2\constSourceCond^2$. Plugging both bounds into \eqref{eq:proof:upper:thm:temp} gives us
\begin{align*}
\|[\optRegFD[\lambda_n] - \optRegFP[\lambda_n]]_\nu\|_\parPowerNorm^2
&\leq 576 \frac{\tau^2}{n \lambda_n^\parPowerNorm}\biggl(K_1 \lambda_n^{-\max\{\parEigDecay,\parEmbedding-\parSourceCond\}} + 2\constEmbedding^2 K_0\frac{1}{n\lambda_n^{\parEmbedding + (\parEmbedding - \parSourceCond)_+}} \biggr)\\
&= 576 \frac{\tau^2}{n \lambda_n^{\parPowerNorm + \max\{\parEigDecay,\parEmbedding-\parSourceCond\}}}\biggl(K_1 + 2\constEmbedding^2 K_0\frac{1}{n\lambda_n^{\parEmbedding + (\parEmbedding - \parSourceCond)_+ - \max\{\parEigDecay,\parEmbedding-\parSourceCond\}}} \biggr)\;\;.
\end{align*}
Next, we show that the second term in the brackets is bounded. To this end, we consider the cases $\parSourceCond+\parEigDecay\leq\parEmbedding$ and $\parSourceCond+\parEigDecay>\parEmbedding$ separately.

\ref{it:res:upper_rates:i} In the case of $\parSourceCond+\parEigDecay\leq\parEmbedding$ we have $0<\parEigDecay\leq \parEmbedding-\parSourceCond$ and
\begin{equation*}
\parEmbedding + (\parEmbedding - \parSourceCond)_+ - \max\{\parEigDecay,\parEmbedding-\parSourceCond\}
= \parEmbedding\;\;.
\end{equation*}
Since $\lambda_n\asymp\bigl(\sfrac{n}{\log^r(n)}\bigr)^{-\sfrac{1}{\parEmbedding }}$ for some $r>1$ we get
\begin{equation*}
\frac{1}{n\lambda_n^{\parEmbedding + (\parEmbedding - \parSourceCond)_+ - \max\{\parEigDecay,\parEmbedding-\parSourceCond\}}}
= \frac{1}{n\lambda_n^\parEmbedding}
\asymp \frac{1}{\log^r(n)}\;\;.
\end{equation*}

\ref{it:res:upper_rates:ii} In the case of $\parSourceCond + \parEigDecay>\parEmbedding$ we have $\parEigDecay>\parEmbedding-\parSourceCond$, $\lambda_n\asymp n^{-\sfrac{1}{(\parSourceCond + \parEigDecay)}}$, and hence
\begin{equation*}
\frac{1}{n\lambda_n^{\parEmbedding + (\parEmbedding - \parSourceCond)_+ - \max\{\parEigDecay,\parEmbedding-\parSourceCond\}}}
= \frac{1}{n\lambda_n^{\parEmbedding + (\parEmbedding - \parSourceCond)_+ - \parEigDecay}}
\asymp \Bigl(\frac{1}{n}\Bigr)^{1-\frac{\parEmbedding + (\parEmbedding-\parSourceCond)_+ -\parEigDecay}{\parSourceCond + \parEigDecay}}\;\;.
\end{equation*}
Using $\parEigDecay>\parEmbedding - \parSourceCond$ again gives us
\begin{equation*}
1-\frac{\parEmbedding + (\parEmbedding-\parSourceCond)_+ -\parEigDecay}{\parSourceCond + \parEigDecay}
= \frac{2\parEigDecay- (\parEmbedding-\parSourceCond) - (\parEmbedding-\parSourceCond)_+}{\parSourceCond + \parEigDecay} > 0
\end{equation*}
Consequently, there is a constant $K_2>0$ with 
\begin{equation*}
\|[\optRegFD[\lambda_n] - \optRegFP[\lambda_n]]_\nu\|_\parPowerNorm^2
= 576 \frac{\tau^2}{n \lambda_n^{\parPowerNorm + \max\{\parEigDecay,\parEmbedding-\parSourceCond\}}}\bigl(K_1 + 2\constEmbedding^2 K_0 K_2 \bigr)
\end{equation*}
for all $n\geq n_0$. Combining this with the splitting in \eqref{eq:proof:upper:prep} and with Lemma~\ref{lem:proof:bounds:power_norm} yields, for $K_3 \coloneqq 576(K_1 + 2\constEmbedding^2 K_0 K_2)$,
\begin{align*}
\bigl\|[\optRegFD]_\nu - \optFP\bigr\|_{\parPowerNorm}^2
&\leq 2\constSourceCond^2\lambda_n^{\parSourceCond-\parPowerNorm} + 2 K_3\frac{\tau^2}{n \lambda_n^{\parPowerNorm + \max\{\parEigDecay,\parEmbedding-\parSourceCond\}}}\\
&\leq \tau^2\lambda_n^{\parSourceCond-\parPowerNorm}\biggl(2\constSourceCond^2 + 2 K_3\frac{1}{n\lambda_n^{\max\{\parEmbedding, \parSourceCond+\parEigDecay\}}}\biggr)\;\;.
\end{align*}
Since in both cases, $\parSourceCond+\parEigDecay \leq \parEmbedding$ and $\parSourceCond+\parEigDecay >\parEmbedding$, we have $\lambda_n \succcurlyeq n^{-\sfrac{1}{\max\{\parEmbedding,\parSourceCond + \parEigDecay\}}}$ the term inside the brackets is bounded by some constant $K>0$ and hence we have
\begin{equation*}
\bigl\|[\optRegFD]_\nu - \optFP\bigr\|_{\parPowerNorm}^2
\leq \tau^2 K \lambda_n^{\parSourceCond - \parPowerNorm}
\end{equation*}
for all $n\geq n_0$. This is the assertion, in both cases.
\end{myproof}


\subsection{Lower Rates} 

We establish the following lower bound in order to prove $\parPowerNorm$-lower rates .

\begin{lem}[Lower Bound]\label{lem:proof:lower:bound}
Let $(X,\mathcal{B})$ be a measurable space, $H$ be a separable RKHS on $X$ w.r.t.\ a bounded and measurable kernel $k$, and $\nu$ be a probability distribution on $X$ such that \eqref{eq:res:embedding_property} and \eqref{eq:res:eigenvalue_decay_exact} are satisfied for some $0<\parEigDecay\leq\parEmbedding\leq 1$. Then, for all parameters $0<\parSourceCond\leq 2$, $0\leq\parPowerNorm\leq 1$ with $\parPowerNorm<\parSourceCond$ and all constants $\varb,\supb, \constSourceCond,\constInftyBound>0$, there exist constants $0<\varepsilon_0\leq 1$ and $\constLB_1,\constLB_2>0$ such that the following statement is satisfied. For all $0<\varepsilon\leq\varepsilon_0$ there is a $M_\varepsilon\geq 1$ with 
\begin{equation}\label{eq:proof:lower:size}
2^{\constLB_2\varepsilon^{-u}} \leq M_\varepsilon \leq 2^{3\constLB_2\varepsilon^{-u}}
\end{equation}
for $u \coloneqq \frac{\parEigDecay}{\max\{\parEmbedding, \parSourceCond\} - \parPowerNorm}$ and there are probability measures $P_0,P_1,\ldots,P_{M_\varepsilon}$ with marginal distribution $(P_j)_X = \nu$ on $X$, $\|\optFZ{P_j}\|_{L_\infty(\nu)}\leq\constInftyBound$, \eqref{eq:res:source_condition} w.r.t.\ $\parSourceCond,\constSourceCond$, and \eqref{eq:res:moment_condition} w.r.t.\ $\varb,\supb$. Moreover, $P_0,P_1,\ldots,P_{M_\varepsilon}$ satisfy
\begin{equation}\label{eq:proof:lower:packing}
\|\optFZ{P_i} - \optFZ{P_j}\|_{\parPowerNorm}^2 \geq 4\varepsilon
\end{equation}
for all $i,j\in\{0,1,\ldots,M_\varepsilon\}$ with $i\not=j$ and
\begin{equation}\label{eq:proof:lower:lower_bound}
\max_{j=0,1,\ldots,M_\varepsilon} P_j^n\bigl(D:\ \Psi(D)\not=j\bigr) 
\geq \frac{\sqrt{M_\varepsilon}}{1+\sqrt{M_\varepsilon}} \Bigl(1 - \constLB_1 n\varepsilon^{\frac{\max\{\parEmbedding,\parSourceCond\} + \parEigDecay}{\max\{\parEmbedding,\parSourceCond\} - \parPowerNorm}} - \frac{1}{2\log(M_\varepsilon)}\Bigr)
\end{equation}
for all $n\geq 1$ and all measurable functions $\Psi:(X\times \R)^n\to \{0,1,\ldots,M_\varepsilon\}$.
\end{lem}

Note that the probability measures $P_j$ also depend on $\varepsilon$ although we omit this in the notation. Moreover, just one probability measure $\nu$ on $X$ with the required properties is needed to construct distributions on $X\times\R$ that are \emph{difficult} to learn. The proof of Lemma~\ref{lem:proof:lower:bound} is an application of \cites[Proposition~2.3]{Ts2009} stated in the following theorem. To this end, recall that the \emph{Kullback-Leibler divergence} of two probability measures $P_1, P_2$ on some measurable space $(\Omega,\mathcal{A})$ is given by
\begin{equation*}
K(P_{1},P_{2}) \coloneqq \int_\Omega \log\biggl(\frac{\d P_1}{\d P_2}\biggr)\ \d P_1
\end{equation*}
if $P_1\ll P_2$ and otherwise $K(P_{1},P_{2}) \coloneqq \infty$. 

\begin{thm}[Lower Bound]\label{thm:lower:general}
Let $M\geq 2$, $(\Omega,\mathcal{A})$ be a measurable space, $P_0,P_1,\ldots,P_M$ be probability measures on $(\Omega,\mathcal{A})$ with $P_{j}\ll P_{0}$ for all $j=1,\ldots,M$, and $0<\alpha_\ast<\infty$ with 
\begin{equation*}
\frac{1}{M}\sum_{j=1}^M K(P_{j},P_{0}) \leq \alpha_\ast\;\;.
\end{equation*}
Then, for all measurable functions $\Psi:\Omega\to\{0,1,\ldots,M\}$, the following bound is satisfied
\[ 
\max_{j=0,1,\ldots,M} P_{j} \bigl(\omega\in\Omega:\ \Psi(\omega)\not=j \bigr) 
\geq \frac{\sqrt{M}}{1 + \sqrt{M}} \biggl(1 - \frac{3\alpha_\ast}{\log(M)} - \frac{1}{2\log(M)}\biggr)\;.
\]
\end{thm}

\begin{proof}
From \cites[Proposition~2.3]{Ts2009} we know, that
\[ 
\max_{j=0,1,\ldots,M} P_{j} \bigl(\omega\in\Omega:\ \Psi(\omega)\not=j \bigr)
\geq \sup_{0<\tau<1} \frac{\tau M}{1 + \tau M} \biggl(1 + \frac{\alpha_\ast + \sqrt{\sfrac{\alpha_\ast}{2}}}{\log(\tau)}\biggr)
\]
is satisfied. If we choose $\tau = M^{-\sfrac{1}{2}}$, then we get
\begin{align*}
\max_{j=0,1,\ldots,M} P_{j} \bigl(\omega\in\Omega:\ \Psi(\omega)\not=j \bigr)
&\geq \frac{\sqrt{M}}{1 + \sqrt{M}} \biggl(1 - \frac{2\alpha_\ast + \sqrt{2\alpha_\ast}}{\log(M)}\biggr)\\
&\geq \frac{\sqrt{M}}{1 + \sqrt{M}} \biggl(1 - \frac{3\alpha_\ast}{\log(M)} - \frac{1}{2\log(M)}\biggr)\;\;,
\end{align*}
where we used the estimate $\sqrt{2\alpha_\ast}\leq \sfrac{1}{2} + \alpha_\ast$ in the last inequality.
\end{proof}

We use this theorem for the measurable space $\Omega=(X\times \R)^n$ and follow the suggestion of \citet{CaDe2007} and \citet{BlM2017} in order to construct a family of probability measures $P_0,P_1,\ldots,P_M$. To this end, let the assumptions of Lemma~\ref{lem:proof:lower:bound} be satisfied and set $\varLB \coloneqq \min\{\varb, \supb\}$. Moreover, we define for a measurable function $f:X\to \R$ and $x\in X$ the conditional distribution $P_f(\,\cdot\,|x) \coloneqq \mathcal{N}(f(x),\varLB^2)$ as the normal distribution on $\R$ with mean $f(x)$ and variance $\varLB^2$. Consequently, 
\begin{equation}\label{eq:proof:lower:measure}
P_f(A) \coloneqq \int_X \int_\R \indicator{A}(x,y)\ P_f(\d y|x)\ \d\nu(x)\;,
\end{equation}
for $A\in\mathcal{B}\otimes\algBorel{\R}$, defines a probability measure on $X\times \R$ with marginal distribution $(P_f)_X=\nu$ on $X$. For this reason, the corresponding power spaces $[H]^\parEmbedding_\nu$ are independent of $f$. Since $P_f = P_{f'}$ is satisfied if $f'=f$ $\nu$-a.s.\ we define $P_{[f]_\nu}$ for $\nu$-equivalence classes. Moreover, for $f\in L_2(\nu)$, we get $|P_f|_2^2 = \varLB^2 + \|f\|_{L_2(\nu)}^2 <\infty$ and the conditional mean function $\optFZ{P_f}$ of $P_f$ coincides with $f$. 

\begin{lem}[Moment Condition]\label{lem:proof:lower:moment_condition}
For a measurable function $f:X\to\R$ the probability measure $P_f$ defined in \eqref{eq:proof:lower:measure} satisfies \eqref{eq:res:moment_condition} for $\varb=\supb=\varLB$.
\end{lem}

\begin{proof}
Let us fix an $x\in X$ and an $m\geq 2$. Since $P_f(\,\cdot\,|x) = \mathcal{N}(f(x),\varLB^2)$, the mapping $y\mapsto \sfrac{(y-f(x))}{\varLB}$ is standard normally distributed under the measure $P_f(\,\cdot\,|x)$ and
\begin{equation*}
\int_\R |y - f(x)|^m\ P_f(\d y|x) = \varLB^m \E |Z|^m
\end{equation*}
with some standard normally distributed random variable $Z$. Consequently, it remains to show $\E |Z|^m\leq \sfrac{m!}{2}$. For $m = 2 k$ with some $k\geq 1$ the moments of $Z$ are well-known, see e.g.\ \cites[Equation~\eqnr{4.20}]{Ba1996},
\begin{equation}\label{eq:proof:lower:moment_condition:temp}
\E |Z|^m 
= (m-1)(m-3)\cdot\ldots\cdot 3\cdot 1
\leq \sfrac{m!}{m}
\leq \sfrac{m!}{2}\;\;.
\end{equation}
For $m = 2 k - 1$ with some $k\geq 2$ we use Hölder's inequality to get $(\E|Z|^m)^{\sfrac{1}{m}} \leq (\E|Z|^{m+1})^{\sfrac{1}{(m+1)}}$. Using \eqref{eq:proof:lower:moment_condition:temp} with $m + 1 = 2 k$ and $m\geq 3$ we get
\begin{equation*}
\E |Z|^m 
\leq \bigl(m (m-2)\cdot\ldots\cdot 3\cdot 1\bigr)^{\frac{m}{m+1}}
\leq \bigl(\sfrac{m!}{2}\bigr)^{\frac{m}{m+1}}
\leq\sfrac{m!}{2}\;\;
\end{equation*}
This gives the assertion for all $m\geq 2$.
\end{proof}

To sum up, we reduced the construction of probability measures to the construction of functions $f_0,f_1,\ldots,f_M\in L_\infty(\nu)\cap[H]_\nu^\parSourceCond$ with $\|f_j\|_{L_\infty(\nu)}^2\leq\constInftyBound$ and $\|f_j\|_{\parSourceCond}^2\leq\constSourceCond$ for $j=0,1,\ldots,M$. Before we start with the construction we recall the following lemma from \cites[Proposition~6.2]{BlM2017}.

\begin{lem}[Kullback-Leibler Divergence]\label{lem:lower:kullback_leibler}
For $f,f'\in L_2(\nu)$ and the corresponding probability measures $P_f$, $P_{f'}$ defined in \eqref{eq:proof:lower:measure} the Kullback-Leibler divergence satisfies, for $n\geq 1$,
\[ 
K(P_f^n,P_{f'}^n) = \frac{n}{2\varLB^2} \|f - f'\|_{L_2(\nu)}^2\;\;.
\]
\end{lem}

For the construction of suitable functions we use binary strings $\omega=(\omega_1,\ldots,\omega_m)\in\{0,1\}^m$ and define
\begin{equation}\label{eq:proof:lower:function}
f_\omega \coloneqq 2 \Bigl(\frac{8\varepsilon}{m}\Bigr)^{\sfrac{1}{2}} \sum_{i=1}^m \omega_i\,\eigw_{i+m}^{\sfrac{\parPowerNorm}{2}}\,[\eigv_{i+m}]_\nu
\end{equation}
for $0<\varepsilon\leq 1$. Since the sum is finite we have $f_\omega\in [H]_\nu\subseteq L_\infty(\nu)\cap [H]_\nu^\parSourceCond$. Next, we establish conditions on $\varepsilon$ and $m$ that ensure $\|f_\omega\|_{L_\infty(\nu)}^2\leq\constInftyBound$ and $\|f_\omega\|_{\parSourceCond}^2\leq\constSourceCond$.

\begin{lem}\label{lem:proof:lower:valid_strings}
Under the assumptions of Lemma~\ref{lem:proof:lower:bound} and $u=\frac{\parEigDecay}{\max\{\parEmbedding, \parSourceCond\} - \parPowerNorm}$, for all $0\leq\parSourceCond\leq 2$ and $0\leq\parPowerNorm\leq 1$ with $\parPowerNorm<\parSourceCond$, there are constants $\constValidString>0$ and $0<\varepsilon_1\leq 1$ such that for all $0<\varepsilon\leq\varepsilon_1$ and all $m\leq\constValidString\varepsilon^{-u}$ the function $f_\omega$ defined in \eqref{eq:proof:lower:function} satisfies the bounds $\|f_\omega\|_{\parSourceCond} \leq \constSourceCond$ and $\|f_\omega\|_{L_\infty(\nu)} \leq \constInftyBound$ for all $\omega\in\{0,1\}^m$.
\end{lem}

Note that, if we do \emph{not} require the functions $f_\omega$ to be uniformly bounded, i.e.\ we omit the condition $\|f_\omega\|_{L_\infty(\nu)}\leq\constInftyBound$, then the same result is satisfied for $u=\frac{\parEigDecay}{\parSourceCond - \parPowerNorm}$.

\begin{proof}
Let us fix $m\in\N$ and $0<\varepsilon\leq 1$. First, we consider the condition $\|f_\omega\|_{\parSourceCond}\leq\constSourceCond$. Using \eqref{eq:res:eigenvalue_decay_exact} and $\parPowerNorm<\parSourceCond$ we get
\begin{equation*}
\|f_\omega\|_{\parSourceCond}^2 = \frac{32\varepsilon}{m} \sum_{i=1}^m \omega_i^2 \eigw_{i+m}^{-(\parSourceCond-\parPowerNorm)} \leq 32\, \varepsilon\eigw_{2m}^{-(\parSourceCond-\parPowerNorm)} \leq 32 \constEigDecayLB^{\parPowerNorm-\parSourceCond} 2^{\frac{\parSourceCond - \parPowerNorm}{\parEigDecay}} \varepsilon m^{\frac{\parSourceCond-\parPowerNorm}{\parEigDecay}} \leq\constSourceCond^2
\end{equation*}
for $m \leq \constValidString_1 \varepsilon^{-\frac{\parEigDecay}{\parSourceCond-\parPowerNorm}}$ with $\constValidString_1 \coloneqq \bigl(\sfrac{\constSourceCond^2}{32}\bigr)^{\frac{\parEigDecay}{\parSourceCond-\parPowerNorm}}\sfrac{\constEigDecayLB^\parEigDecay}{2}$.
Next, we consider the condition $\|f_\omega\|_{L_\infty(\nu)} \leq \constInftyBound$ for the cases $\parPowerNorm < \parEmbedding$ and $\parPowerNorm\geq\parEmbedding$ separately. In the case of $\parPowerNorm < \parEmbedding$, \eqref{eq:res:embedding_property} together with an analogues argument with $\parEmbedding$ instead of $\parSourceCond$ yields 
\begin{equation*}
\|f_\omega\|_{L_\infty(\nu)}^2\leq \constEmbedding^2 \|f_\omega\|_{\parEmbedding}^2 \leq \constInftyBound^2
\end{equation*}
for $m\leq \constValidString_2 \varepsilon^{-\frac{\parEigDecay}{\parEmbedding - \parPowerNorm}}$ with $\constValidString_2 \coloneqq \bigl(\sfrac{(\sfrac{\constInftyBound}{\constEmbedding})^2}{32}\bigr)^{\frac{\parEigDecay}{\parEmbedding-\parPowerNorm}}\sfrac{\constEigDecayLB^\parEigDecay}{2}$. Consequently, for $\constValidString \coloneqq \min\{\constValidString_1,\constValidString_2\}$, both conditions, $\|f_\omega\|_{\parSourceCond} \leq \constSourceCond$ and $\|f_\omega\|_{L_\infty(\nu)} \leq \constInftyBound$, are satisfied if 
\begin{equation*}
m
\leq \constValidString \min\Bigl\{\varepsilon^{-\frac{\parEigDecay}{\parSourceCond - \parPowerNorm}}, \varepsilon^{-\frac{\parEigDecay}{\parEmbedding - \parPowerNorm}}\Bigr\} 
= \constValidString \varepsilon^{- \min\{\frac{\parEigDecay}{\parSourceCond - \parPowerNorm},\frac{\parEigDecay}{\parEmbedding - \parPowerNorm}\}}
= \constValidString \varepsilon^{-u}\;\;.
\end{equation*}
Note that there is some $m\in\N$ satisfying this bound since we ensure $\constValidString \varepsilon^{-u}\geq 1$ by choosing $0<\varepsilon\leq\varepsilon_1 \coloneqq \min\{1,\constValidString^{\sfrac{1}{u}}\}$.
In the case of $\parPowerNorm\geq\parEmbedding$, \eqref{eq:res:embedding_property} and \eqref{eq:res:eigenvalue_decay} implies
\begin{align*}
\|f_\omega\|_{L_\infty(\nu)}^2
&\leq\constEmbedding^2\|f_\omega\|_{\parEmbedding}^2 
\leq \frac{32\varepsilon}{m} \constEmbedding^2 \sum_{i=1}^m \eigw_{i+m}^{\parPowerNorm-\parEmbedding} 
\leq 32\constEmbedding^2\, \varepsilon \eigw_{m}^{\parPowerNorm-\parEmbedding}\\ 
&\leq 32\constEmbedding^2 \constEigDecay^{\parPowerNorm-\parEmbedding} \varepsilon m^{-\frac{\parPowerNorm-\parEmbedding}{\parEigDecay}} 
\leq 32\constEmbedding^2 \constEigDecay^{\parPowerNorm-\parEmbedding} \varepsilon
\leq \constInftyBound^2
\end{align*}
for all $m\geq 1$ and $0<\varepsilon\leq\sfrac{\constInftyBound^2}{(32\constEmbedding^2\constEigDecay^{\parPowerNorm - \parEmbedding})}$. Since $\parPowerNorm\geq\parEmbedding$ and $\parSourceCond>\parPowerNorm$ implies $\parSourceCond>\parEmbedding$ and $u=\frac{p}{\parSourceCond - \parPowerNorm}$, both conditions, $\|f_\omega\|_{\parSourceCond} \leq \constSourceCond$ and $\|f_\omega\|_{L_\infty(\nu)} \leq \constInftyBound$, are satisfied for $m\leq\constValidString\varepsilon^{-u}$ and $0<\varepsilon\leq\varepsilon_1$ with $\constValidString \coloneqq \constValidString_1$ and $\varepsilon_1 \coloneqq \min\{\sfrac{\constInftyBound^2}{(32\constEmbedding^2\constEigDecay^{\parPowerNorm - \parEmbedding})}, \constValidString_1^{\sfrac{1}{u}}\}$.
\end{proof}

If $\omega'=(\omega_1',\ldots, \omega_m')\in\{0,1\}^m$ is an other binary string, we investigate the norm of the difference $f_\omega-f_{\omega'}$. To this end, we set $v \coloneqq \sfrac{\parPowerNorm}{\parEigDecay}$ and use \eqref{eq:res:eigenvalue_decay}
\begin{equation}\label{eq:lower:L2bound}
\|f_\omega - f_{\omega'}\|_{L_2(\nu)}^2 
= \frac{32\varepsilon}{m}\sum_{i=1}^m (\omega_i - \omega_i')^2\eigw_i^\parPowerNorm
\leq 32\varepsilon\eigw_m^\parPowerNorm
\leq 32\constEigDecay^\parPowerNorm\,\varepsilon m^{-v}\;\;.
\end{equation}
In order to obtain a lower bound on the $\parPowerNorm$-power norm, we assume $\sum_{i=1}^m(\omega_i - \omega_i')^2\geq\sfrac{m}{8}$, i.e.\ the distance between $\omega$ and $\omega'$ is \emph{large}:
\begin{equation}\label{eq:lower:lower_bound}
\|f_\omega - f_{\omega'}\|_{\parPowerNorm}^2 = \frac{32\varepsilon}{m}\sum_{i=1}^m(\omega_i - \omega_i')^2 \geq 4\varepsilon\;\;.
\end{equation}
The following lemma is from \cites[Lemma~2.9]{Ts2009} and claims that there are many binary strings with large distances.

\begin{lem}[Gilbert-Varshamov Bound]\label{lem:lower:gilbert_varshamov}
For $m\geq 8$ there exists some $M \geq 2^{\sfrac{m}{8}}$ and some binary strings $\omega^{(0)},\ldots,\omega^{(M)}\in\{0,1\}^m$ with $\omega^{(0)}=(0,\ldots,0)$ and 
\begin{equation}\label{eq:proof:lower:large_distance}
\sum_{i=1}^m \bigl(\omega^{(j)}_i - \omega^{(k)}_i\bigr)^2 \geq \sfrac{m}{8}
\end{equation}
for all $j\not=k$, where $\omega^{(k)}=(\omega^{(k)}_1,\ldots,\omega^{(k)}_m)$.
\end{lem}


\begin{myproof}[\jmlrtext{}{Proof }of Lemma~\ref{lem:proof:lower:bound}]
Using the constants $\constValidString>0$ and $0<\varepsilon_1\leq 1$ from Lemma~\ref{lem:proof:lower:valid_strings} we define $\varepsilon_0 \coloneqq \min\{\varepsilon_1,(\sfrac{\constValidString}{9})^{\sfrac{1}{u}}\}$ and $m_\varepsilon \coloneqq \lfloor\constValidString \varepsilon^{-u}\rfloor$. Now, we fix an $n\geq 1$ and an $0<\varepsilon\leq\varepsilon_0$. Since $m_\varepsilon\geq 9$, Lemma~\ref{lem:lower:gilbert_varshamov} yields at least $M_\varepsilon \coloneqq \lceil 2^{\sfrac{m_\varepsilon}{8}}\rceil \geq 3$ binary strings $\omega^{(0)},\omega^{(1)},\ldots,\omega^{(M_\varepsilon)}\in\{0,1\}^{m_\varepsilon}$ satisfying \eqref{eq:proof:lower:large_distance}. According to Lemma~\ref{lem:proof:lower:valid_strings}, for $j=0,1,\ldots, M_\varepsilon$, the corresponding functions $f_j \coloneqq f_{\omega^{(j)}}$ defined in \eqref{eq:proof:lower:function} satisfy the bounds $\|f_j\|_{L_\infty(\nu)}\leq\constInftyBound$ and $\|f_j\|_\parSourceCond\leq\constSourceCond$. Consequently, for $j=0,1,\ldots, M_\varepsilon$, the corresponding probability distribution $P_j \coloneqq P_{f_j}$ defined in \eqref{eq:proof:lower:measure} satisfies  $\|\optFZ{P_j}\|_{L_\infty(\nu)}\leq\constInftyBound$ and \eqref{eq:res:source_condition} w.r.t.\ $\parSourceCond,\constSourceCond$. Recall that $P_j$ additionally satisfies $(P_j)_X=\nu$ and \eqref{eq:res:moment_condition} w.r.t.\ $\varb,\supb$ according to Lemma~\ref{lem:proof:lower:moment_condition}. It remains to prove the Statements~\eqref{eq:proof:lower:size},\eqref{eq:proof:lower:packing}, and \eqref{eq:proof:lower:lower_bound}. 
Due to the definitions of $M_\varepsilon$, $m_\varepsilon$ and $m_\varepsilon\geq 9$ we get $\sfrac{8\constValidString}{9}\;\varepsilon^{-u} \leq m_\varepsilon\leq \constValidString\varepsilon^{-u}$ and 
\begin{equation*}
2^{\sfrac{\constValidString}{9}\; \varepsilon^{-u}} 
\leq 2^{\sfrac{m_\varepsilon}{8}}
\leq M_\varepsilon
\leq 2^{\sfrac{m_\varepsilon}{4}} 
\leq 2^{\sfrac{\constValidString}{3}\;\varepsilon^{-u}}\;\;.
\end{equation*}
Consequently, \eqref{eq:proof:lower:size} is satisfied for $\constLB_2 \coloneqq \sfrac{\constValidString}{9}$. 
The inequality in \eqref{eq:proof:lower:packing} is a consequence of our choice of the binary strings with \eqref{eq:proof:lower:large_distance} and the inequality in \eqref{eq:lower:lower_bound}. 
Lemma~\ref{lem:lower:kullback_leibler} and \eqref{eq:lower:L2bound}  yield
\[ 
\frac{1}{M_\varepsilon}\sum_{j=1}^{M_\varepsilon} K(P_{j}^n,P_{0}^n) = \frac{n}{2\varLB^2 M_\varepsilon}\sum_{j=1}^{M_\varepsilon} \|f_j - f_0\|_{L_2(\nu)}^2 
\leq \frac{16 \constEigDecay^\parPowerNorm}{\varLB^2}\, n\varepsilon m_\varepsilon^{-v}\;\;.
\]
Furthermore, using $m_\varepsilon\geq\sfrac{8\constValidString}{9}\;\varepsilon^{-u}$ we find
\begin{equation*}
\frac{1}{M_\varepsilon}\sum_{j=1}^{M_\varepsilon} K(P_{j}^n,P_{0}^n)
\leq \constLB_3 n\varepsilon^{1 + uv} \eqqcolon \alpha_\ast
\end{equation*}
with $\constLB_3 \coloneqq \frac{16 \constEigDecay^\parPowerNorm 9^{v}}{\varLB^2(8\constValidString)^{v}}$. For a measurable function $\Psi:(X\times \R)^n\to \{0,1,\ldots,M_\varepsilon\}$, Theorem~\ref{thm:lower:general} and $M_\varepsilon\geq 2^{\constLB_2\varepsilon^{-u}}$ from \eqref{eq:proof:lower:size} yields 
\begin{align*}
\max_{j=0,1,\ldots,M_\varepsilon} P_j^n\bigl(D:\ \Psi(D)\not=j\bigr) 
&\geq \frac{\sqrt{M_\varepsilon}}{1 + \sqrt{M_\varepsilon}} \biggl(1 - \frac{3\constLB_3 n\varepsilon^{1 + uv}}{\log(M_\varepsilon)} - \frac{1}{2\log(M_\varepsilon)}\biggr)\\
&\geq \frac{\sqrt{M_\varepsilon}}{1 + \sqrt{M_\varepsilon}} \biggl(1 - \frac{3\constLB_3 }{\constLB_2\log(2)}n\varepsilon^{1 + uv+u} - \frac{1}{2\log(M_\varepsilon)}\biggr)\;\;.
\end{align*}
Since $1 + u v + u = \frac{\max\{\parEmbedding,\parSourceCond\} + \parEigDecay}{\max\{\parEmbedding,\parSourceCond\} - \parPowerNorm}$, this gives us \eqref{eq:proof:lower:lower_bound} for $\constLB_1 \coloneqq \frac{3\constLB_3}{\constLB_2\log(2)}$.
\end{myproof}

Now, the proof of Theorem~\ref{thm:res:lower_rate} remains an application of Lemma~\ref{lem:proof:lower:bound} and the general reduction scheme from \citet[Section~2.2]{Ts2009}.

\begin{myproof}[\jmlrtext{}{Proof }of Theorem~\ref{thm:res:lower_rate}]
Let $D\mapsto\optRegFD$ be a (measurable) learning method. Furthermore, we use the notation of Lemma~\ref{lem:proof:lower:bound}, set $r \coloneqq \frac{\max\{\parEmbedding,\parSourceCond\} - \parPowerNorm}{\max\{\parEmbedding,\parSourceCond\} + \parEigDecay}$, and fix $\tau>0$ and $n\geq 1$ with $\varepsilon_n \coloneqq \tau n^{-r} \leq \varepsilon_0$. It remains to show that there is a distribution $P$ which is difficult to learn for the considered learning method. Lemma~\ref{lem:proof:lower:bound}, for $\varepsilon=\varepsilon_n$, provides possible candidates $P_0,P_1,\ldots,P_{M_n}$, with $M_n \coloneqq M_{\varepsilon_n}$, each satisfying the requirements of Theorem~\ref{thm:res:lower_rate}. Next, we estimate the left hand side of the inequality in \eqref{eq:proof:lower:lower_bound}. To this end, we consider the measurable function $\Psi:(X\times \R)^n \to \{0,1,\ldots,M_n\}$ defined by
\begin{equation}\label{eq:proof:lower:psi}
\Psi(D) \coloneqq \argmin_{j=0,1,\ldots,M_n} \|[f_D]_\nu - f_j\|_{\parPowerNorm}\;\;.
\end{equation}
For $j\in\{0,1,\ldots,M_n\}$ and $D\in(X\times \R)^n$ with $\Psi(D)\not=j$ we have
\[ 
2\sqrt{\varepsilon_n} 
\leq \|\optFZ{P_{\Psi(D)}} - \optFZ{P_j}\|_{\parPowerNorm} 
\leq \|\optFZ{P_{\Psi(D)}} - [f_D]_\nu\|_{\parPowerNorm} + \|[f_{D}]_\nu - \optFZ{P_j}\|_{\parPowerNorm} \leq 2\|[f_{D}]_\nu - \optFZ{P_j}\|_{\parPowerNorm}\;\;.
\]
Consequently, for all $j=0,1,\ldots,M_n$ we find
\begin{equation*}
P_j^n\bigl(D:\Psi(D)\not=j\bigr) \leq P_j^n\bigl(D: \|[f_D]_\nu - \optFZ{P_j}\|_{\parPowerNorm}^2 \geq \varepsilon_n\bigr)\;\;. 
\end{equation*}
According to \eqref{eq:proof:lower:lower_bound}, for $\Psi$ defined in \eqref{eq:proof:lower:psi}, we have
\begin{align*}
\max_{j=0,1,\ldots,M_n}P^n\bigl(D: \|[f_D]_\nu - \optFP\|_{\parPowerNorm}^2 \geq \varepsilon_n\bigr)
&\geq \max_{j=0,1,\ldots,M_n}P^n\bigl(D:\Psi(D)\not=j\bigr)\\
&\geq \frac{\sqrt{M_n}}{\sqrt{M_n} + 1} \Bigl(1 - \constLB_1\tau^{\sfrac{1}{r}} - \frac{1}{2\log(M_n)}\Bigr)\;\;.
\end{align*}
Since $M_n\to\infty$ for $n\to\infty$ we can choose $n$ sufficiently large such that the right hand side is bounded from below by $1-2\constLB_1\tau^{\sfrac{1}{r}}$.
\end{myproof}

\appendix


\section{Auxiliary Results and Concentration Inequalities}\label{sec:apx}

\begin{lem}\label{lem:apx:estimate}
Let, for $\lambda>0$ and $0\leq\alpha\leq1$, the function $f_{\lambda,\alpha}:[0,\infty)\to\R$ be defined by $f_{\lambda,\alpha}(t) \coloneqq \sfrac{t^\alpha}{(\lambda + t)}$. In the case $\alpha=0$ the function $f_{\lambda,\alpha}$ is decreasing and in the case of $\alpha=1$ the function $f_{\lambda,\alpha}$ is increasing. Furthermore, the supremum of $f_{\lambda,\alpha}$ satisfies the following bound
\[
\sfrac{\lambda^{\alpha-1}}{2}  \leq \sup_{t\geq 0} f_{\lambda,\alpha}(t) \leq \lambda^{\alpha-1}\;\;.
\]
In the case of $0<\alpha<1$ the function $f_{\lambda,\alpha}$ attain its supremum at $t^\ast \coloneqq \sfrac{\lambda\alpha}{(1-\alpha)}$.
\end{lem}

\begin{proof}
In order to prove this statement we use the derivative of $f_{\lambda,\alpha}$, which is given by
\[ 
f_{\lambda,\alpha}' (t) = \frac{\alpha t^{\alpha -1}(\lambda + t) - t^\alpha}{(\lambda + t)^2}\;\;.
\]
For $\alpha=0$ we have $f_{\lambda,\alpha}' (t) = -(\lambda + t)^{-2} < 0$ and hence $\sup_{t\geq 0}f_{\lambda,\alpha}(t) = f_{\lambda,\alpha}(0) = \lambda^{\alpha-1}$. For $\alpha=1$ we have $f_{\lambda,\alpha}' (t) = \lambda (\lambda + t)^{-2} > 0$ and hence $\sup_{t\geq 0}f_{\lambda,\alpha}(t) = \lim_{t\to\infty}f_{\lambda,\alpha}(t) = 1=\lambda^{\alpha-1}$. For $0<\alpha<1$ the derivative $f_{\lambda,\alpha}'$ has a unique root at $t^\ast = \sfrac{\alpha\lambda}{(1-\alpha)}$. Since $f_{\lambda,\alpha}(0)=0$ and $\lim_{t\to\infty}f_{\lambda,\alpha}(t)=0$ holds, $f_{\lambda,\alpha}$ attains its global maximum at $t^\ast$ and
\[ 
\sup_{t\geq 0}f_{\lambda,\alpha}(t) = f_{\lambda,\alpha}(t^\ast) = \lambda^{\alpha-1} \alpha^\alpha(1-\alpha)^{1-\alpha}\;\;.
\]
Since $g(\alpha) \coloneqq \alpha^\alpha(1-\alpha)^{1-\alpha}$ is bounded by $1$ the upper bound follows. The derivative 
\[ 
g'(\alpha) = g(\alpha) \log\biggl(\frac{\alpha}{1-\alpha}\biggr)
\]
of $g$ has a unique root at $\alpha=\sfrac{1}{2}$ and hence the lower bound follows from $g(\alpha)\geq g(\sfrac{1}{2}) = \sfrac{1}{2}$ for all $0<\alpha< 1$.
\end{proof}

The following Bernstein type inequality for Hilbert space valued random variables is due to \citet{PiSa1986}. However we use a version from \cites[Proposition~2]{CaDe2007}.

\begin{thm}[Bernstein's Inequality]\label{thm:apx:bernstein}
Let $(\Omega,\mathcal{B},P)$ be a probability space, $H$ be a separable Hilbert space, and $\zvi:\Omega\to H$ be a random variable with 
\[ 
\E_P \|\zvi\|_H^m \leq\frac{1}{2}m!\varb^2\supb^{m-2}
\]
for all $m\geq 2$. 
Then, for $\tau\geq 1$ and $n\geq 1$, the following concentration inequality is satisfied
\[ 
P^n\biggl((\omega_1,\ldots,\omega_n)\in\Omega^n:\ \Bigl\|\frac{1}{n}\sum_{i=1}^n \zvi(\omega_i) - \E_P \zvi \Bigr\|_H^2 \geq 32\frac{\tau^2}{n}\biggl(\varb^2+ \frac{\supb^2}{n}\biggr) \biggr) \leq 2 e^{-\tau}\;\;.
\]
\end{thm}

\begin{proof} 
The $m$-th moment of the centered random variable $\zvi - \E_P\zvi$ is bounded by
\begin{equation*}
\E_P \|\zvi - \E_P\zvi\|_H^m 
\leq 2^{m-1} \bigl(\E_P\|\zvi\|_H^m + \|\E_P\zvi\|_H^m\bigr) 
\leq 2^m\E_P\|\zvi\|_H^m 
\leq \frac{1}{2}m! (2\supb)^{m-2} 4\varb^2\;\;. 
\end{equation*}
Since we consider the squared norm the assertion is a direct consequence of \cites[Proposition~2]{CaDe2007} with $\eta = 2 e^{-\tau}$, $L=2\supb$, and $\sigma^2 = 4\varb^2$.
\end{proof}

The following Bernstein type inequality for Hilbert-Schmidt operator valued random variables is due to \citet{Mi2017}. However we use a version from \cites[Lemma~26]{LiCe2018}, see also \citet{Tr2015} for an introduction to this topic.

\begin{thm}\label{thm:apx:bernstein_operator}
Let $(\Omega,\mathcal{B},P)$ be a probability space, $H$ be a separable Hilbert space, and $\zvi:\Omega\to\mathcal{L}_2(H)$ be a random variable with values in the set of self-adjoint Hilbert-Schmidt operators. 
Furthermore, let the operator norm be $P$-a.s.\ bounded, i.e.\ $\|\zvi\|\leq B$ $P$-a.s. and $V$ be a self-adjoint positive semi-definite trace class operator with $\E_P(\zvi^2) \preccurlyeq V$, i.e.\ $V - \E_P(\zvi^2)$ is positive semi-definite. 
Then, for $g(V) \coloneqq \log\bigl(\sfrac{2 e \tr(V)}{\|V\|}\bigr)$, $\tau\geq 1$, and $n\geq 1$, the following concentration inequality is satisfied
\[ 
P^n\biggl((\omega_1,\ldots,\omega_n)\in\Omega^n:\ \Bigl\|\frac{1}{n}\sum_{i=1}^n \zvi(\omega_i) - \E_P \zvi \Bigr\|\geq \frac{4 \tau B g(V)}{3 n}+ \sqrt{\frac{2\tau\|V\|\,g(V)}{n}} \biggr) \leq 2 e^{-\tau}\;\;.
\]
\end{thm}

Recall that $\|V\|$ denotes the operator norm and $\tr$ the trace operator.

\begin{proof}
This is a direct consequence of Lemma~26 from \cite{LiCe2018} with $\delta = 2 e^{-\tau}$ applied to the centered random variable $\zvi - \E_P\zvi$. Furthermore, we used $\|\zvi - \E_P\zvi\| \leq 2 B$ and $\E_P(\zvi - \E_P\zvi)^2 \preccurlyeq \E_P (\zvi^2)\preccurlyeq V$. Finally, $\beta$ defined by \cites[Lemma~26]{LiCe2018} can be bounded by 
\[ 
\beta 
\coloneqq \log\biggl(\frac{4\tr(V)}{\|V\|\delta}\biggr) 
= \log\biggl(\frac{2\tr(V)}{\|V\|}\biggr) + \tau
\leq \tau g(V)
\] 
because of $\tau\geq 1$ and $\log\bigl(\sfrac{2\tr(V)}{\|V\|}\bigr)>0$.
\end{proof}


\begin{singlespace}

\bibliographystyle{../_style/bibliographystyle}
\bibliography{../_style/literatur}

\end{singlespace}


\end{document}